\documentclass[10pt, conference]{IEEEtran}

\usepackage{graphicx} 
\usepackage{amsmath}
\usepackage{amsfonts}
\usepackage{fancyhdr}
\usepackage{datetime}
\usepackage{comment}
\usepackage{centernot}
\usepackage{float}
\usepackage{algorithm}
\usepackage{algpseudocode} 
\usepackage[authoryear]{natbib}
\usepackage[dvipsnames]{xcolor}
\definecolor{deeppink}{RGB}{255,20,147}
\definecolor{royalblue}{HTML}{00B4CE}
\usepackage[colorlinks=true, linkcolor=WildStrawberry, citecolor=WildStrawberry, urlcolor=WildStrawberry]{hyperref}
\usepackage{cuted}
\usepackage{multicol}

\renewcommand\baselinestretch{1}
\DeclareMathOperator{\EX}{\mathbb{E}}

\title{Observability conditions for neural state-space models with eigenvalues and their roots of unity}
\author{
  Andrew Gracyk
  \\
  Department of Mathematics, Purdue University
  \\
  West Lafayette, IN 47907, United States
  \\
  \texttt{agracyk@purdue.edu}
}
\date{}

\begin{document}

\maketitle

\begin{abstract}
\noindent We operate through the lens of ordinary differential equations and control theory to study the concept of observability in the context of neural state-space models and the Mamba architecture. We develop strategies to enforce observability, which are tailored to a learning context, specifically where the hidden states are learnable at initial time, in conjunction to over its continuum, and high-dimensional. We also highlight our methods emphasize eigenvalues, roots of unity, or both. Our methods effectuate computational efficiency when enforcing observability, sometimes at great scale. We formulate observability conditions in machine learning based on classical control theory and discuss their computational complexity. Our nontrivial results are fivefold. We discuss observability through the use of permutations in neural applications with learnable matrices without high precision. We present two results built upon the Fourier transform that effect observability with high probability up to the randomness in the learning. These results are worked with the interplay of representations in Fourier space and their eigenstructure, nonlinear mappings, and the observability matrix.  We present a result for Mamba that is similar to a Hautus-type condition, but instead employs an argument using a Vandermonde matrix instead of eigenvectors. Our final result is a shared-parameter construction of the Mamba system, which is computationally efficient in high exponentiation. We develop a training algorithm with this coupling, showing it satisfies a Robbins-Monro condition under certain orthogonality, while a more classical training procedure fails to satisfy a contraction with high Lipschitz constant.
\end{abstract}

\medskip
\noindent
\textbf{Key words.} Mamba, state-space models, neural ordinary differential equations, observability, control theory, eigenvalues, roots of unity, diagonalizable, Robbins-Monro, Hautus lemma, Vandermonde, contraction mapping, Bauer-Fike

\tableofcontents

\section{Introduction}

\begin{figure*}[htbp]
  \vspace{0mm}
  \centering
  \includegraphics[scale=0.6]{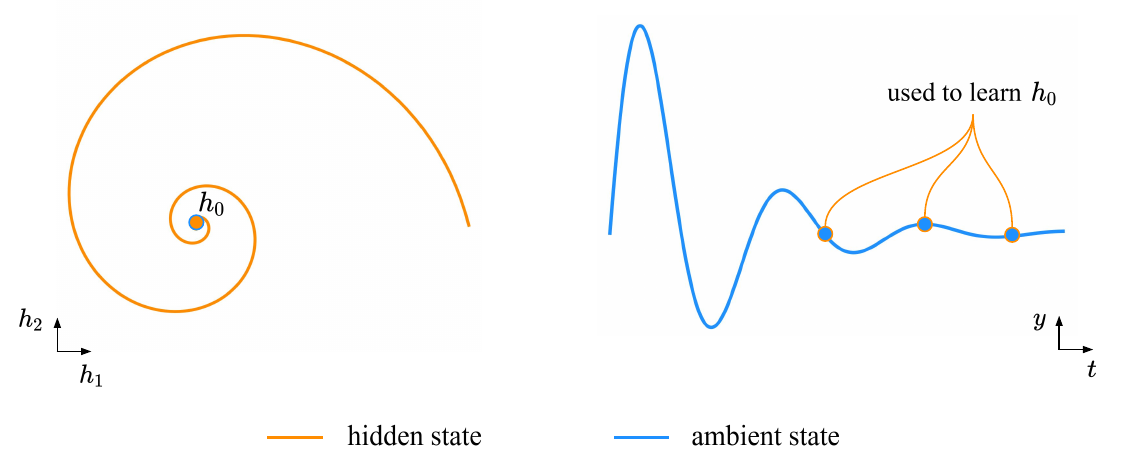}
  \caption{We illustrate observability: ambient output states may be used to learn the initial hidden state. We focus on scenarios when this hidden state is learnable, i.e. not fixed, otherwise the learning task of observability is trivial.}
  \label{fig:obs_diagram}
\end{figure*}

We design methods in ordinary differential equations to effectuate observability \cite{evans_control_course} \cite{dahleh2011lectures} in neural applications, placing significance on the Mamba architecture \cite{gu2024mambalineartimesequencemodeling} \cite{hu2024statespacemodelsaccurateefficient} \cite{hamdan2024sparsemambaintroducingcontrollability} \cite{jafari2025mambalrpexplainingselectivestate} \cite{anthony2024blackmambamixtureexpertsstatespace}, but also considering more traditional neural state-space systems \cite{gu2022efficientlymodelinglongsequences} \cite{alonso2024statespacemodelsfoundation} \cite{forgione2023empirical} \cite{smith2023simplifiedstatespacelayers} that rely on the Fourier transform. To begin, we provide a description the state-space model. A state-space model has more broad relevance in mathematics beyond machine learning, and is a coupled linear ordinary differential equation dynamical system with a learnable hidden stage undergoing concurrent dynamics to underscore and characterize the ambient solution of the data. For us, observability is the ability to utilize output information to infer $h_0$, the initial latent state, although this qualitative definition has more rigorous mathematical significance. Moreover, by learning the initial hidden state, observability ensures inference of the hidden evolution for all times, solvable from the governing output, given that the underlying state-space system is known and allows computation throughout its flow.

\vspace{2mm}

Observability has meaningful relevance in engineering, physics, and science, beyond its significance in theoretical mathematics  \cite{pickard2025dynamicsensorselectionbiomarker} \cite{Montanari_2022} \cite{Vaidya2007ObservabilityGF} \cite{karamched2024observabilitycomplexsystemsconserved} \cite{Montanari_2021} \cite{doi:10.1073/pnas.1215508110} \cite{Kunapareddy2018RecoveringOV} \cite{PhysRevE.86.026205} \cite{Rafieisakhaei_2017}. By intertwining these widened scientific contexts with machine learning, our methods provide a rich basis when combining practical, efficient computation with a more broad scientific scope. Thus, we provide foundations for analyzing high-dimensional data with latent, dynamic representations in real-world environments. To exemplify, \cite{Vaidya2007ObservabilityGF} combines control theory with stochastic systems and Bayesian inference. \cite{Kunapareddy2018RecoveringOV} and \cite{Rafieisakhaei_2017} places greater emphasis in practical engineering with sensors and robotics respectively. \cite{PhysRevE.86.026205} demonstrates more mathematical-based requisitions with regard to chaos theory.

\vspace{2mm}

\subsection{Related work}

Dominating literature on neural state-space models and Mamba are \cite{gu2022efficientlymodelinglongsequences} \cite{gu2024mambalineartimesequencemodeling}, although, for other foundational literature on state-space models, we refer to \cite{forgione2023empirical} \cite{ansari2023neuralcontinuousdiscretestatespace},  \cite{hu2024statespacemodelsaccurateefficient} \cite{cheng2024mambaneuraloperatorwins} for neural operator literature, \cite{toloubidokhti2022neuralstatespacemodelinglatent}\cite{bae2025causalmambascalableconditionalstate} for causal learning, on baseline improvement \cite{lahoti2026mamba3improvedsequencemodeling} with Mamba-3 and \cite{nguyen2026muonssmorthogonalizingstatespace} with MuonSSM, on diffusion \cite{ruder2026diffusiondrivenstatespacemodels}. 

\vspace{2mm}

For greater relevance to our results, \cite{gilyen2025arbeitsgemeinschaft} \cite{alexis2024quantumsignalprocessingnonlinear} are signal-analytic literature that emphasize the nonlinearities and intricacies involved in the Fourier transform, although our results are less typical in literature but well-founded in strategies from linear algebra. Permutations affecting rowspaces is well established in traditional theory, but also has broader applications such as in \cite{balan2022permutationinvariantrepresentationsapplications} \cite{yilmaz2026arrowflowhierarchicalmachinelearning} in more modern contexts. Many of our techniques are adaptations of well-established theorems in classical mathematics, such as through the Bauer-Fike Theorem \cite{katende2024nonlineargeneralizationbauerfiketheorem} and the Hautus lemma, in which we elaborate more later. Lastly, our final result in \ref{app:codependence} relies on certain gradient descent scaling, reasonable in optimization literature such as \cite{ward2021adagradstepsizessharpconvergence} \cite{arjevani2022lowerboundsnonconvexstochastic} \cite{carmon2017lowerboundsfindingstationary} \cite{carmon2019lowerboundsfindingstationary}, and a Robbins-Monro condition \cite{fraysse2013robbinsmonroprocedureestimationparametric}, another known mathematical mechanism.

\vspace{2mm}

\subsection{Contributions}

We detail our contributions as follows:
\begin{itemize}
\item We provide fundamental machine learning perspectives to enforce observability based on well-established, foundational results, including relating to the observability matrix (which is closely related to the discrete Observability Gramian based on definition), and the Hautus lemma. These strategies are simple and come at high computational expense.
\item We present a series of results based on permutations, which are inherent to the enforcement of observability since a wide matrix can never be full column rank. While more general matrices often have permutation-like qualities in an observability context, they are not necessarily permutation matrices in regard to column orthogonality and root of unity eigenvalues. A key cost of this enforcement is that it is restrictive over the parameter space with respect to $A$.
\item We give results for more traditional, non-Mamba state-space models whose underlying mechanisms are supported with the Fourier transform \cite{gu2022efficientlymodelinglongsequences}. We give two results: one regarding a generalized discretized impulse response, and another regarding the setup as in \cite{gu2022efficientlymodelinglongsequences}. These results are built upon diagonalization arguments and kernel conditions. Since the Fourier transform involves exponentiation and is equivalent to an observability matrix-type formulation, we have the necessary grounds to establish the connections. As we will see, these results have general relations to permutation-type properties or roots of unity.
\item Towards Mamba, we provide a result that is quite similar to a Hautus-type condition in the sense that no eigenvector is annihilated by $C$. In fact, we will show equivalence of this result with the Hautus lemma if this proof were to be extended analytically over through use of machine learning. Our argument relies on efficiently extracting select rows of the observability matrix by leveraging Kronecker products and a Vandermonde matrix. We demonstrate, by use of the Vandermonde matrix, we improve efficiency. Normally, matrix multiplication would dominate the computational complexity of this method. By constructing a dependence on $C$ and the diagonalization of $A$, we eliminate this dependence for matrix multiplication, in which our method outperforms significantly.
\item Our last result is built on a parameter-reduction among the control matrices by introducing a coupling. The number of parameters of the state-space system is drastically reduced and is of notable scale depending on the matrix dimensions. We develop an algorithm that satisfies a Robbins-Monro condition \cite{ward2021adagradstepsizessharpconvergence} under enforceable conditions. Furthermore, we show a typical training algorithm built on this coupling has high Lipschitz constant a typical metric, thus convergence is not supported in this sense because a contraction mapping is not satisfied. We remark the major drawback of this method is that it is also restrictive over the parameter space.
\end{itemize}

\section{Preliminaries}

\subsection{Control theory}
\label{sec:control_theory}

We provide background on control theory to establish our framework. First, we define the \textit{reachable set} at $t$ and the \textit{overall reachable set} respectively as
\begin{align}
\label{eqn:reachable_set}
&\mathcal{C}(t) = \text{set of } \ y_0 \ \text{s.t. } \exists \text{ a control so} \ y(t) = 0
\\
& \mathcal{C} = \bigcup_{t \geq 0} \mathcal{C}(t) .
\end{align}
for some function $\alpha : [0, \infty) \rightarrow A$ we call a \textit{control}. We say an ODE system is \textit{controllable} if $\mathcal{C} = \mathbb{R}^n$. We say an ODE system
\begin{equation}
\label{eqn:ODE_system}
\begin{cases}
\dot{h}(t) = A h(t) + Bx(t)
\\
y(t) = C h(t) + Dx(t)
\end{cases} 
\end{equation}
is observable if the system
\begin{equation}
\dot{z}(t) = A^T z(t) + C^T \alpha(t)
\end{equation}
is controllable \cite{evans_control_course}, where $\alpha$ is the control. We will restrict
\begin{equation}
A \in \mathbb{R}^{n \times n}, B \in \mathbb{R}^{n \times m}, C \in \mathbb{R}^{m \times n}, n \geq m   
\end{equation}
for the remainder of this research. We will not work with $D$ at all, and we will regard it is as irrelevant for us and it does not affect observability (neither does $B$ typically, but our results will sometimes involve $B$). Moreover, the system is \textit{observable} if knowledge of $y$ on any $[0,t]$ allows us to compute $h(0)$ under the system.

\vspace{2mm}

Generally, it is not reasonable to examine all of $\mathbb{R}^d$ to enforce observability. We can greatly simplify observability with conditions instead of examining controllability and the reachable set. In particular, we can examine the \textit{observability matrix} to examine observability. Define the observability matrix as
\begin{equation}
\label{eqn:obs_matrix}
\mathcal{O} = \begin{pmatrix} C  \\ CA \\ \vdots \\
C A^{n-1}
\end{pmatrix}
\in \mathbb{R}^{(nm) \times n} .
\end{equation}
It is a known result that the system is observable if this matrix is full column rank. Furthermore, we define the unobservable subspace $N$ to be the kernel of the map $\mathcal{G}$ to be
\begin{align}
\mathcal{G} : h(0) \rightarrow C e^{At} h(0) .
\end{align}
We say the system is observable if and only if $N$ is the zero vector, i.e. if we have
\begin{equation}
C e^{At} h(0) = 0 \implies h(0) = 0 \ \text{only} ,
\end{equation}
then the system is observable.

\subsection{The neural state-space model}
\label{sec:neural_ssm}

The neural state-space model is outlined by the ODE system of equations \cite{gu2022efficientlymodelinglongsequences}
\begin{equation}
\label{eqn:state_space_model}
\begin{cases}
\dot{h}(t) = A h(t) + Bx(t)
\\
y(t) = C h(t) + Dx(t)
\end{cases} .
\end{equation}
The parameters used in gradient descent updates are the matrices $A,B,C,D$. Note there is a well-known recurrent formulation of this system for discrete data, but methods regarding this are computationally impracticable. Ultimately, we do not consider recurrent methodology in our investigation. Instead, we will consider the convolutional state-space system
\begin{equation}
\label{eqn:state_space_model}
\begin{cases}
h_k = \sum_{i=0}^k \overline{A}^i \overline{B} x_{k-i}
\\
y_k = \sum_{i=0}^k \overline{C} \overline{A}^i \overline{B} x_{k-i}
\end{cases} ,
\end{equation}
where we have defined the quantities
\begin{align}
\overline{A} & = ( I - \frac{\Delta}{2} \cdot A)^{-1} ( I + \frac{\Delta}{2} \cdot A )
\\
\overline{B} & = ( I - \frac{\Delta}{2} \cdot A)^{-1} \Delta B
\\
\overline{C} & = C .
\end{align}
Here, $\Delta \in \mathbb{R}^+$ is the step size. There is relevance of the Fourier transform to the state-space model system \cite{gu2022efficientlymodelinglongsequences}, and in our analysis we will be studying
\begin{align}
\hat{K} = \hat{\mathcal{K}}_L(z; \overline{A}, \overline{B}, \overline{C}) 
\end{align}
such that
\begin{align}
\overline{K} & = (C \overline{B}, C \overline{A} \overline{B}, \hdots, C \overline{A}^{L-1} \overline{B})
\\
\hat{K}_j & = \sum_{k=0}^{L-1} \overline{K}_k \exp\{ - 2\pi i \frac{jk}{L} \} 
\\
& = \sum_{k=0}^{L-1} \overline{C} \overline{A}^k \overline{B} \exp \{ -2 \pi i \frac{jk}{L} \} .
\end{align}

\subsection{Mamba setup}
\label{sec:ssm}

The Mamba architecture \cite{gu2024mambalineartimesequencemodeling} proceeds as the state-space model
\begin{equation}
\label{eqn:state_space_model}
\begin{cases}
\dot{h}(t) = A h(t) + Bx(t)
\\
y(t) = C h(t)
\end{cases}
\end{equation}
for hidden state $h(t)$. The hidden state is an intermediary, learned state to conduct the learning task. Using a discretization, our ODE system becomes \cite{gu2024mambalineartimesequencemodeling}
\begin{equation}
\begin{cases}
h_t = \overline{A}' h_{t-1} + \overline{B}' x_t
\\
y_t = C h_t
\end{cases}
\end{equation}
using the equations
\begin{equation}
\overline{A}' = \exp \{ \Delta \cdot A \}, \ \ \ \ \ \overline{B}' = (\Delta \cdot A )^{-1} ( \exp \{ \Delta \cdot A \} - I ) \cdot \Delta B .
\end{equation}
Lastly, we remark we can convert between $A,B$ and $\overline{A}, \overline{B}$ using the inverses
\begin{align}
A & = \frac{1}{\Delta} \log( \overline{A}' )
\\
B & = \frac{1}{\Delta} (\Delta \cdot A) \cdot \overline{B}' \cdot ( \exp\{ \Delta \cdot A \} - I )^{-1} 
\\
& =  A \cdot \overline{B}' \cdot ( \exp\{ \Delta \cdot A \} - I )^{-1} .
\end{align}
Here, $\log$ is the matrix logarithm, which can be computed efficiently using standard computing libraries.

\subsection{Matrix setup}

In our analyses, we will make restrictions on the structure of certain matrices. These are notably ensuring a full rank condition, and the ability to be diagonalized. 

\vspace{2mm}

In practice, we can always enforce a matrix to be full rank by taking the desired block, or entire matrix, to be of the form $Q^T Q + \epsilon I$. The proof follows by taking $v^T (Q^T Q + \epsilon I) v > 0$ and using the rank-nullity theorem. For diagonalization, a sufficient condition is that all eigenvalues are distinct. Alternatively, we can parameterize $P = Q^{-1} \Lambda Q$ directly by constructing $Q, \Lambda$ as the learnable parameters.

\vspace{2mm}

As an additional remark, sometimes the conditions we will establish are satisfied because of imprecision in machine learning. For example, slight numerical rounding error can establish a matrix as full column rank. Thus, practical implementation of our methods implies the conditions are met sufficiently well so they are not automatically satisfied with numerical imprecision.

\section{Introductory methods}
\label{sec:intro_methods}

We present a series of preliminary strategies for observability based on classical, well-known results from control theory.

\subsection{The observability matrix as full column rank}

The most clear method for enforcing observability is to ensure the observability matrix is full column rank, but working with the observability matrix $\mathcal{O} \in \mathbb{R}^{(nm) \times n}$ as it is presents challenges because it is often very high-dimensional in the sense that $nm$ is typically very large. The computational cost of this is $\mathcal{O}^T \mathcal{O}$ is $O(n^3 m + n^3 +(n-2)n^3)$, the first from the matrix multiplication, the second from the determinant, and the third is from computing all powers of $A^k$ needed to formulate the matrix. While modern hardware can certainly do this, there are computational gains to be made. Thus, a very classical objective to enforce observability is to augment any loss function with the term
\begin{equation}
\mathcal{L} = \text{relu} \Bigg( \text{positive constant} - \text{det} \Big( \mathcal{O}^T \mathcal{O} \Big) \Bigg) .
\end{equation}
We will use determinants throughout our analyses. Generally, we require $\text{det}(\mathcal{O}^T \mathcal{O}) > 0$ only, so the small positive constant in the loss is to help ensure nondegeneracy, i.e. it is a small regularization.

\vspace{2mm}

Working with the observability matrix presents other challenges. By taking $\mathcal{O}^T \mathcal{O}$, small entries are magnified in scale by taking the products, which continually yields negligible numbers. Furthermore, determinants scale exponentially, so the effect of these small numbers is amplified even more.

\vspace{2mm}

Note that a well-known from control theory states that $(C,A)$ is observable if and only if $(C,e^{\Delta A})$ is observable assuming $\Delta$ does not cause aliasing, i.e. $\text{Im}(\lambda_i - \lambda_j) \neq \frac{2\pi k}{\Delta} \ \text{for any non-zero integer } k$. In the Mamba architecture specifically, this is notable as we can now exponentiate $e^{\Delta A}$ to high polynomial order $e^{k \Delta A}$ instead of just taking $A^k$, and the result is the same.

\subsection{Loss from the Hautus lemma}
\label{sec:hautus}

Alternatively, and still simply, we can enforce the Hautus lemma in a learning task to achieve observability. The Hautus lemma states if
\begin{equation}
\text{dim} \Bigg( \text{col} \Bigg( \begin{pmatrix} A - \lambda I \\ C 
\end{pmatrix} \Bigg) \Bigg) = n
\end{equation}
for all eigenvalues of $A$, $\lambda$, then the pair $(C,A)$ is observable. In a machine learning task, we take
\begin{align}
\label{eqn:hautus_loss}
\mathcal{L} & = \sum_{j} \text{relu} \Bigg( \text{positive constant} \\
& \ \ \ \  - \text{det} \Bigg( (A - \lambda_j I)^{\dagger} (A - \lambda_j I )  + C^T C \Bigg) \Bigg) .
\end{align}
Computing all eigenvalues is $O(n^3)$, but we only need do this once before formulating each term; however, computing the determinant here is $O(n^3)$, and since this is done $n$ times our leading order is $O(n^4 + n^3 + \text{smaller terms})$. Thus, this loss about as expensive than formulating the observability matrix as is, since this method diverts computing high order exponents of $A$.

\vspace{2mm}

It is notable to us that it is necessary $C v \neq 0$ for all eigenvectors $v$ of $A$. This is because $Av - \lambda I v = 0$, therefore the matrix can never be full column rank if $v$ is in the kernel of $C$. Indeed, we can improve efficiency of this by diagonalizing $A$ and then taking the loss
\begin{equation}
\label{eqn:hautus_efficient}
\mathcal{L} \stackrel{!}{=} \sum_j \text{relu}\Bigg( \text{positive constant} - ||Cv_j||_2^2 \Bigg) .
\end{equation}
Here, $v_j$ are the eigenvectors of $A$. We remark this condition is necessary, not sufficient, and so we have denoted that this loss function is incomplete for observability. Consider the counterexample $A=I = I^3, C = \text{concat}\{ (1 \ 0 \ 1), ( 0 \ 1 \ 1) \}$, in which no eigenvector is mapped to triviality but the system is clearly not observable. The computational cost of this is $O( mn^2 + mn )$, from the $CV$ formulation and the norm respectively. As we will see, we can improve this by not needing to compute $CV$ but instead only needing light conditions on the eigenvalues, thus our methods are superior.

\section{Theoretical results}

\subsection{Observability with permutations}

In this subsection, we discuss ways to enforce observability through the use of permutation matrices. Key advantages of these techniques is that they are computationally efficient, easy to implement, and suitable for both Mamba and non-Mamba state-space models. The key pitfalls of these methods is that they are restrictive over the parameter space with respect to parameter $A$, and thus they are non-expressive; however, we make no restrictions on $B$ at all, and only light restrictions on $C$.

\vspace{2mm}

We say $Q$ has root of unity eigenvalues if the eigenvalues of $Q$ are given by
\begin{align}
& \text{eigenvalues}(Q) = \{ \lambda_i, \hdots, \lambda_d \} = \{ e^{-2 \pi i j / d} : j \in \{1,\hdots,d\} \} \\
& = \Big\{ \cos \Big( \frac{2 \pi j }{d} \Big) - i \sin \Big( \frac{2 \pi j }{d} \Big) : j \in \{1,\hdots,d\} \Big\} .
\end{align}

We can enforce observability by matching the learned matrix $A$ to match distinct root of unity eigenvalues with column and row size conditions. In particular, we minimize the loss
\begin{align}
\label{eqn:permutation_loss}
\mathcal{L} & = \sum_{k=1}^n | \lambda_k(A) - \frac{1}{\sqrt{n}} e^{-2 \pi i k / n} |^2 
\\
& +  \sum_j | \sum_i A_{ij} - 1 | +  \sum_i | \sum_j A_{ij} - 1  | .
\end{align}
If $A$ is parameterized as $V \text{diag}(\lambda_1,\hdots,\lambda_n)V^{-1}$, the computational cost of this loss is $O(2n^2 + \text{negligible terms})$. This is the most efficient loss we will develop

\vspace{2mm}

It is acceptable that this is not learned with extreme precision, i.e. there is some error. We will mathematically demonstrate both why this loss function enforces observability and why an error tolerance is acceptable. In particular, we will elaborate more on all terms on this loss and why this works. 

\vspace{2mm}

\textbf{Lemma 2.} Suppose $Q \in \mathbb{R}_+^{n \times n}$ has distinct root of unity eigenvalues. Suppose 
\begin{equation}
\sum_{i} Q_{ij} = \sum_j Q_{ij} = 1 .
\end{equation}
Then $Q$ is necessarily a permutation matrix.

\vspace{2mm}

\textbf{Theorem 1.} Let $A \in \mathbb{R}^{n \times n}$ be a permutation matrix with distinct root of unity eigenvalues, and let $C \in \mathbb{R}^{m \times n}, n > m$ be the matrix of equation \ref{eqn:ODE_system} (hence not identically the zero matrix). Furthermore, suppose $C$ is nonconstant among permutations in the sense that no column in the column space of $C$ can be formulated with constant coefficients among other linearly independent columns in the space that span the image the of $C$. More specifically,
\begin{gather}
\psi_k \in \{ \{ \psi_i \}_i : \psi_i \ \text{is a column of} \ C \}, 
\\
\psi_k \neq \sum_{j \neq k} \alpha \varphi_j, \text{Im} \Big( \text{span} \{ \varphi_j 
\\
: \varphi_j \ \text{is a linearly independent column of} \ C \} \Big) = \mathbb{R}^m  .   
\end{gather}
Then the pair $(C,A)$ is observable.

\vspace{2mm}

\begin{figure}[h]
  \vspace{0mm}
  \centering
  \includegraphics[scale=0.75]{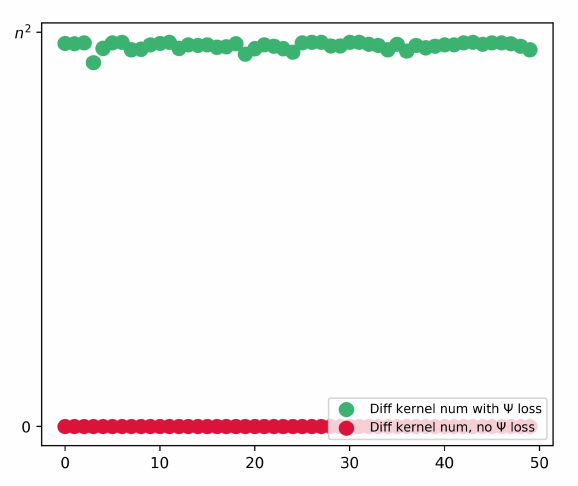}
  \caption{This figure primarily highlights Theorem 4 is valid. First, this figure illustrates the number of distinct kernels across $\Psi^j$ with respect to $j$, with similarity to the figures in the appendix. Higher is better, so our method clearly outperforms. Moreover, it illustrates the loss functions we develop in our Fourier-based theorems have nontrivial significance, and they outperform eigenvalues that are simply random in ensuring the distinct kernel conditions hold. Here, \textcolor{Green}{green} corresponds to the eigenvalues learned with our custom loss functions, while \textcolor{red}{red} corresponds to random eigenvalues, i.e. a baseline. A second concept this figure illustrates is that we can control $\Delta$ predictably. In particular, the same $\Delta$ used in the loss function is the same $\Delta$ used to recreate these kernel results. We chose $\Delta = 0.1$, $n=40,m=25.$}
  \label{fig:kernel_with_Psi_loss}
\end{figure}

\textbf{Lemma 3.} Let $Q$ be a permutation matrix, and let $P$ be a matrix such that
\begin{equation}
| \lambda_i(Q) - \lambda_i(P) | = \epsilon_i ,
\end{equation}
and even moreover, the closest eigenvalue to $\lambda_i(P)$ is $\lambda_i(Q)$. Then there exists a matrix $\Xi$ such that $
\sum_i \epsilon_i^2 = \mathcal{O}(n||\Xi||_2)$
and $P = Q +  \Xi $.

\vspace{2mm}

\textbf{Theorem 2.} Let $C \in \mathbb{R}^{m \times n}$ be the matrix as defined in \ref{eqn:ODE_system}. Let the hypotheses of Theorem 1 on $C$ hold. Let $Q \in \mathbb{R}^{n \times n}$ be a permutation matrix and let $P$ be a matrix with distinct eigenvalues sufficiently close to roots of unity in the sense that $P = Q + \Xi$, $| \lambda_i(Q) - \lambda_i(P)| = \epsilon_i, \sum_i \epsilon_i^2 = \mathcal{O}(||\Xi||_F^2)$. Let
\begin{equation}
\sum_i \Big|\Big| (P^i)^T C^T C P^i - (Q^i)^T C^T C Q^i \Big|\Big|_F = \mathcal{O}( n ||\Xi||_F^2) .
\end{equation}
If $|| \Xi ||_F^2 \leq \epsilon$ holds for sufficiently small $\epsilon$, then the pair $(C,P)$ is observable.

\subsection{Observability with the Fourier transform: simple non-degeneracy conditions}

\textbf{Theorem 3.} Consider the Fast Fourier transform of the matrix $C e^{A k \Delta t} B$, $\mathcal{F}[C e^{A k \Delta t} B]$.
$C \in \mathbb{R}^{m \times n}$ be full row rank, $A \in \mathbb{R}^{n \times n}, B\in \mathbb{R}^{n \times m}, n > m$, where $A$ is full column rank. Suppose $A = V \Lambda V^{-1}$ is diagonalizable. Let $C$ satisfy the permutation-invariant property as in Theorem 1. Suppose
\begin{equation}
\text{ker} \Bigg( CV \Phi^{j_1} V^{-1}  \Bigg) \not\subseteq  \text{ker} \Bigg( CV \Phi^{j_2} V^{-1}  \Bigg) .
\end{equation}
More specifically, let the loss of \ref{eqn:loss_thm4} be exactly satisfied, where $ 1 \leq j_1,j_2 \leq L-1$ and $I - e^{A L \Delta t}$ is full rank. We have defined
\begin{align}
\Phi^j = \text{diag} \Bigg( \frac{ 1 - e^{\lambda_1 \Delta t}}{ 1 - e^{\lambda_1 \Delta t - 2 \pi i \frac{j}{L}}}, \hdots, \frac{ 1 - e^{\lambda_n \Delta t}}{ 1 - e^{\lambda_n \Delta t - 2 \pi i \frac{j}{L}}} \Bigg) .
\end{align}
$\theta_k(\Lambda)$ denotes the angular part of the complex exponentials of the entries of $\Lambda$. Then with high probability the system is observable.

\vspace{2mm}

\textit{Corollary.} We can relax the condition that $(I -e^{A L \Delta t})$ is full column rank by taking
\begin{equation}
\text{det}\Bigg(   \mathcal{F} \Bigg[ C e^{A k \Delta t} B  \Bigg]_j  \Bigg) > 0  
\end{equation}
for all $j$, but generally this is computationally expensive.

\vspace{2mm}

\begin{figure*}[h]
  \vspace{0mm}
  \centering
  \includegraphics[scale=0.68]{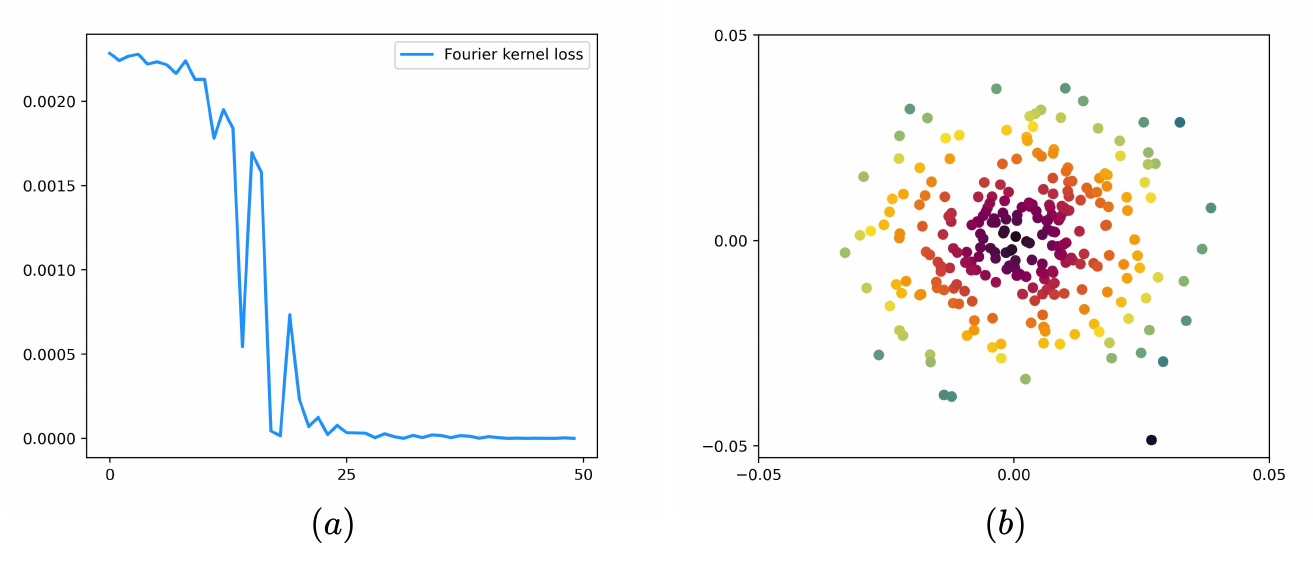}
  \caption{This figures illustrates the Fourier kernel loss as in equation \ref{eqn:fourier_kernel_loss} is solvable, and that $\mathcal{L} = 0$ is attainable. We do so by systemically constructing vectors in the kernel, as is the method in the appendix. In (a), we depict $\mathcal{L}$  with ((x) training iteration mod 2 versus (y) loss). We have chosen $m=250, n=400$ here. We also chose $\text{positive constant} = 0.05$. In (b), we plot the learned eigenvalues in the complex plane. Color represents the norm.}
  \label{fig:losses_obs}
\end{figure*}

\subsection{Observability with a more applicable Fourier transform: Fourier non-degeneracy}
\label{sec:fourier_nondegeneracy}

\textbf{Theorem 4.} Consider the Fast Fourier transform of the matrix $\overline{K}$. Suppose $C \in \mathbb{R}^{m \times n}$ is full row rank, and $A \in \mathbb{R}^{n \times n}$ full column rank. Suppose $A = V \Lambda V^{-1}$ is diagonalizable with all distinct eigenvalues. Suppose it is true that 
\begin{align}
\label{eqn:thm4_kernel_condition}
\text{ker} \Bigg( CV \Psi^{j_1}  V^{-1} \Bigg) \not\subseteq  \text{ker} \Bigg( CV \Psi^{j_2}  V^{-1} \Bigg)
\end{align}
for all $j_1, j_2$, where $\Psi$ is full column rank and is defined in \ref{eqn:Psi_j}. Then the system is observable with high probability.

\vspace{2mm}

\textit{Remark.} Our loss function to enforce the distinct eigenvalues and kernel condition is using \ref{eqn:Psi_j}
\begin{strip}
\vspace{-4mm}
\par\noindent\rule{\dimexpr(0.5\textwidth-0.5\columnsep-0.4pt)}{0.4pt}%
\rule{0.4pt}{6pt}
\vspace{8mm}
\begin{align}
\label{eqn:loss_thm4}
& \mathcal{L}_{\Xi \in \{ \Phi, \Psi \} } = \ \sum_{j} \text{relu} \Bigg( C - \min_{k_1,k_2} \Big| \Xi_{k_1 k_1}^{j} - \Xi_{k_2 k_2}^{j} \Big| \Bigg)  + \sum_{\substack{j_1 \\ j_2 \neq j_1}} \sum_k \text{relu} \Bigg( C - \Bigg| \Xi_{kk}^{j_1} - \Xi_{kk}^{j_2} \Bigg|^2 \Bigg)  +  \text{relu} \Bigg(  \Bigg| \frac{1}{n} \sum_k e^{i \theta_k(\Lambda) } \Bigg| - C    \Bigg) .
\end{align}
\par
\vspace{2mm}
\vspace{\belowdisplayskip}\hfill\rule[-6pt]{0.4pt}{6.4pt}%
\rule{\dimexpr(0.5\textwidth-0.5\columnsep-1pt)}{0.4pt}
\vspace{-4mm}
\end{strip}%
\noindent\ignorespaces We use $\Xi$ to denote either $\Phi, \Psi$ (the loss is the same for each). If this is satisfied, then with high probability the system is observable.

\vspace{2mm}

\textit{Remark.} The probability of observability is higher the closer $m$ is to $n$. This is because we use a pigeonhole-type argument, which is easier to fulfill the higher $m$ is with respect to $n$.

\vspace{2mm}

\textit{Remark.} The choice of positive constants in the loss can vary between the two terms, and this will impact performance.

\vspace{2mm}

\textit{Remark.} The choice of the Fourier transform is canonical, and is essential to the argument. The Fourier transform produces a canonical family of geometric series evaluations in \ref{app:thm4} of the Krylov sequence $\{C, C\overline{A}, C\overline{A}^2, \dots, C\overline{A}^{L-1}\}$. It is possible to get the evaluation 
\begin{align}
\sum_{k=0}^{L-1} z^k \overline{A}^k = (I - z\overline{A})^{-1} \left(I - z^L \overline{A}^L\right) 
\end{align}
for other $z$. What is special about Fourier is that the $z = \omega_j$ form a complete invertible basis transformation of the entire sequence $I, \overline{A}, \dots, \overline{A}^{L-1}$. In particular, our final step is
\begin{align}
CV\Psi^jV^{-1}\widetilde\psi = \sum_{k=0}^{L-1}\omega_j^k C\overline A^k\widetilde\psi \neq 0.
\end{align}
From this alone, $\exists k \ \text{s.t. } C\overline A^k\widetilde\psi\neq 0$, because if every $C\overline A^k\widetilde\psi$ vanished, every linear combination would vanish (a linear combination of zero vectors is still zero). That direction does not require the Fourier transform to be invertible. Invertibility becomes valuable because we also have the inverse 
\begin{align}
C\overline A^k = \frac{1}{L}\sum_{j=0}^{L-1}\omega_j^{-k}CV\Psi^jV^{-1}.
\end{align}
Therefore,\begin{align}
\bigcap_{j=0}^{L-1}\ker(CV\Psi^jV^{-1}) = \bigcap_{k=0}^{L-1}\ker(C\overline A^k). 
\end{align} 
This is stronger, which says the Fourier transformation neither creates nor destroys an unobservable direction. The concluding remarks in \ref{app:thm4} as to why observability is enforced via the section surround \ref{eqn:cayley} would fail if $\Psi$ was associated arbitrarily.

\vspace{2mm}

\textit{Corollary.} Again, we can relax the condition $A$ is full column rank if we enforce
\begin{align}
\text{det} \Bigg( \mathcal{F} \Big[ \overline{K} \Big]_j \Bigg) > 0
\end{align}
for all $j$, but this is computationally harder.

\begin{figure*}[h]
  \vspace{0mm}
  \centering
  \includegraphics[scale=0.64]{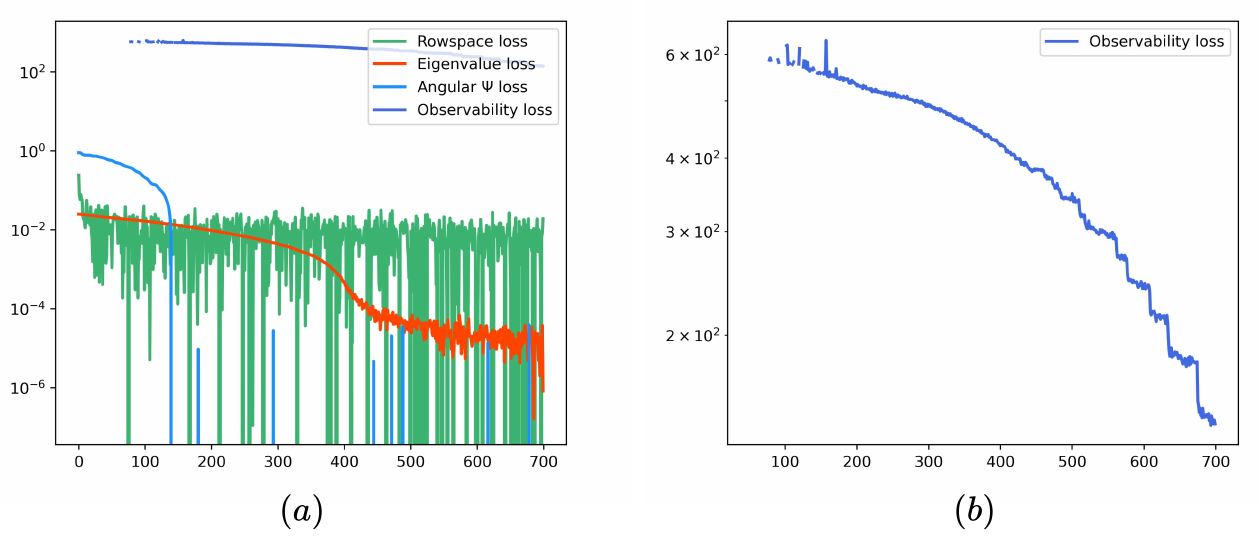}
  \caption{We further illustrate that Theorem 4 is empirically valid. (a) illustrates observability and the Fourier and eigenvalue loss as in equation \ref{eqn:loss_thm4} (b) illustrates observability loss with the observability matrix determinant. We take the log of the determinant, which helps nondegeneracy from the accumulation of small values. If the observability matrix were low rank, the blue loss would be infinite, which we do not have (except near initialization only), so the system is observable. As we can see, loss achieves exactly zero from \ref{eqn:loss_thm4} because we use a relu-type loss.}
  \label{fig:losses_fourier_eigen}
\end{figure*}

\begin{figure*}[h]
  \vspace{0mm}
  \centering
  \includegraphics[scale=0.64]{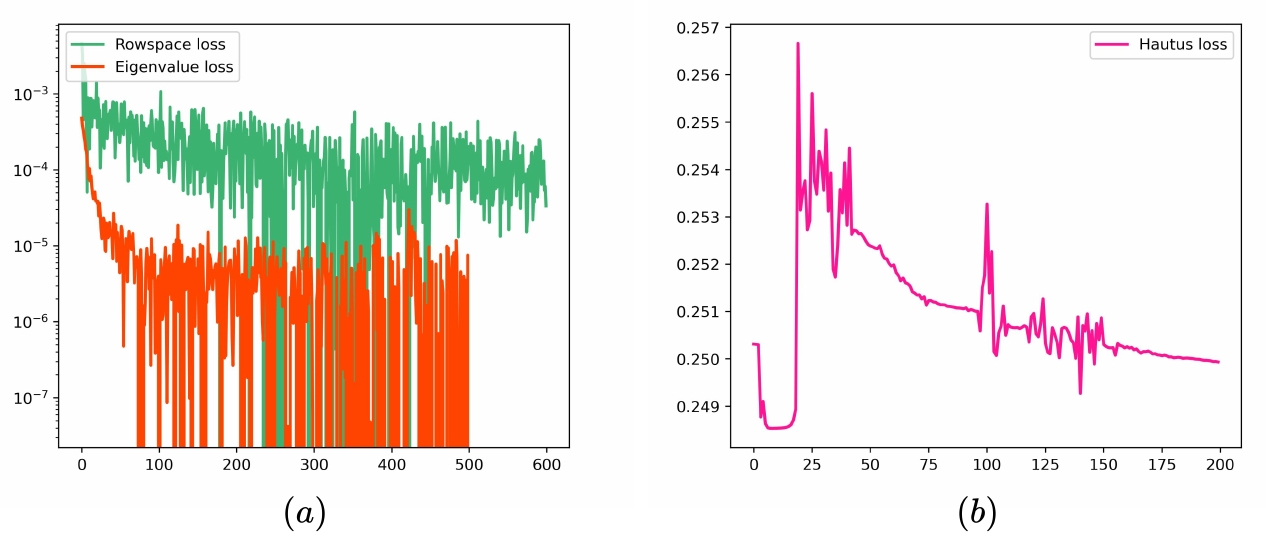}
  \caption{In (a), regarding Theorem 4, as we can see, satisfying the distinct eigenvalues condition does not imply the distinctiveness of $\Psi$ with respect to $j$ is automatically satisfied. Thus, both of these loss terms of equation \ref{eqn:loss_thm4} are valuable in enforcing observability with high probability. In (b), we provide a Hautus loss condition and show that the minimum norm on a column of $CV$ is nonzero regarding Theorem 4 in a very simple state-space model setup with learnable initial hidden state. Again, this eigenvector condition is necessary, not sufficient, as we provided a counterexample in section \ref{sec:hautus}.}

\end{figure*}

\vspace{2mm}

We can note the loss of \ref{eqn:loss_thm4} is $\mathcal{O}(n^3)$, which is faster than \ref{eqn:hautus_loss} by an order of magnitude. Our loss is dominated by the cost of computing the eigenvalues. The minimum is $\mathcal{O}(n^2)$ and this is done across the powers which makes it $\mathcal{O}(n^3)$. Formulating $\Psi$ from the eigenvalues is done just once and is $\mathcal{O}(n)$, and taking powers of $\Psi$ is $\mathcal{O}(n)$.

\vspace{2mm}

Our key arguments for Theorem 3 and Theorem 4 reside in the fact that the exponentiation across the Fourier-analog of the respective matrices, such as $C e^{A k \Delta t} B$, have a more rich exponentiation process than simply exponentiating a diagonalized $A$ via $A^k = V \Lambda^k V^{-1}$. Our argument is centralized on a pigeonhole principle via the kernel such as in \ref{eqn:thm4_kernel_condition}. In regard to the exponentiation process, it can be noted the complex eigenvalues in an imaginary-exponential format scale as $r e^{i \theta} = r^k e^{i k \theta}$ upon exponentiation, which is a highly predictable, uniform process overall. The exponentiation of $\Psi$ is significantly richer and does not cohere to such structure as with $\Lambda, \Lambda^k$. Therefore, our loss is primarily a way to enforce the kernel condition of \ref{eqn:thm4_kernel_condition} by taking advantage of this rich scaling. Our loss is therefore a way to ensure the scaling across $\Psi^j$ utilizes sufficient nondegeneracy across $\Psi$ to ensure this rich exponentiation is fully utilized. $\mathcal{L}$ is overall effectively a nondegeneracy condition, and it is empirically supported.

\vspace{2mm}

Let us elaborate theoretically that each term in \ref{eqn:loss_thm4} has value. Let us denote $W = CV$. Since $\Psi^j$ is diagonal, one matrix layer collection in the observability matrix can be written 

\begin{strip}
\vspace{-8mm}
\par\noindent\rule{\dimexpr(0.5\textwidth-0.5\columnsep-0.4pt)}{0.4pt}%
\rule{0.4pt}{6pt}
\vspace{8mm}
\begin{align}
\label{eqn:Psi_j}
\Psi^j & :=  \text{diag} \Bigg(  \Big(1- e^{-2 \pi i \frac{j}{L}} (\frac{1 + \frac{\Delta \lambda_1}{2}}{1 - \frac{\Delta \lambda_1}{2}}) \Big)^{-1} \Bigg[ 1 - \Big(\frac{1 + \frac{\Delta \lambda_1}{2}}{1 - \frac{\Delta \lambda_1}{2}} \Big)^L \Bigg], \hdots ,  \Big(1- e^{-2 \pi i \frac{j}{L}} (\frac{1 + \frac{\Delta \lambda_n}{2}}{1 - \frac{\Delta \lambda_n}{2}})\Big)^{-1} \Bigg[ 1 - \Big(\frac{1 + \frac{\Delta \lambda_n}{2}}{1 - \frac{\Delta \lambda_n}{2}} \Big)^L \Bigg] \Bigg) .
\end{align}
\par
\vspace{2mm}
\vspace{\belowdisplayskip}\hfill\rule[-6pt]{0.4pt}{6.4pt}%
\rule{\dimexpr(0.5\textwidth-0.5\columnsep-1pt)}{0.4pt}
\vspace{-6mm}
\end{strip}

\begin{align}
W\Psi^j = \begin{pmatrix} w_1 \Psi_{11}^j & w_2 \Psi_{22}^j & \dots & w_n \Psi_{nn}^j \end{pmatrix},
\end{align}
up to a surjective change-of-basis matrix $V^{-1}$ on the right, where $w_i$ is a column of $W$. Therefore, the $k$-th columns of the observability matrix can be written
\begin{align}
\widetilde{o}_k = \begin{pmatrix} w_k \Psi_{kk}^{j_1} \\ w_k \Psi_{kk}^{j_2} \\ \vdots \\ w_k \Psi_{kk}^{j_L} \end{pmatrix} .
\end{align}
To enforce observability, we must have sufficient linear independence up to a tolerance on the $\widetilde{o}$, so we examine the inner product on particular columns
\begin{align}
\label{eqn:observability_column_innerprod}
& \langle \tilde{o}_{k_1}, \tilde{o}_{k_2} \rangle = \sum_{l=1}^L \langle w_{k_1} \Psi_{k_1 k_1}^{j_l}, w_{k_2} \Psi_{k_2 k_2}^{j_l} \rangle_{\mathbb{C}^m} 
\\
& = \langle w_{k_1}, w_{k_2} \rangle \sum_{l=1}^L \overline{\Psi_{k_1 k_1}^{j_l}} \Psi_{k_2 k_2}^{j_l} .
\end{align}
If the first loss term of \ref{eqn:loss_thm4} fails, so there exist $\Psi_{k_1 k_1}^j \approx \Psi_{k_2 k_2}^j$, we get $
\sum_{l=1}^L \overline{\Psi_{k_1 k_1}^{j_l}} \Psi_{k_1 k_1}^{j_l} = \sum_{l=1}^L |\Psi_{k_1 k_1}^{j_l}|^2 $. The above provides zero de-correlation. Let us elaborate more via a linear dependence argument. Suppose $\Psi_{kk}^j \approx \psi^j$ holds for a sufficiently high number of $k$.
\begin{align}
\sum_{k=1}^n c_k \widetilde{o}_k
&= \sum_{k \in \mathcal{K}} c_k
\begin{pmatrix} w_k \psi^{j_1} \\ w_k \psi^{j_2} \\ \vdots \\ w_k \psi^{j_L} \end{pmatrix}
+ \sum_{k \notin \mathcal{K}} c_k
\begin{pmatrix} w_k \Psi_{kk}^{j_1} \\ w_k \Psi_{kk}^{j_2} \\ \vdots \\ w_k \Psi_{kk}^{j_L} \end{pmatrix} .
\end{align}
It is necessarily the case $W=CV$ has columns $w_i$ that are linearly dependent since it is wide. The multipliers $\psi^{j_{\ell}}$ are independent of $k$ and are therefore the same across $k \in \mathcal{K}$ in the first sum. Since $W$ possibly has columns that are linearly dependent in the first sum, $\psi^{j_{\ell}}$ can be absorbed into the constants per $k$ and provides no contribution to ensure linear independence. For example,
\begin{align}
& \sum_{k \in \mathcal{K}} c_k
\begin{pmatrix} w_k \psi^{j_1} \\ w_k \psi^{j_2} \\ \vdots \\ w_k \psi^{j_L} \end{pmatrix} = 
\text{diag} \left( \psi^{j_1} I, \hdots \right) \left( \sum_{k \in \mathcal{K}} c_k
\begin{pmatrix} w_k  \\ w_k  \\ \vdots \\ w_k  \end{pmatrix} \right) .
\end{align}
Now, let us attempt to develop some analysis as to why the second term as in \ref{eqn:loss_thm4} matters. Now let us assume for a subset of $k$, $\mathcal{K}$, we get $\Psi_{kk}^{j} \approx \psi_k$ for all $j \in \mathcal{J}$ for some collection $\mathcal{J}$. Let us examine a subset of the columns of the observability matrix
\begin{align}
\widetilde{o}_k^{\mathcal{J}} = \left. \begin{pmatrix} w_k \Psi_{kk}^{j_{a_1}} \\ w_k \Psi_{kk}^{j_{a_2}} \\ \vdots \end{pmatrix} \right|_{j \in \mathcal{J}} \approx \begin{pmatrix} w_k \psi_k \\ w_k \psi_k \\ \vdots \end{pmatrix} .
\end{align}
As before, since $C$ is wide, we have for certain $\mathcal{K}$,  $\sum_{k \in \mathcal{K}} c_k w_k = 0$. Attempting to examine linear dependence for this subset,
\begin{align}
& \sum_{k \in \mathcal{K}} d_k \widetilde{o}_k^{\mathcal{J}} = \sum_{k \in \mathcal{K}} d_k \begin{pmatrix} w_k \psi_k \\ w_k \psi_k \\ \vdots \end{pmatrix} 
\\
& = \begin{pmatrix} \sum_{k \in \mathcal{K}} d_k \psi_k w_k \\ \sum_{k \in \mathcal{K}} d_k \psi_k w_k \\ \vdots \end{pmatrix} = \begin{pmatrix} \sum_{k \in \mathcal{K}} c_k w_k \\ \sum_{k \in \mathcal{K}} c_k w_k \\ \vdots \end{pmatrix} = 0 
\end{align}
by defining a new set of constants $d_k = c_k / \psi_k$. Therefore, this second loss term helps prevent the above, probabilistically. The third term affects the kernel condition \ref{eqn:loss_thm4} via empirical results, and we refer to Figure \ref{fig:losses_fourier_eigen} for this.

\vspace{2mm}

Let us attempt to illustrate more closely why exponentiating $\Psi^j$ has greater likelihood to enforce the kernel condition of \ref{eqn:thm4_kernel_condition} rather than exponentiating $\Lambda^j$. Let us examine the necessary conditions for nontrivial kernel intersection. Let $D^j$ denote a generalized diagonal matrix. Suppose there exists a non-zero vector $x$ that resides in the kernel for two distinct indices $j_1 \neq j_2$, and so
\begin{align}
x \in \text{ker} \Bigg( CV D^{j_1}  V^{-1} \Bigg) \bigcap  \text{ker} \Bigg( CV D^{j_2}  V^{-1} \Bigg) .
\end{align}
Letting $y = V^{-1}x$ and $W = CV$, by definition of the kernel, we have $W D^{j_1} y = 0$ and $W D^{j_2} y = 0$. Consequently, for any scalar $\alpha \in \mathbb{C}$, the vector $y$ must satisfy 
\begin{align}
\label{eqn:kernel_overlap}
W \Big( D^{j_1} - \alpha D^{j_2} \Big) y = 0 
\end{align}
since each separate term annihilates $y$. If we take $D^j = e^{\Lambda j \Delta}$ (recall our setup as defined in section \ref{sec:ssm} is with a matrix exponential), the diagonal entries of the difference matrix in \ref{eqn:kernel_overlap} become $e^{\lambda_k j_1 \Delta} - \alpha e^{\lambda_k j_2 \Delta}$, since the matrix exponential of a diagonal matrix is the exponential of each element. The relative scaling between modes, given by the ratio $e^{\lambda_k (j_1 - j_2) \Delta}$, changes monotonically in magnitude and linearly in phase. In particular, we decompose $\lambda_k = \sigma_k + i \omega_k$ and denote $\tau = (j_1 - j_2)\Delta$. We can write
\begin{align}
\label{eqn:exp}
e^{\lambda_k \tau} = e^{\sigma_k \tau} e^{i \omega_k \tau} .
\end{align}
We observe from this
\begin{itemize}
\item Monotonic magnitude: the modulus $|e^{\lambda_k \tau}| = e^{\sigma_k \tau}$ strictly grows or decays. It never oscillates or exhibits local extrema.
\item Linear phase: the argument $\arg(e^{\lambda_k \tau}) = \omega_k \tau \pmod{2\pi}$ winds around the origin at a constant angular velocity.
\end{itemize}
Let us show this argument fails for $\Psi^j$. Let us concatenate the terms constant in $j$ in $\Psi$ into a constant and so rewrite
\begin{align}
\Psi_{kk}^j = \frac{C_k}{1 - \gamma_k e^{-2 \pi i \frac{j}{L}}} .
\end{align}
As before, examining a ratio,
\begin{align}
\frac{\Psi_{kk}^{j_1}}{\Psi_{kk}^{j_2}} = \frac{1 - \gamma_k e^{-2 \pi i \frac{j_2}{L}}}{1 - \gamma_k e^{-2 \pi i \frac{j_1}{L}}}  .
\end{align}
Express $\gamma_k = |\gamma_k| e^{i \phi_k}$, and returning to equation \ref{eqn:observability_column_innerprod}, notice
\begin{align}
\Bigg| \frac{\Psi_{kk}^{j_1}}{\Psi_{kk}^{j_2}} \Bigg|^2 = \frac{1 + |\gamma_k|^2 - 2|\gamma_k| \cos\left(2\pi \frac{j_2}{L} - \phi_k \right)}{1 + |\gamma_k|^2 - 2|\gamma_k| \cos\left(2\pi \frac{j_1}{L} - \phi_k \right)} 
\end{align}
for suitable intrinsic phase $\phi$. The magnitude fluctuates based on a cosine, and experiences local resonances/extrema, and is widely mercurial in behavior. Moreover, we can note the phase
\begin{align}
\arg \Bigg( \frac{\Psi_{kk}^{j_1}}{\Psi_{kk}^{j_2}} \Bigg) & = \arg\left(1 - \gamma_k e^{-2 \pi i \frac{j_2}{L}}\right)
\\
& \ \ \ \ \ \ \ - \arg\left(1 - \gamma_k e^{-2 \pi i \frac{j_1}{L}}\right) 
\end{align}
does not wind at constant angular velocity.

\vspace{2mm}

We make some concluding remarks on this theory. It is still necessary $Cv \neq 0$ is valid for the result to hold, which is the Hautus lemma. Instead, the eigenvalues act more of a guide to help the permutation structure of the observability matrix. For example, the observability matrix $(CV^T \ CV^T \ CV^T \ \hdots)^T$ still cannot be full column rank because $C$ is wide, where we have diagonalized $A$ and the eigenvalue matrix is the identity. Thus, eigenvalues work at the interplay of these systems, and the high probability necessity holds.

\subsection{An adaptation of the Hautus lemma: Mamba observability with exponential diagonalization and a Vandermonde matrix}

\begin{figure*}[h]
  \vspace{0mm}
  \centering
  \includegraphics[scale=0.5]{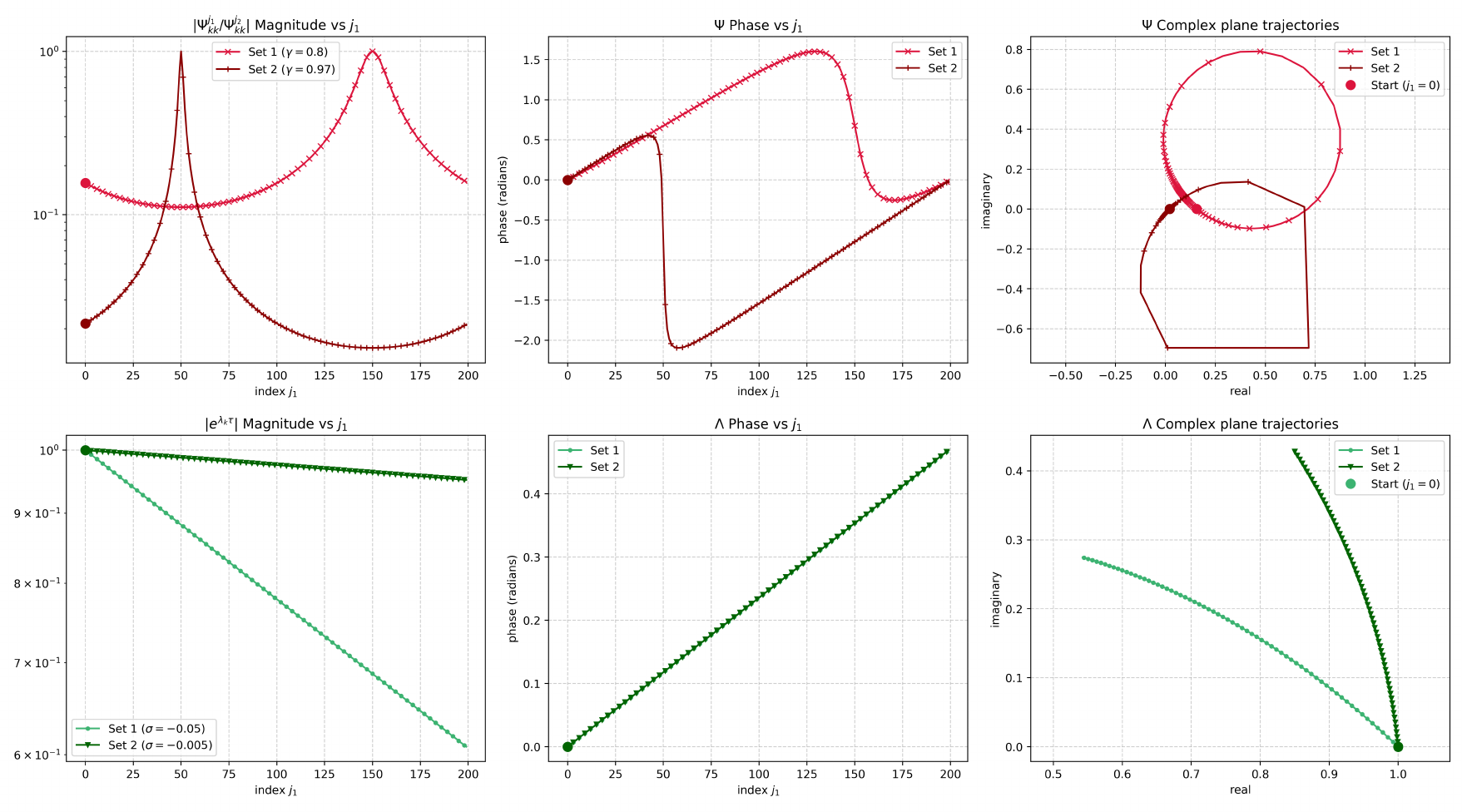}
  \caption{We illustrate intricate scaling using two specific typical samples of $\Psi$ versus using two typical samples of $\Lambda$ directly. We use (1) $j_2=0, \Delta=0.05,\sigma_k=-0.05,\omega_k=3\pi/200,\gamma_k = 0.8 \exp( 3 \pi i /2)$ and (2) $j_2=0, \Delta=0.05,\sigma_k=-0.005,\omega_k=3\pi/200,\gamma_k = 0.97 \exp(  \pi i /2)$, both with normalization. The left panel show the magnitude (in the sense of $\sqrt{\text{Re}(z)^2 + \text{Im}(z)^2}$) of the ratios of $e^{\lambda_k \tau}$ in \ref{eqn:exp} and between $ \Psi_{kk}^{j_1} / \Psi_{kk}^{j_2}$ for fixed $j_2$ across varying $j_1$. The middle panel shows $\text{arg}(z) = \text{arg}(re^{i \theta}) = \theta$ for $e^{\lambda_k \tau},  \Psi_{kk}^{j_1} / \Psi_{kk}^{j_2}$. The right panel shows $e^{\lambda_k \tau},  \Psi_{kk}^{j_1} / \Psi_{kk}^{j_2}$ in the complex plane varying across $j_1$, where $j_2$ is fixed.}
  \label{fig:ratio}
\end{figure*}

A consequence of the Hautus lemma is that the if the matrix $C$ does not annihilate every eigenvector of $A$, then the system is observable. As we saw in \ref{sec:hautus}, by enforcing an eigenvector condition directly, we have a necessary but not sufficient condition. We also saw in \ref{sec:hautus} that we can enforce the Hautus lemma by finding the eigenvalues and taking a determinant, but a determinant is computationally nontrivial.

\vspace{2mm}

Let us explain some key facts of our theorem here. Our theorem here implicitly tests the Hautus lemma because we will examine $\widetilde{C}=CV$, which uses the eigenvectors. Thus, we must compute eigenvectors to enforce observability here, but this is not as nontrivial as computing a determinant, partially because determinants are sensitive and can experience blow-up effects. Both eigenvectors and determinants are $\mathcal{O}(n^3)$, but the Hautus lemma loss in \ref{sec:hautus} is $\mathcal{O}(n^4)$ because there are $n$ eigenvectors and we must enforce the loss for each. One primary advantage of this theorem is that there is no need to exponentiate matrix $A$, therefore mitigating effects of catastrophe of computing the determinant of the observability matrix. As mentioned, a second quality of this theorem is it that it effectively reproves the Hautus lemma. Our result relies on a diagonalization-type argument of $A$ and taking the matrix exponential, using Vandermonde matrices and Kronecker products along the way. We will take select rows and enforce a full rank condition on the result, which forces the entire observability matrix to be full column rank.

\vspace{2mm}

\textbf{Theorem 5 (modification of the Hautus-Lemma eigenvector test without testing eigenvectors).} Suppose $A \in \mathbb{R}^{n \times n}$ takes the form
\begin{equation}
A = V \text{diag} \Big( \lambda_1, \lambda_2, \hdots, \lambda_n \Big) V^{-1}.
\end{equation}
Here $A$ is diagonalized, $V$ is surjective. Suppose that $C \in \mathbb{R}^{m \times n}$. Suppose the loss
\begin{align}
\mathcal{L} = & \ \text{relu} \Bigg( \text{constant}  - \min_{k_1,k_2} \Bigg| \lambda_{k_1}  - \lambda_{k_2} \Bigg| \Bigg)  
\\
& + \sum_j \text{relu} \Bigg( \text{constant}  - | \widetilde{C}_{1j} | \Bigg) 
\end{align}
is exactly satisfied, where $\widetilde{C} = CV$. Then the pair $(C,\overline{A})$ is observable. Moreover, the pair $(C,A)$ is observable.

\vspace{2mm}

Alternatively, we can enforce observability over more rows by taking
\begin{align}
\mathcal{L} = & \ \underbrace{ \text{relu} \Bigg( \text{constant}  - \min_{k_1,k_2} \Bigg| \lambda_{k_1}  - \lambda_{k_2} \Bigg| \Bigg) }_{\text{distinct Vandermonde column condition}}
 \\
 & + \underbrace{ \sum_i \sum_j \text{relu} \Bigg( \text{constant}  - | \widetilde{C}_{ij} | \Bigg) }_{\text{non-annihilation of eigenvectors in} \ C}.
\end{align}

\begin{figure*}[h]
  \vspace{0mm}
  \centering
  \includegraphics[scale=0.68]{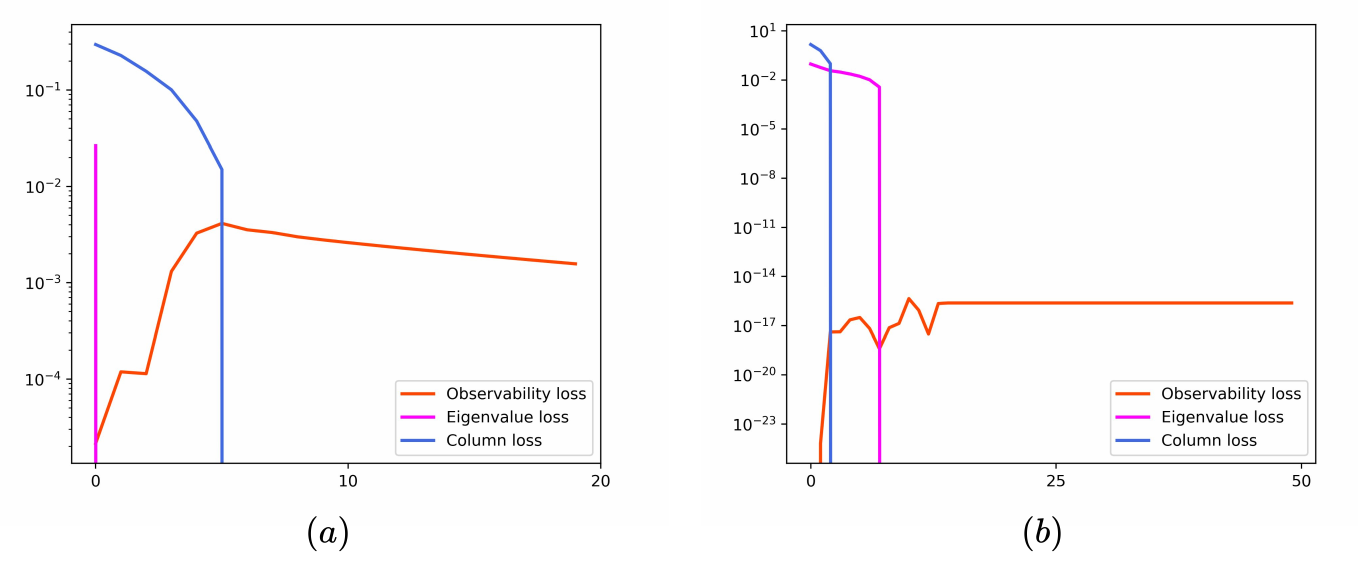}
  \caption{This figure helps demonstrate that the Hautus lemma eigenvector test indeed yields an observable matrix. This figure is based off the continuation of Theorem 5 in its relation to the Hautus test: this is why we include a distinct eigenvalue condition. In essence, the zero values of eigenvalue and column loss are desirable, and the fact that the observability loss has plateaued and is not decreasing significantly if at all is desirable. We depict objective terms (with (x) training iteration mod 20 versus (y) log loss) regarding this theorem to enforce observability: the first is $|\text{det}(\mathcal{O}^T \mathcal{O})|$; the second is $\text{relu}(\text{const} - \min | \lambda_i - \lambda_j|)$; and the third is $\text{relu}(\text{const} - ||[CV]_i||_2)$. In (a), we normalize the determinant with $\text{det}(\Phi) / \prod_i || [\Phi]_i ||_2$ to better illustrate nondegeneracy. Here, $m=2,n=3$. In (b), the determinant is not normalized, and $m=10,n=15$. As we can see, the determinant increases with respect to training iteration here, and eventually stabilizes. The determinant is quite small because the entries are small, not due to imprecision with a low rank matrix. In the (a) case, we examined $\mathcal{O}$ specifically and saw it was full column rank, although this is hard to do in the (b) case because it is much more high-dimensional.}
  \label{fig:losses_obs}
\end{figure*}

\subsection{Parameter sharing and algorithmic convergence}

In this section, we construct a coupling between $A$ and $B$. This does two things for us. The primary thing this done for is that it creates a drastic parameter reduction since we can know construct our high-dimensional $A$ from lower dimensional $B$ while parameterizing both simultaneously. Also, we enforce specific structure on $A$ and $B$ that makes exponentiating $A$ computationally efficient. Moreover, since our general form of $B$ is already diagonalized, our extraction of $B$ is efficient by introducing a supplemental matrix to extract the appropriate eigenvectors to construct $B$ non-square out of square with an eigendecomposition.

\vspace{2mm}

Let $n = pm$ for $p \in \mathbb{N}$. Consider $A =  (Q \otimes I_p)^T ( U \otimes I_p)^T \text{diag}( \{\lambda_i\}_i) (U \otimes I_p) (Q \otimes I_p), B = \text{Eigendecomp}( (U \otimes I)^T f( \text{diag}(\{\lambda_i\}_i)) ( U \otimes I) ) $. Here, $Q,U \in \mathbb{R}^{m \times m}$ are symmetric. Now, note that $B$ is taking an eigendecomposition of a matrix already in a diagonalized form. Thus, we can avoid a more computationally demanding eigendecomposition and simply formulate $B$ by taking
\begin{equation}
B = f(\text{diag}(\lambda_1,\hdots,\lambda_n))^{1/2}(U \otimes I) S ,
\end{equation}
where $S \in \mathbb{R}^{n \times m}$ is learned, and satisfies
\begin{equation}
S_{kj} = \begin{cases}
    1 \ \text{if} \ k = i_j
    \\
    0 \ \text{otherwise}
\end{cases} ,
\end{equation}
where $i_j$ corresponds to the $i_j-th$ column to be selected in the new position $j$. We also require
\begin{align}
( U \otimes I)^{T} = (U \otimes I)^{-1}, \ \ \ \ \ (Q \otimes I)^T = (Q \otimes I)^{-1} .
\end{align}
Thus, exponentiation of $A$ is fast and we compute
\begin{equation}
A^k = (Q \otimes I)^T (U \otimes I)^T \text{diag}(\lambda_1^k,\hdots,\lambda_n^k) (U \otimes I) (Q \otimes I ).
\end{equation}

\subsection{Algorithmic convergence and training}

\textbf{Theorem 6.} Suppose loss $f$ decays at a rate
\begin{align}
& \EX \Big[ || (U_{k+1},Q_{k+1},S_{k+1}) - (U_{k},Q_{k},S_{k}) || \Big] \\
= \ & \EX \Big[ || \nabla f(U_k,Q_k,S_k) || \Big] \sim \Theta ( \frac{1}{k^q} ) ,
\end{align} 
where $\frac{1}{2} \leq q \leq 1$. We refer to \cite{ward2021adagradstepsizessharpconvergence} \cite{arjevani2022lowerboundsnonconvexstochastic} \cite{carmon2017lowerboundsfindingstationary} \cite{carmon2019lowerboundsfindingstationary} \cite{convergence_rate_nonconvex} \cite{bubeck2015convexoptimizationalgorithmscomplexity} \cite{ghadimi2013stochasticfirstzerothordermethods} \cite{lei2019stochasticgradientdescentnonconvex} for relevant literature. Define
\begin{align}
& \mathcal{A}(Q,U,\Lambda) = (Q \otimes I_p)^T ( U \otimes I_p)^T \\
& \quad \quad \quad \quad \text{diag}(\lambda_1,\hdots,\lambda_n) (U \otimes I_p) (Q \otimes I_p)   
\\
& \mathcal{B}(\Lambda,U,S) = f(\text{diag}(\lambda_1,\hdots,\lambda_n))^{1/2}(U \otimes I) S .
\end{align}
Here, $p$ is some integer. Consider the training algorithm
\begin{align}
& A_{k+1} \leftarrow A_k  + || \mathcal{A}(Q_k,U_k) 
\\
& \quad \quad - \mathcal{A}(Q_{k-1},U_{k-1}) ||_F^{1/2q} \cdot \mathcal{A}(Q_k,U_k,\Lambda_k)
\\
& B_{k+1} \leftarrow B_k + || \mathcal{B}(U_k,S_k) 
\\
& \quad \quad - \mathcal{B}(U_{k-1},S_{k-1}) ||_F^{1/q}  \cdot \mathcal{B}(\Lambda_k,U_k,S_k)
\\
& \text{Update} \ Q_{k+1},U_{k+1},S_{k+1} .
\end{align}
Suppose $Q,U$ are orthogonal and take the diagonalized form $Q = V_Q D_Q V_Q^T, U_i = V_U D_U V_U$ (we remark the entries can be complex-valued) where $D_Q, D_U$ are diagonal and learnable. Suppose that orthogonal $V$ are either fixed or updated only sometimes and finitely often. Then the algorithm satisfies a Robbins-Monro condition.

\vspace{2mm}

\textit{Remark.} To enforce the conditions on $Q,U$, we can construct them as $V D (D^T D)^{-1/2} V^T$, where $D$ is diagonal and $V$ is any orthogonal matrix. We can construct $V$ as $P (P^T P)^{-1/2}$, which is any orthogonal matrix.

\vspace{2mm}

\textbf{Theorem 7.} Suppose the hypotheses of Lemma 9 are satisfied. Let $\mathcal{A}, \mathcal{B}$ be as in Theorem 6. Let $Q,U \in \mathbb{R}^{m \times m}$ be any matrices. Then the algorithm
\begin{align}
& A_{k+1} \leftarrow \mathcal{A}(Q_k,U_k,\Lambda_k)
\\
& B_{k+1} \leftarrow \mathcal{B}(\Lambda_k,U_k,S_k)
\\
& \text{Update} \ Q_{k+1},U_{k+1},S_{k+1}  .
\end{align}
does not satisfy a contraction mapping with Lipschitz constant $L \gg 1$.

\section{Conclusions and limitations}

We presented a series of results, ranging in complexity and rigor, on how to enforce observability in neural contexts. Our results demonstrate improvements in computational gains in order to instill observability. Our results have focused on traditional approaches, permutations, full rank conditions in Fourier space, modifications of Hautus-type conditions, and couplings to reduce learning complexity over the parameter space.

\vspace{2mm}

The primary drawback of our methods is that they are restrictive over the parameter space, sometimes by a lot but sometimes not by much. While observability in itself is inherently a restriction on the state-space matrices, our methods extend this further and diminish parameter-learning autonomy even moreso than may be necessary to enforce observability at its more essential level. For example, constructing $A$ to necessarily be a permutation matrix means that this matrix is sparse, and this is fundamentally our most restrictive condition out of all we propose. Our research provides insight towards efficiency in control theory learning at such a cost of expressivity.

\section{Acknowledgements}

Andrew Gracyk is currently affiliated with Purdue University Department of Mathematics. I acknowledge and am grateful for previous financial support from University of Illinois Urbana-Champaign (UIUC) Department of Statistics and Data and Informatics Graduate Intern-Traineeship: Materials at the Atomic Scale (DIGIMAT) with Grant No. 1922758 at UIUC. I would like to thank Hanming Zhou at University of California, Santa Barbara, whose course on control theory and ordinary differential equations largely laid the foundational basis for this work.

\bibliographystyle{plainnat}
\bibliography{bibliography}

@misc{evans_control_course,
  author       = {Lawrence C. Evans},
  title        = {An Introduction to Mathematical
  Optimal Control Theory
  },
  year         = {2024},
  url = {https://math.berkeley.edu/~evans/control.course.pdf}
}

@article{cheng2024mambaneuraloperatorwins,
title = {Mamba neural operator: Who wins? transformers vs. state-space models for PDEs},
journal = {Journal of Computational Physics},
volume = {548},
pages = {114567},
year = {2026},
issn = {0021-9991},
doi = {https://doi.org/10.1016/j.jcp.2025.114567},
url = {https://www.sciencedirect.com/science/article/pii/S0021999125008496},
author = {Chun-Wun Cheng and Jiahao Huang and Yi Zhang and Guang Yang and Carola-Bibiane Schönlieb and Angelica I. Aviles-Rivero}
}

@article{hu2024statespacemodelsaccurateefficient,
title = {State-space models are accurate and efficient neural operators for dynamical systems},
journal = {Neural Networks},
volume = {197},
pages = {108496},
year = {2026},
issn = {0893-6080},
doi = {https://doi.org/10.1016/j.neunet.2025.108496},
url = {https://www.sciencedirect.com/science/article/pii/S0893608025013772},
author = {Zheyuan Hu and Nazanin Ahmadi Daryakenari and Qianli Shen and Kenji Kawaguchi and George Em Karniadakis}
}

@inproceedings{
gu2024mambalineartimesequencemodeling,
title={Mamba: Linear-Time Sequence Modeling with Selective State Spaces},
author={Albert Gu and Tri Dao},
booktitle={First Conference on Language Modeling},
year={2024},
url={https://openreview.net/forum?id=tEYskw1VY2}
}

@Article{signals5030023,
AUTHOR = {Zhou, Qihou},
TITLE = {On the Impulse Response of Singular Discrete LTI Systems and Three Fourier Transform Pairs},
JOURNAL = {Signals},
VOLUME = {5},
YEAR = {2024},
NUMBER = {3},
PAGES = {460--473},
URL = {https://www.mdpi.com/2624-6120/5/3/23},
ISSN = {2624-6120},
DOI = {10.3390/signals5030023}
}

@article{Wang2001TechnicalCI,
  title={Technical Communique: Impulse observability and impulse controllability of linear time-varying singular systems},
  author={Chi-Jo Wang and Ho-En Liao},
  journal={Automatica},
  year={2001},
  volume={37},
  pages={1867-1872},
  url={https://api.semanticscholar.org/CorpusID:9880960}
}

@article{SUBASI20141669,
title = {Quantitative measure of observability for linear stochastic systems},
journal = {Automatica},
volume = {50},
number = {6},
pages = {1669-1674},
year = {2014},
issn = {0005-1098},
doi = {https://doi.org/10.1016/j.automatica.2014.04.008},
url = {https://www.sciencedirect.com/science/article/pii/S0005109814001381},
author = {Yuksel Subasi and Mubeccel Demirekler}
}

@misc{hamdan2024sparsemambaintroducingcontrollability,
      title={Sparse Mamba: Introducing Controllability, Observability, And Stability To Structural State Space Models}, 
      author={Emadeldeen Hamdan and Hongyi Pan and Ahmet Enis Cetin},
      year={2024},
      eprint={2409.00563},
      archivePrefix={arXiv},
      primaryClass={cs.LG},
      url={https://arxiv.org/abs/2409.00563}, 
}

@misc{gu2022efficientlymodelinglongsequences,
      title={Efficiently Modeling Long Sequences with Structured State Spaces}, 
      author={Albert Gu and Karan Goel and Christopher Ré},
      year={2022},
      eprint={2111.00396},
      archivePrefix={arXiv},
      primaryClass={cs.LG},
      url={https://arxiv.org/abs/2111.00396}, 
}

@misc{alonso2024statespacemodelsfoundation,
      title={State Space Models as Foundation Models: A Control Theoretic Overview}, 
      author={Carmen Amo Alonso and Jerome Sieber and Melanie N. Zeilinger},
      year={2024},
      eprint={2403.16899},
      archivePrefix={arXiv},
      primaryClass={eess.SY},
      url={https://arxiv.org/abs/2403.16899}, 
}

@inproceedings{forgione2023empirical,
title={{Neural State-Space Models: Empirical Evaluation
of Uncertainty Quantification}},
author={Forgione, Marco and Piga, Dario},
booktitle={Proc. of the 22nd IFAC World Congress, Yokohama, Japan},
year={2023}
}

@inproceedings{
smith2023simplifiedstatespacelayers,
title={Simplified State Space Layers for Sequence Modeling},
author={Jimmy T.H. Smith and Andrew Warrington and Scott Linderman},
booktitle={The Eleventh International Conference on Learning Representations },
year={2023},
url={https://openreview.net/forum?id=Ai8Hw3AXqks}
}

@inproceedings{
jafari2025mambalrpexplainingselectivestate,
title={Mamba{LRP}: Explaining Selective State Space Sequence Models},
author={Farnoush Rezaei Jafari and Gr{\'e}goire Montavon and Klaus Robert Muller and Oliver Eberle},
booktitle={The Thirty-eighth Annual Conference on Neural Information Processing Systems},
year={2024},
url={https://openreview.net/forum?id=2n1Ysn1EDl}
}

@misc{anthony2024blackmambamixtureexpertsstatespace,
      title={BlackMamba: Mixture of Experts for State-Space Models}, 
      author={Quentin Anthony and Yury Tokpanov and Paolo Glorioso and Beren Millidge},
      year={2024},
      eprint={2402.01771},
      archivePrefix={arXiv},
      primaryClass={cs.CL},
      url={https://arxiv.org/abs/2402.01771}, 
}

@InProceedings{ward2021adagradstepsizessharpconvergence,
  title = 	 {{A}da{G}rad Stepsizes: Sharp Convergence Over Nonconvex Landscapes},
  author =       {Ward, Rachel and Wu, Xiaoxia and Bottou, Leon},
  booktitle = 	 {Proceedings of the 36th International Conference on Machine Learning},
  pages = 	 {6677--6686},
  year = 	 {2019},
  editor = 	 {Chaudhuri, Kamalika and Salakhutdinov, Ruslan},
  volume = 	 {97},
  series = 	 {Proceedings of Machine Learning Research},
  month = 	 {09--15 Jun},
  publisher =    {PMLR},
  url = 	 {https://proceedings.mlr.press/v97/ward19a.html}
}

@article{arjevani2022lowerboundsnonconvexstochastic,
  author    = {Arjevani, Yossi and Carmon, Yair and Duchi, John C. and Foster, Dylan J. and Srebro, Nathan and Woodworth, Blake},
  title     = {Lower bounds for non-convex stochastic optimization},
  journal   = {Mathematical Programming},
  year      = {2023},
  volume    = {199},
  number    = {1},
  pages     = {165--214},
  doi       = {10.1007/s10107-022-01822-7},
  url       = {https://doi.org/10.1007/s10107-022-01822-7}
}

@misc{pickard2025dynamicsensorselectionbiomarker,
      title={Dynamic Sensor Selection for Biomarker Discovery}, 
      author={Joshua Pickard and Cooper Stansbury and Amit Surana and Lindsey Muir and Anthony Bloch and Indika Rajapakse},
      year={2025},
      eprint={2405.09809},
      archivePrefix={arXiv},
      primaryClass={q-bio.MN},
      url={https://arxiv.org/abs/2405.09809}, 
}

@article{Montanari_2022,
   title={Functional observability and subspace reconstruction in nonlinear systems},
   volume={4},
   ISSN={2643-1564},
   url={http://dx.doi.org/10.1103/PhysRevResearch.4.043195},
   DOI={10.1103/physrevresearch.4.043195},
   number={4},
   journal={Physical Review Research},
   publisher={American Physical Society (APS)},
   author={Montanari, Arthur N. and Freitas, Leandro and Proverbio, Daniele and Gonçalves, Jorge},
   year={2022},
   month=dec }

@misc{karamched2024observabilitycomplexsystemsconserved,
      title={Observability of complex systems via conserved quantities}, 
      author={Bhargav Karamched and Jack Schmidt and David Murrugarra},
      year={2024},
      eprint={2408.00143},
      archivePrefix={arXiv},
      primaryClass={math.DS},
      url={https://arxiv.org/abs/2408.00143}, 
}

@article{Montanari_2021,
   title={Functional observability and target state estimation in large-scale networks},
   volume={119},
   ISSN={1091-6490},
   url={http://dx.doi.org/10.1073/pnas.2113750119},
   DOI={10.1073/pnas.2113750119},
   number={1},
   journal={Proceedings of the National Academy of Sciences},
   publisher={Proceedings of the National Academy of Sciences},
   author={Montanari, Arthur N. and Duan, Chao and Aguirre, Luis A. and Motter, Adilson E.},
   year={2021},
   month=dec }

@article{
doi:10.1073/pnas.1215508110,
author = {Yang-Yu Liu  and Jean-Jacques Slotine  and Albert-László Barabási },
title = {Observability of complex systems},
journal = {Proceedings of the National Academy of Sciences},
volume = {110},
number = {7},
pages = {2460-2465},
year = {2013},
doi = {10.1073/pnas.1215508110},
URL = {https://www.pnas.org/doi/abs/10.1073/pnas.1215508110},
eprint = {https://www.pnas.org/doi/pdf/10.1073/pnas.1215508110}}

@article{Kunapareddy2018RecoveringOV,
  title={Recovering Observability via Active Sensing},
  author={Abhinav Kunapareddy and Noah J. Cowan},
  journal={2018 Annual American Control Conference (ACC)},
  year={2018},
  pages={2821-2826},
  url={https://api.semanticscholar.org/CorpusID:2720673}
}

@article{PhysRevE.86.026205,
  title = {Influence of the singular manifold of nonobservable states in reconstructing chaotic attractors},
  author = {Frunzete, Madalin and Barbot, Jean-Pierre and Letellier, Christophe},
  journal = {Phys. Rev. E},
  volume = {86},
  issue = {2},
  pages = {026205},
  numpages = {8},
  year = {2012},
  month = {Aug},
  publisher = {American Physical Society},
  doi = {10.1103/PhysRevE.86.026205},
  url = {https://link.aps.org/doi/10.1103/PhysRevE.86.026205}
}

@inproceedings{Rafieisakhaei_2017,
   title={On the use of the observability gramian for partially observed robotic path planning problems},
   url={http://dx.doi.org/10.1109/CDC.2017.8263868},
   DOI={10.1109/cdc.2017.8263868},
   booktitle={2017 IEEE 56th Annual Conference on Decision and Control (CDC)},
   publisher={IEEE},
   author={Rafieisakhaei, Mohammadhussein and Chakravorty, Suman and Kumar, P. R.},
   year={2017},
   month=dec, pages={1523–1528} }

@article{Vaidya2007ObservabilityGF,
  title={Observability gramian for nonlinear systems},
  author={Umesh Vaidya},
  journal={2007 46th IEEE Conference on Decision and Control},
  year={2007},
  pages={3357-3362},
  url={https://api.semanticscholar.org/CorpusID:2275515}
}

@misc{dahleh2011lectures,
  title = {Lectures on Dynamic Systems and Control},
  author = {Dahleh, Mohammed and Dahleh, Munther A. and Verghese, George},
  year = {2011},
  institution = {Massachusetts Institute of Technology},
  url = {https://ocw.mit.edu/courses/6-241j-dynamic-systems-and-control-spring-2011/2f03f88e1a714f3ccdb0b7b3f05a2c55_MIT6_241JS11_chap24.pdf}
}

@article{carmon2017lowerboundsfindingstationary,
  author    = {Carmon, Yair and Duchi, John C. and Hinder, Oliver and Sidford, Aaron},
  title     = {Lower bounds for finding stationary points II: first-order methods},
  journal   = {Mathematical Programming},
  year      = {2021},
  volume    = {185},
  number    = {1},
  pages     = {315--355},
  doi       = {10.1007/s10107-019-01431-x},
  url       = {https://doi.org/10.1007/s10107-019-01431-x}
}

@misc{carmon2019lowerboundsfindingstationary,
      title={Lower Bounds for Finding Stationary Points I}, 
      author={Yair Carmon and John C. Duchi and Oliver Hinder and Aaron Sidford},
      year={2019},
      eprint={1710.11606},
      archivePrefix={arXiv},
      primaryClass={math.OC},
      url={https://arxiv.org/abs/1710.11606}, 
}

@misc{convergence_rate_nonconvex, 
  title        = {The proof for worst-case convergence rate of non-smooth convex optimization},
  author = {Leontina},
  note         = {StackExchange},
}

@article{bubeck2015convexoptimizationalgorithmscomplexity,
author = {Bubeck, S{\'e}bastien},
title = {Convex Optimization: Algorithms and Complexity},
year = {2015},
issue_date = {Nov 2015},
publisher = {Now Publishers Inc.},
address = {Hanover, MA, USA},
volume = {8},
number = {3–4},
issn = {1935-8237},
url = {https://doi.org/10.1561/2200000050},
doi = {10.1561/2200000050},
journal = {Found. Trends Mach. Learn.},
month = nov,
pages = {231–357},
numpages = {130}
}

@article{ghadimi2013stochasticfirstzerothordermethods,
  author    = {Ghadimi, Saeed and Lan, Guanghui},
  title     = {Stochastic First- and Zeroth-Order Methods for Nonconvex Stochastic Programming},
  journal   = {SIAM Journal on Optimization},
  year      = {2013},
  volume    = {23},
  number    = {4},
  pages     = {2341--2368},
  doi       = {10.1137/120880811},
  url       = {https://doi.org/10.1137/120880811}
}

@misc{lei2019stochasticgradientdescentnonconvex,
      title={Stochastic Gradient Descent for Nonconvex Learning without Bounded Gradient Assumptions}, 
      author={Yunwen Lei and Ting Hu and Guiying Li and Ke Tang},
      year={2019},
      eprint={1902.00908},
      archivePrefix={arXiv},
      primaryClass={cs.LG},
      url={https://arxiv.org/abs/1902.00908}, 
}

@misc{Chafai2011,
  author       = {Djalil Chafai},
  title        = {The Hoffman-Wielandt Inequality},
  url          = {https://djalil.chafai.net/blog/2011/12/03/the-hoffman-wielandt-inequality/},
}

@misc{eigenvector_deriv, 
  title        = {Derivative of eigenvectors of a matrix with respect to its components},
  author = {Jeff Tehrani},
  note         = {StackExchange}
}

@InProceedings{ansari2023neuralcontinuousdiscretestatespace,
  title = 	 {Neural Continuous-Discrete State Space Models for Irregularly-Sampled Time Series},
  author =       {Ansari, Abdul Fatir and Heng, Alvin and Lim, Andre and Soh, Harold},
  booktitle = 	 {Proceedings of the 40th International Conference on Machine Learning},
  pages = 	 {926--951},
  year = 	 {2023},
  editor = 	 {Krause, Andreas and Brunskill, Emma and Cho, Kyunghyun and Engelhardt, Barbara and Sabato, Sivan and Scarlett, Jonathan},
  volume = 	 {202},
  series = 	 {Proceedings of Machine Learning Research},
  month = 	 {23--29 Jul},
  publisher =    {PMLR},
  url = 	 {https://proceedings.mlr.press/v202/ansari23a.html}
}

@misc{toloubidokhti2022neuralstatespacemodelinglatent,
      title={Neural State-Space Modeling with Latent Causal-Effect Disentanglement}, 
      author={Maryam Toloubidokhti and Ryan Missel and Xiajun Jiang and Niels Otani and Linwei Wang},
      year={2022},
      eprint={2209.12387},
      archivePrefix={arXiv},
      primaryClass={cs.LG},
      url={https://arxiv.org/abs/2209.12387}, 
}

@misc{bae2025causalmambascalableconditionalstate,
      title={CausalMamba: Scalable Conditional State Space Models for Neural Causal Inference}, 
      author={Sangyoon Bae and Jiook Cha},
      year={2025},
      eprint={2510.17318},
      archivePrefix={arXiv},
      primaryClass={cs.CV},
      url={https://arxiv.org/abs/2510.17318}, 
}

@misc{lahoti2026mamba3improvedsequencemodeling,
      title={Mamba-3: Improved Sequence Modeling using State Space Principles}, 
      author={Aakash Lahoti and Kevin Y. Li and Berlin Chen and Caitlin Wang and Aviv Bick and J. Zico Kolter and Tri Dao and Albert Gu},
      year={2026},
      eprint={2603.15569},
      archivePrefix={arXiv},
      primaryClass={cs.LG},
      url={https://arxiv.org/abs/2603.15569}, 
}

@misc{ruder2026diffusiondrivenstatespacemodels,
      title={Diffusion-Driven State Space Models}, 
      author={Jack Ruder and Michael Wojnowicz},
      year={2026},
      eprint={2606.21036},
      archivePrefix={arXiv},
      primaryClass={stat.ML},
      url={https://arxiv.org/abs/2606.21036}, 
}

@misc{nguyen2026muonssmorthogonalizingstatespace,
      title={MuonSSM: Orthogonalizing State Space Models for Sequence Modeling}, 
      author={Thai-Khanh Nguyen and Ngoc-Bich-Uyen Vo and Thieu N. Vo and Tan M. Nguyen and Cuong Pham},
      year={2026},
      eprint={2606.30461},
      archivePrefix={arXiv},
      primaryClass={cs.LG},
      url={https://arxiv.org/abs/2606.30461}, 
}

@article{gilyen2025arbeitsgemeinschaft,
  author  = {Gily{\'e}n, Andr{\'a}s and Lin, Lin and Thiele, Christoph},
  title   = {Arbeitsgemeinschaft: Quantum Signal Processing and Nonlinear Fourier Analysis},
  journal = {Oberwolfach Reports},
  year    = {2024},
  volume  = {21},
  number  = {4},
  pages   = {2671--2750},
  doi     = {10.4171/OWR/2024/46},
  url     = {https://ems.press/journals/owr/articles/14298801}
}

@article{alexis2024quantumsignalprocessingnonlinear,
  author    = {Alexis, Michel and Mnatsakanyan, Gevorg and Thiele, Christoph},
  title     = {Quantum signal processing and nonlinear Fourier analysis},
  journal   = {Revista Matem{\'a}tica Complutense},
  year      = {2024},
  volume    = {37},
  number    = {3},
  pages     = {655--694},
  doi       = {10.1007/s13163-024-00494-5},
  url       = {https://doi.org/10.1007/s13163-024-00494-5}
}

@article{balan2022permutationinvariantrepresentationsapplications,
title = {Permutation-invariant representations with applications to graph deep learning},
journal = {Applied and Computational Harmonic Analysis},
volume = {79},
pages = {101798},
year = {2025},
issn = {1063-5203},
doi = {https://doi.org/10.1016/j.acha.2025.101798},
url = {https://www.sciencedirect.com/science/article/pii/S1063520325000521},
author = {Radu Balan and Naveed Haghani and Maneesh Singh}
}

@misc{yilmaz2026arrowflowhierarchicalmachinelearning,
      title={ArrowFlow: Hierarchical Machine Learning in the Space of Permutations}, 
      author={Ozgur Yilmaz},
      year={2026},
      eprint={2604.04087},
      archivePrefix={arXiv},
      primaryClass={cs.LG},
      url={https://arxiv.org/abs/2604.04087}, 
}

@misc{katende2024nonlineargeneralizationbauerfiketheorem,
      title={A Nonlinear Generalization of the Bauer-Fike Theorem and Novel Iterative Methods for Solving Nonlinear Eigenvalue Problems}, 
      author={Ronald Katende},
      year={2024},
      eprint={2409.11098},
      archivePrefix={arXiv},
      primaryClass={math.NA},
      url={https://arxiv.org/abs/2409.11098}, 
}

@misc{fraysse2013robbinsmonroprocedureestimationparametric,
      title={A Robbins-Monro procedure for the estimation of parametric deformations on random variables}, 
      author={Philippe Fraysse and Hélène Lescornel and Jean-Michel Loubès},
      year={2013},
      eprint={1302.0110},
      archivePrefix={arXiv},
      primaryClass={math.ST},
      url={https://arxiv.org/abs/1302.0110}, 
}

\onecolumn

\appendix

\subsection{Observability via permutations}

\textbf{Lemma 1.} A real-valued matrix with distinct root of unity eigenvalues is not necessarily a permutation matrix in the sense that row space is preserved for non-full column rank matrices.

\vspace{2mm}

\textit{Proof.} Consider the counterexample
\begin{equation}
Q = \begin{pmatrix} -\frac{1}{2} & -\frac{\sqrt{3}}{2} & 0
\\
\frac{\sqrt{3}}{2} & -\frac{1}{2} & 0 
\\ 0 & 0 & 1
\end{pmatrix}, \{\lambda_1,\lambda_2,\lambda_3 : \lambda_i \in \text{eigenvalues}(Q)\} = \{ e^{2 \pi i /3}, e^{- 2 \pi i /3}, 1\}.
\end{equation}
Now, consider the full row rank
\begin{equation}
A = \begin{pmatrix} 1 & 0 & 0
\\
0 & 1 & 0 
\end{pmatrix} , \ \ \ \ \ AQ = \begin{pmatrix} -\frac{1}{2} & -\frac{\sqrt{3}}{2} & 0
\\
\frac{\sqrt{3}}{2} & -\frac{1}{2} & 0 
\end{pmatrix} 
\end{equation}
These two matrices have the same row space and so the columns are not permuted in the notion that we require and we are done.

$\square$

\vspace{2mm}

This proof is simple yet foundational for the following work. It is not sufficient that we restrict $A$ to have distinct root of unity eigenvalues to obtain observability. Instead, we must enforce the permutation property in the classical sense with $0's$ and $1's$; however, achieving this in a machine learning context is challenging. If were to create such a matrix by construction alone, it highly nontrivial to create an orthonormal, binary matrix that is learned. We will examine enforcing a permutation with a loss function, but learning binary values with high precision with neural networks is also nontrivial. This leads us to introduce permutation matrices with error tolerances.

\vspace{2mm}

\textbf{Lemma 2.} Suppose $Q \in \mathbb{R}_+^{n \times n}$ has distinct root of unity eigenvalues. Suppose 
\begin{equation}
\sum_{i} Q_{ij} = \sum_j Q_{ij} = 1 .
\end{equation}
Then $Q$ is necessarily a permutation matrix.

\vspace{2mm}

\textit{Proof.} By definition of $Q$, it is doubly stochastic. By the Birkhoff-von Neumann Theorem we have
\begin{equation}
\label{eqn:birkhoff}
Q = \sum_i \alpha_i \widetilde{Q}_i, \ \ \ \ \ \sum_i \alpha_i = 1
\end{equation}
where $\widetilde{Q}_i$ is necessarily a permutation matrix; however, since $Q$ necessarily has distinct root of unity eigenvalues, there is only the possibility $Q = \widetilde{Q}$, thus it is a permutation matrix. We show this more rigorously. Consider an eigenvector of $Q$, $\psi$. Thus, from \ref{eqn:birkhoff}
\begin{equation}
Q \psi = e^{2 \pi i j / d} \psi = \sum_i \alpha_i \widetilde{Q}_i \psi .    
\end{equation}
Now, we do not know for sure that $\psi$ is an eigenvector of $\widetilde{Q}_i$. However, we do know that
\begin{equation}
\label{eqn:squeeze}
1 = || Q \psi ||_2^2 = \Big| \Big| \sum_i \alpha_i \widetilde{Q}_i \psi \Big| \Big|_2^2 = \sum_i \sum_j \alpha_i \alpha_j \langle \widetilde{Q}_i \psi, \widetilde{Q}_j \psi \rangle \stackrel{\text{Cauchy-Schwarz}}{\leq} \sum_i \sum_j \alpha_i \alpha_j = \Big( \sum_i \alpha_i \Big)^2 = 1 . 
\end{equation}
Therefore, by a squeeze-theorem type argument, it must be true that $\widetilde{Q}_i \psi = \widetilde{Q}_j \psi$ since equality in the Cauchy-Schwarz inequality is only attained at equivalence. Therefore, we get from \ref{eqn:squeeze}
\begin{equation}
Q \psi = \sum_i \alpha_i \widetilde{Q}_i \psi = \sum_i \alpha_i \widetilde{Q} \psi = \widetilde{Q} \psi .
\end{equation}
Since $\psi$ was any eigenvector, all eigenvectors and eigenvalues are the same for $Q$ and $\widetilde{Q}$. Since permutation matrices are diagonalizable, their eigenbasis diagonalizations are the same, thus they are the same matrices and we are done with the first part of the proof.

$ \square $

\vspace{2mm}

\textbf{Theorem 1.} Let $A \in \mathbb{R}^{n \times n}$ be a permutation matrix with distinct root of unity eigenvalues, and let $C \in \mathbb{R}^{m \times n}, n > m$ be the matrix of equation \ref{eqn:ODE_system} (hence not identically the zero matrix). Furthermore, suppose $C$ is nonconstant among permutations in the sense that no column in the column space of $C$ can be formulated with constant coefficients among other linearly independent columns in the space that span the image the of $C$. More specifically,
\begin{align}
& \psi_k \in \{ \{ \psi_i \}_i : \psi_i \ \text{is a column of} \ C \}, 
\\
& \psi_k \neq \sum_{j \neq k} \alpha \varphi_j, \text{Im} \Big( \text{span} \{ \varphi_j : \varphi_j \ \text{is a linearly independent column of} \ C \} \Big) = \mathbb{R}^m  .   
\end{align}
Then the pair $(C,A)$ is observable.

\vspace{2mm}

\textit{Remark.} The general idea of this proof is that $Q$ sufficiently permutes the columns of $C$ into linear independence in the column space of the observability matrix.

\vspace{2mm}

\textit{Proof.} We use induction over the columns in the column space of $A$. Let $A$ have $n$ columns. We begin with a base case of $n=2$. Let $C$ be such that its width is $2$. We show linear independence. We show the second column is not linearly dependent of the first (i.e. it is not the first scaled by a constant).
Consider
\begin{equation}
\begin{pmatrix}
[C]_1  \\ [CA]_1 \\ \vdots \\ [C A^{n-1}]_1 
\end{pmatrix}   \ \ \ \ \ \text{and} \ \ \ \ \ \begin{pmatrix}
[C]_2  \\ [CA]_2 \\ \vdots \\ [C A^{n-1}]_2 
\end{pmatrix} ,
\end{equation}
where $[\cdot]_k$ denotes the column. We remark $A^i \in \mathbb{R}^{2 \times 2}$ is also a permutation matrix. If the second column of $C$ is not a scaling of the first, the proof is automatically complete. Note that if $C \in \mathbb{R}^{1 \times 2}$, then the second column is automatically a scaling of the first; if $C \in \mathbb{R}^{2 \times 2}$, this is not necessarily true. So that the proof is not trivial, suppose that the second column of $C$ is not a scaling of the first, and denote this constant $\alpha$. Since $A$ is a permutation matrix, there exists some exponent $i$ such that $A^i$ swaps the two columns of $C$. Thus, the new constant relating the two is $1/\alpha$ for this block.
Hence, we have a full rank condition on the observability matrix.

\vspace{2mm}

We induct on the general case. Select $m$ columns of the observability matrix $\mathcal{O}$ be linearly independent. We show the $(m+1)$-th column is not a linear combination of the previous $m$. Now, for the first $m$ columns, we must have
\begin{equation}
\text{dim} \Bigg( \Big\{ \text{concat}_i [CA^i]_j \Big\}_{j=1}^m \Bigg) = \text{dim} \Bigg(\Big\{ \text{concat}_i [C]_{j, \sigma_i} \Big\}_{j=1}^m \ \Bigg) = m .
\end{equation}
Here, we have denoted $[C]_{j, \sigma_i}$ the $j$-th column of $C$ under permutation $\sigma_i$.
We will use notation $[\cdot]_{k,j,\sigma_i}$ to denote the $(kj)$ entry of $\mathcal{O}$ under permutation $\sigma_i$. Suppose for the sake of contradiction
\begin{align}
[C]_{k,m+1,\sigma_k} = \sum_{j < m+1} \alpha_j  [C]_{k,j,\sigma_k}, \ \ \ \ \ [C]_{l,m+1,\sigma_{l}} = \sum_{j < m+1} \alpha_j  [C]_{l,j,\sigma_l}
\end{align}
for $k \neq l$. The above is saying that the final entries in each column can be expressed with the same coefficients under different permutations. Thus a solution cannot exist without triviality (constancy), which violates the constant coefficient assumption. Thus, we have linear independence, and we are done.

$ \square $

\vspace{2mm}

\textit{Remark.} If $A$ does not have distinct eigenvalues, $A$ does not necessarily permute all entries (take, for example, the identity matrix, which is a trivial permutation).

\vspace{2mm}

\textbf{Lemma 3.} Let $Q$ be a permutation matrix, and let $P$ be a matrix such that
\begin{equation}
| \lambda_i(Q) - \lambda_i(P) | = \epsilon_i^2 ,
\end{equation}
and even moreover, the closest eigenvalue to $\lambda_i(P)$ is $\lambda_i(Q)$. Then there exists a matrix $\Xi$ such that
\begin{equation}
\sum_{i=1}^n \epsilon_i^2 \leq \mathcal{O}(n||\Xi||_2)
\end{equation}
and
\begin{equation}
P = Q +  \Xi .
\end{equation}

\vspace{2mm}

\textit{Proof.} The proof is an application of the Bauer-Fike Theorem, which states that
\begin{equation}
\label{eqn:bauer_fike}
| \lambda - \mu | \leq \kappa_2(V) || \Xi ||_2 = ||V||_2 ||V^{-1}||_2 ||\Xi||_2 .
\end{equation}
Here, $\lambda = \lambda(P)$ is an eigenvalue of $Q = V \Lambda V^{-1}$ (since every permutation matrix is diagonalizable over $\mathbb{C}$, and diagonalization over $\mathbb{R}$ is not needed for the Bauer-Fike Theorem), and $\mu = \mu(Q) = \lambda(Q)$ is an eigenvalue of $Q$. In particular, we have chosen the spectral norm. Now, summing over the eigenvalues with \ref{eqn:bauer_fike},
\begin{equation}
\sum_i | \lambda_i(P) - \mu_i(Q) | = \sum_i \epsilon_i^2 \leq \sum_i || V||_2 ||V^{-1}||_2 || \Xi||_2 = \mathcal{O}(n||\Xi||_2) ,
\end{equation}
where we also use $||V||_2 = ||V^{-1}||_2 = 1$, which uses the Spectral Theorem, since $Q$ is a permutation (hence normal, so unitarily diagonalizable).

$ \square $

\vspace{2mm}

\textit{Remark.} This proof tells us that as the differences in eigenvalues become small, the magnitude of the perturbation matrix becomes small. This is useful for us because we will be matching eigenvalues in a loss function.

\vspace{2mm}

\textbf{Theorem 2.} Let $C \in \mathbb{R}^{m \times n}$ be the matrix as defined in \ref{eqn:ODE_system}. Let the hypotheses of Theorem 1 on $C$ hold. Let $Q \in \mathbb{R}^{n \times n}$ be a permutation matrix and let $P$ be a matrix with distinct eigenvalues sufficiently close to roots of unity in the sense that $P = Q + \Xi$, $| \lambda_i(Q) - \lambda_i(P)| = \epsilon_i, \sum_i \epsilon_i^2 \leq \mathcal{O}(||\Xi||_F^2)$. Let
\begin{equation}
\sum_i \Big|\Big| (P^i)^T C^T C P^i - (Q^i)^T C^T C Q^i \Big|\Big|_F = \mathcal{O}( n ||\Xi||_F^2) .
\end{equation}
If
\begin{equation}
|| \Xi ||_F^2 \leq \epsilon
\end{equation}
holds for sufficiently small $\epsilon$, then the pair $(C,P)$ is observable.

\vspace{2mm}

\textit{Proof.} First, we employ Theorem 1. Immediately, we have the pair $(C,A)$ is observable. Now, notice
\begin{align}
\sum_i ( (P^i)^T C^T C P^i - (Q^i)^T C^T C Q^i ) & = \text{concat}_i \Big(  CP^i \Big)^T \text{concat}_i \Big( CP^i \Big)  - \text{concat}_i \Big( CQ^i \Big)^T \text{concat}_i \Big( CQ^i \Big)  \\
& = \mathcal{O}_{P}^T \mathcal{O}_{P}  - \mathcal{O}^T \mathcal{O} ,
\end{align}
where $\mathcal{O}_{P}$ is the observability matrix with respect to $P$ instead of $A$. Hence, by the triangle inequality and our assumptions, we have
\begin{equation}
\Big|\Big|\mathcal{O}_{P}^T \mathcal{O}_{P} - \mathcal{O}^T \mathcal{O} \Big|\Big|_F \leq \sum_i \Big|\Big| (P^i)^T C^T C P^i - (Q^i)^T C^T C Q^i \Big|\Big|_F  = \mathcal{O}(n || \Xi ||_F^2 ) = \mathcal{O}(n \epsilon ) .
\end{equation}
In other words, the new observability matrix under $P$ is sufficiently close to the true observability matrix, which is known to be full column rank by Theorem 1, if $||\Xi||_F^2$ is sufficiently small. Now, we have $\mathcal{O}^T \mathcal{O}$ has all nonzero eigenvalues since $\mathcal{O}$ is full column rank. Also, any matrix of the form $A^T A$ is normal. By Lemma 4, we must have 
\begin{equation}
\min \Bigg\{ \lambda_1(\mathcal{O}_P^T\mathcal{O}_P), \hdots, \lambda_n(\mathcal{O}_P^T\mathcal{O}_P) \Bigg\} > 0 ,
\end{equation}
i.e. all eigenvalues are sufficiently close (and note that they are positive), which implies $\mathcal{O}_P$ is full column rank, and we are done.

$ \square $

\vspace{2mm}

\textbf{Lemma 4.} Let $X \in \mathbb{R}^{n \times n}$ be a square, finite-valued, normal, full column rank matrix. Let $Y \in \mathbb{R}^{n \times n}$ also be normal and sufficiently close to $X$ in the sense that
\begin{equation}
|| Y - X ||_F^2 \leq \epsilon || X||_F^2 .
\end{equation}
Then $Y$ is also full column rank if $\epsilon$ is sufficiently small.

\vspace{2mm}

\textit{Proof.} The proof is also an application of the Hoffman-Wielandt inequality \cite{Chafai2011}. By this inequality, we have
\begin{equation}
\min_{\sigma \in \mathcal{S}_s} \sum_{i=1}^n | \lambda_i(Y) - \lambda_{\sigma(i)}(X) 
|^2 \leq || Y - X ||_F^2 \leq \epsilon ||X||_F^2 .
\end{equation}
Taking $\epsilon$ sufficiently small, we have the eigenvalues of $Y$ approach the eigenvalues of $X$. Even though there is a summation on the left-hand side, all individual terms in the summation must be sufficiently close. Since the eigenvalues of $X$ are nonzero due to $X$ being full column rank, so are the eigenvalues of $Y$, and we are done.

$ \square $

\subsection{Observability with the Fourier transform}

Although the relation between the Fourier transform and the impulse response has some relation to observability \cite{signals5030023} \cite{Wang2001TechnicalCI} \cite{SUBASI20141669}, we find the topic remains to be explored further, specifically for machine learning. Thus, our result is meaningful both in ODE control theory and specifically for machine learning.

\subsection{Observability with the Fourier transform with the impulse response}
\textbf{Theorem 3.} Consider the Fast Fourier transform of the matrix $C e^{A k \Delta t} B$, $\mathcal{F}[C e^{A k \Delta t} B]$.
$C \in \mathbb{R}^{m \times n}$ be full row rank, $A \in \mathbb{R}^{n \times n}, B\in \mathbb{R}^{n \times m}, n > m$, where $A$ is full column rank. Suppose $A = V \Lambda V^{-1}$ is diagonalizable. Let $C$ satisfy the permutation-invariant property as in Theorem 1. Suppose
\begin{equation}
\label{eqn:kernel_cond_thm3}
\text{ker} \Bigg( CV \Phi^{j_1} V^{-1}  \Bigg) \not\subseteq  \text{ker} \Bigg( CV \Phi^{j_2} V^{-1}  \Bigg) .
\end{equation}
More specifically, let the loss
\begin{align}
\mathcal{L} = & \ \sum_k \text{relu} \Bigg( \text{positive constant} - \text{min}_{k_1,k_2} \Big| \Lambda_{k_1 k_1} - \Lambda_{k_2 k_2} \Big| \Bigg) 
\\
& + \sum_{j_1} \sum_{j_2 \neq j_1} \sum_k \text{relu} \Bigg( \text{positive constant} - \Big| \Phi_{kk}^{j_1} - \Phi_{kk}^{j_2} \Big| \Bigg)
\\
& +  \text{relu} \Bigg(  \Bigg| \frac{1}{n} \sum_k e^{i \theta_k(\Lambda) } \Bigg| - \text{positive constant}    \Bigg)  
\end{align}
be exactly satisfied, where $ 1 \leq j_1,j_2 \leq L-1$ and $I - e^{A L \Delta t}$ is full rank. We have defined
\begin{align}
\Phi^j = \text{diag} \Bigg( \frac{ 1 - e^{\lambda_1 \Delta t}}{ 1 - e^{\lambda_1 \Delta t - 2 \pi i \frac{j}{L}}}, \hdots, \frac{ 1 - e^{\lambda_n \Delta t}}{ 1 - e^{\lambda_n \Delta t - 2 \pi i \frac{j}{L}}} \Bigg) .
\end{align}
Then with high probability the system is observable.

\vspace{2mm}

\textit{Corollary.} We can relax the condition that $(I -e^{A L \Delta t})$ is full column rank by taking
\begin{equation}
\text{det}\Bigg(   \mathcal{F} \Bigg[ C e^{A k \Delta t} B  \Bigg]_j  \Bigg) > 0  
\end{equation}
for all $j$, but generally this is computationally expensive.

\vspace{2mm}

\textit{Remark.} We can enforce the necessary hypothesis constraint by adding the following loss to the objective:
\begin{align}
\mathcal{L} & = \sum_{j_1} \sum_{j_2 \neq j_1} \text{relu} \Bigg( \text{positive constant} - \Bigg| \Bigg| \underbrace{ CV \varphi_{j_1,j_1} }_{=0} -  CV \varphi_{j_1,j_2 } \Bigg| \Bigg|_2^2 \Bigg) 
\\
& =  \sum_{j_1} \sum_{j_2 \neq j_1} \text{relu} \Bigg( \text{positive constant} - \Bigg| \Bigg|  CV \varphi_{j_1,j_2 } \Bigg| \Bigg|_2^2 \Bigg).
\end{align}
Here, we have defined
\begin{equation}
\varphi_{j,j_2} = \text{column} \Bigg( \frac{ - [(Q^{j})^{-1} \widetilde{P}^{j}]_1 }{ 1 - e^{\lambda_1 \Delta t - 2 \pi i {j_2} / L }}, \hdots, \frac{ - [(Q^{j})^{-1} \widetilde{P}^{j}]_m }{ 1 - e^{\lambda_m \Delta t - 2 \pi i {j_2} / L }}, \frac{ 1 }{ 1 - e^{\lambda_1 \Delta t - 2 \pi i {j_2} / L }},\hdots, \frac{ 1 }{ 1 - e^{\lambda_n \Delta t - 2 \pi i {j_2} / L }} \Bigg)  
\end{equation} 
for appropriate choices of $Q,\widetilde{P}$ (see proof). The above means that, under suitable assumptions, there exists a vector $\psi$ so that $\psi$ is in the kernel of one but not the other \ref{eqn:kernel_cond_thm3}. In general, here, we are more interested in the kernels not being equal rather than containment, because $\psi$ could be in the kernel of the larger one but not the smaller.

\vspace{2mm}

\textit{Proof.} First, we remark this proof goes hand-in-hand with Theorem 4, so we also advise reading the proof of that theorem if details seem omitted.

\vspace{2mm}

Observe for general $j > 0$, using strategies from \cite{gu2022efficientlymodelinglongsequences}
\allowdisplaybreaks 
\begin{align}
& \mathcal{F} \Bigg[ C e^{A k \Delta t} B \Bigg]_j
\\
& = \sum_{k=0}^{L-1}  C \exp \Big\{ A k \Delta t \Big\} B \exp \Big\{ - 2\pi i \frac{jk}{L} \Big\}
\\
& = C \Bigg(  \sum_{k=0}^{L-1}   \exp \Big\{ A k \Delta t \Big\} \exp \Big\{ - 2\pi i \frac{jk}{L} \Big\} \Bigg) B
\\
& = C \Bigg(  \sum_{k=0}^{L-1}   \Big( \exp \Big\{ A  \Delta t \Big\} \exp \Big\{ - 2\pi i \frac{j}{L} \Big\} \Big)^k  \Bigg) B
\\
& \stackrel{(1)}{=} C \Bigg(  (I - \exp\{ - 2\pi i \frac{j}{L} \} \exp\{ A  \Delta t\})^{-1} (I - \exp\{ - 2\pi i j \} \exp\{ A L \Delta t  \}) \Bigg) B
\\
& = C \Bigg(  (I - \exp\{ - 2\pi i \frac{j}{L} \} \exp\{ A  \Delta t\})^{-1} (I -  \exp\{ A L \Delta t  \}) \Bigg) B.
\end{align}
$(1)$ is by the geometric series formula for matrices. Now, since $A$ is diagonalizable, we have
\begin{align}
\exp\{ A \Delta t\} = V \text{diag} \{ e^{\lambda_1 \Delta t}, \hdots, e^{\lambda_n \Delta t} \} V^{-1} ,
\end{align}
and so
\begin{align}
(I - \exp \{ -2 \pi i \frac{J}{L} \} \exp \{ A \Delta t \} )^{-1} & = (I - \exp \{ -2 \pi i \frac{j}{L} \}  V \text{diag} \{ e^{\lambda_1 \Delta t}, \hdots, e^{\lambda_n \Delta t} \} V^{-1} )^{-1}
\\
& = [ V (I - \exp \{ -2 \pi i \frac{j}{L} \} \text{diag} \{ e^{\lambda_1 \Delta t}, \hdots, e^{\lambda_n \Delta t} \} ) V^{-1} ]^{-1}
\\
& =  V (I - \exp \{ -2 \pi i \frac{j}{L} \} \text{diag} \{ e^{\lambda_1 \Delta t}, \hdots, e^{\lambda_n \Delta t} \} ) ^{-1} V^{-1}
\\
& = V \text{diag} \Bigg\{ \frac{1}{1 - e^{\lambda_1 \Delta t -2 \pi i \frac{j}{L}}} , \hdots, \frac{1}{1 - e^{\lambda_n  \Delta t -2 \pi i \frac{j}{L}}} \Bigg\} V^{-1} .
\end{align}
We will denote
\begin{equation}
\Lambda_j = \text{diag} \Bigg\{ \frac{1}{1 - e^{\lambda_1 \Delta t -2 \pi i \frac{j}{L}}} , \hdots, \frac{1}{1 - e^{\lambda_n  \Delta t -2 \pi i \frac{j}{L}}} \Bigg\} .
\end{equation}
Hence, we see
\begin{align}
\mathcal{F} \Bigg[ C e^{A k \Delta t} B  \Bigg]_j & = C V \text{diag} \Bigg\{ \frac{1}{1 - e^{\lambda_1 \Delta t -2 \pi i \frac{j}{L}}} , \hdots, \frac{1}{1 - e^{\lambda_n  \Delta t -2 \pi i \frac{j}{L}}} \Bigg\} V^{-1}(I - e^{A L \Delta t - 2 \pi i j I } ) B 
\\
& = CV \Lambda V^{-1} (I - e^{A L \Delta t  } ) B .
\end{align}
Now, we will find a sufficient condition so that
\begin{align}
\label{eqn:fourier_rowspace}
\text{row} \Bigg( CV \Lambda_{j_1} V^{-1} (I - e^{A L \Delta t  } ) \Bigg) \neq \text{row} \Bigg( CV \Lambda_{j_2} V^{-1} (I - e^{A L \Delta t  } )  \Bigg)
\end{align}
for $j_1,j_2 \leq n-m$. For each permutation that creates at least $1$ new dimension in the rowspace, we need at least $n-m$ permutations. By the pigeonhole principle, we have
\begin{align}
\text{dim} \Bigg( \text{row} \Bigg( CV \Lambda_{j_1} V^{-1} (I - e^{A L \Delta t  } )  \oplus \hdots \oplus CV \Lambda_{j_p} V^{-1} (I - e^{A L \Delta t } )  \Bigg) \Bigg) = n.
\end{align}
We will not only construct but also enforce a vector $\psi$ to be such that
\begin{align}
0 = CV \Lambda_{j_1} V^{-1} (I - e^{A L \Delta t  } ) \psi, 0 \neq  CV \Lambda_{j_2} V^{-1} (I - e^{A L \Delta t  } ) \psi .
\end{align}
Our argument is to enforce the rowspace of $CV\Lambda_{j}$ to change for all $j$, thus allowing the pigeonhole principle to take effect. We refer to Theorem 4 for an argument for this. In essence if we can ensure $|\Lambda_{j_1,kk} - \Lambda_{j_2,kk}| > 0$ sufficiently large, the rowspace of $CV\Lambda_j V^{-1}$ permutes with respect to $j$ sufficiently better than simply exponentiating the eigenvalue due to the nonlinearity in the definition of $\Lambda_j$.

\vspace{2mm}

Also, we note (following the proof of Theorem 5, see Appendix \ref{app:thm5}),
\begin{align}
I - e^{A L \Delta t} & = I - e^{ V \Lambda V^{-1} \Delta t}
\\
& = V V^{-1} - V \text{diag} \Bigg( e^{\lambda_1 \Delta t}, \hdots, e^{\lambda_n \Delta t} \Bigg) V^{-1} 
\\
& = V \text{diag} \Bigg( 1 - e^{\lambda_1 \Delta t}, \hdots, 1 - e^{\lambda_n \Delta t} \Bigg) V^{-1} .
\end{align} 
Thus, following suit of the procedure as in Theorem 4, our loss function is 
\begin{align}
\mathcal{L} = & \ \sum_k \text{relu} \Bigg( \text{positive constant} - \text{min}_{k_1,k_2} \Big| \Lambda_{k_1 k_1} - \Lambda_{k_2 k_2} \Big| \Bigg) 
\\
& + \sum_{j_1} \sum_{j_2 \neq j_1} \sum_k \text{relu} \Bigg( \text{positive constant} - \Big| \Phi_{kk}^{j_1} - \Phi_{kk}^{j_2} \Big| \Bigg)
\\
& + \text{relu} \Bigg(  \Bigg| \frac{1}{n} \sum_k e^{i \theta_k(\Lambda) } \Bigg| - \text{positive constant}    \Bigg) .
\end{align}
We have defined
\begin{align}
\Phi^j = \text{diag} \Bigg( \frac{ 1 - e^{\lambda_1 \Delta t}}{ 1 - e^{\lambda_1 \Delta t - 2 \pi i \frac{j}{L}}}, \hdots, \frac{ 1 - e^{\lambda_n \Delta t}}{ 1 - e^{\lambda_n \Delta t - 2 \pi i \frac{j}{L}}} \Bigg) .
\end{align}
The last term involving variance ensures dispersion of $\Phi$ across the complex plane.

\vspace{2mm}

We remark the pigeonhole principle only take affects if the permutations add new rowspace dimensions. It is true with enough permutations, there is high probability the rowspace union condition will be fulfilled, but a permutation itself does not guarantee this new rowspace addition. Thus, the high probability argument is crucial here as well. We discuss this more in the proof of Theorem 4.

\vspace{2mm}

Now, assuming the kernel criteria has met and continuing the proof, notice that there necessarily exists a $j$ such that
\begin{align}
C V \Lambda_j V^{-1} (I - e^{A L \Delta t} ) \psi \neq 0,
\end{align}
and observe there is an equivalence
\begin{align}
C V \Lambda_j V^{-1} (I - e^{A L \Delta t} ) \psi & = \mathcal{F} \Bigg[ C e^{A k \Delta t} \Bigg]_j \psi
\\
& = C \Bigg(  \sum_{k=0}^{L-1}   \exp \Big\{ A k \Delta t \Big\} \exp \Big\{ - 2\pi i \frac{jk}{L} \Big\} \Bigg) \psi .
\end{align} 
Thus, it must be true that, for all $\psi$, there exists some $k$ such that $ C e^{A k \Delta t} \psi  \neq 0 $. Therefore, it must be true that $C \exp\{Ak \Delta t\} \neq 0$ and we have the system is observable in the $(C,\overline{A})$ case. The $(C,A)$ case follows (see the proof of Theorem 5).

\vspace{2mm}

There is one final remark: we have not enforced $I - e^{A L \Delta t}$ to be full column rank, and indeed it must be true that $h_0$ is not annihilated by $I - e^{A L \Delta t}$. However, by Lemma 5, we have
\begin{equation}
\text{dim} \Bigg( \text{col} \Bigg( V \Lambda_j V^{-1} (I - e^{A L \Delta t  } ) \Bigg) \Bigg) = n .
\end{equation}
Also, we can assume this, which we did in our theorem hypotheses. Hence, this matrix has trivial nullspace, and the nullspace condition only comes from the shifting of $C$. We remark while $h_0$ does need to be any vector in $\mathbb{R}^n$, $\widetilde{\psi}$ does not. This completes the proof.

$ \square $

\vspace{2mm}

\textit{Remark.} We can systemically construct such a $\psi$ with a dependence on $j$ so that the above holds for all $j_1,j_2$. Observe $V^{-1}$ is surjective, but we do not yet have $I - e^{A L \Delta t}$ is surjective. Instead, we make some notes. Denote
\begin{equation}
\widetilde{\psi} = V^{-1} (I - e^{A L \Delta t} ) \psi .
\end{equation}
If we construct $\widetilde{\psi}$ in such a way that
\begin{align}
CV \Lambda_{j_1} \widetilde{\psi} = 0, CV\Lambda_{j_2} \widetilde{\psi} \neq 0
\end{align}
holds for appropriate $j_1,j_2$, then it does not necessarily need to hold that $\widetilde{\psi}$ is any vector in $\mathbb{R}^n$. It is only necessary that $\widetilde{\psi}$ is in the image of $V^{-1} (I - e^{A L \Delta t} )$, which we assumed. Thus, we can apply Lemma 5 later without circular logic. Now, notice
\begin{align}
0 & = \ C V \Lambda_{j_1} \widetilde{\psi}  = C V \text{column} \Bigg( \frac{ \widetilde{\psi}_1 }{ 1 - e^{\lambda_1 \Delta t - 2 \pi i j_1 / L }}, \hdots, \frac{ \widetilde{\psi}_n }{ 1 - e^{\lambda_n \Delta t - 2 \pi i j_1 / L }} \Bigg) 
\\
0 & \neq C V \Lambda_{j_2} \widetilde{\psi} = C V \text{column} \Bigg( \frac{ \widetilde{\psi}_1 }{ 1 - e^{\lambda_1 \Delta t - 2 \pi i j_2 / L }}, \hdots, \frac{ \widetilde{\psi}_n }{ 1 - e^{\lambda_n \Delta t - 2 \pi i j_2 / L }} \Bigg) .
\end{align}
Here $\widetilde{\psi}$ is not necessarily any vector in $\mathbb{R}^n$, but we do not need this fact. Now, observe
\begin{align}
0 & = C V \text{column} \Bigg( \frac{ \widetilde{\psi}_1 }{ 1 - e^{\lambda_1 \Delta t - 2 \pi i {j_1} / L }}, \hdots, \frac{ \widetilde{\psi}_n }{ 1 - e^{\lambda_n \Delta t - 2 \pi i {j_1} / L }} \Bigg) 
\\
& = \text{column} \Bigg( \sum_k \frac{ [CV]_{1k} \widetilde{\psi}_k }{ 1 - e^{\lambda_k \Delta t - 2 \pi i j_1 / L }}, \hdots, \sum_k \frac{ [CV]_{nk} \widetilde{\psi}_k }{ 1 - e^{\lambda_k \Delta t - 2 \pi i j_1 / L }} \Bigg)
\\
0 & \neq  \text{column} \Bigg( \sum_k \frac{ [CV]_{1k} \widetilde{\psi}_k }{ 1 - e^{\lambda_k \Delta t - 2 \pi i j_2 / L }}, \hdots, \sum_k \frac{ [CV]_{nk} \widetilde{\psi}_k }{ 1 - e^{\lambda_k \Delta t - 2 \pi i j_2 / L }} \Bigg)
\\
& =  C V \text{column} \Bigg( \frac{ \widetilde{\psi}_1 }{ 1 - e^{\lambda_1 \Delta t - 2 \pi i j_2 / L }}, \hdots, \frac{ \widetilde{\psi}_n }{ 1 - e^{\lambda_n \Delta t - 2 \pi i j_2 / L }} \Bigg)  .
\end{align}
Thus, it must be true that
\begin{align}
0 & =  \begin{cases} \sum_k \frac{ [CV]_{1k} \widetilde{\psi}_k }{ 1 - e^{\lambda_k \Delta t - 2 \pi i j_1 / L }} \\
\vdots \\
\sum_k \frac{ [CV]_{mk} \widetilde{\psi}_k }{ 1 - e^{\lambda_k \Delta t - 2 \pi i j_1 / L }}
\end{cases}.
\end{align}
One method is to use a kernel solver with computing, but the kernel basis vectors are not consistent among new iterations, thus enforcing a condition to find such a $\widetilde{\psi}$ in this basis but not in another corresponding to $j_2$ will present challenges.

\vspace{2mm}

Instead, we will construct $\widetilde{\psi}$ systematically, which is consistent and reproducible under all kernel bases, and show this constructed vector is not in the kernel corresponding to $j_2$. We will use our assumption that the first $m \times m$ block of $C$ is linearly independent, which is easy to enforce with machine learning by adding a perturbation $+ \epsilon I$. First, note that $k \in \{ 1,\hdots,n\}, p \in \{ 1 \hdots, n\}$ and
\begin{align}
\sum_k \frac{ [CV]_{lk} \widetilde{\psi}_k }{ 1 - e^{\lambda_k \Delta t - 2 \pi i j_1 / L }} = \sum_k \sum_p \frac{ C_{lp} V_{pk} \widetilde{\psi}_k }{ 1 - e^{\lambda_k \Delta t - 2 \pi i j_1 / L }} = 0.
\end{align}
Now, since $C$ is a block matrix with the first $m \times m$ block linearly independent, we will choose, for $m+1 \leq k \leq n$, so that $\widetilde{\psi}_k = 1$.
Hence, we have 
\begin{align}
\sum_{1 \leq k \leq m} \sum_p \frac{ C_{lp} V_{pk} \widetilde{\psi}_k  }{ 1 - e^{\lambda_k \Delta t - 2 \pi i j_1 / L }} + \sum_{m < k \leq n} \sum_p \frac{ C_{lp} V_{pk}   }{ 1 - e^{\lambda_k \Delta t - 2 \pi i j_1 / L }} = 0 . 
\end{align}
We will denote the matrices $Q^j \in \mathbb{R}^{m \times m}, P^j \in \mathbb{R}^{m \times (n-m)}$, and vector $\widetilde{P}^{j_1} \in \mathbb{R}^{m}$ with elements
\begin{align}
Q_{lk}^{j_1} = \sum_p \frac{ C_{lp} V_{pk}  }{ 1 - e^{\lambda_k \Delta t - 2 \pi i j_1 / L }}, \ \ \ \ \ P_{lk}^{j_1} = \sum_p \frac{ C_{lp} V_{p(k+m)}  }{ 1 - e^{\lambda_{k+m} \Delta t - 2 \pi i j_1 / L }}, \ \ \ \ \ \widetilde{P}_l^{j_1} = \sum_k \sum_p \frac{ C_{lp} V_{p(k+m)}  }{ 1 - e^{\lambda_{k+m} \Delta t - 2 \pi i j_1 / L }}.
\end{align}
Thus, our system of equations to solve is
\begin{align}
Q^{j_1} ( \widetilde{\psi}_{1}, \widetilde{\psi}_{2}, \hdots, \widetilde{\psi}_{m} )^T + \widetilde{P}^{j_1}  = 0
\end{align}
Hence, we can solve the remaining $\widetilde{\psi}$ using
\begin{equation}
(\widetilde{\psi}_{1}, \widetilde{\psi}_{2  }, \hdots, \widetilde{\psi}_m )^T = - (Q^{j_1})^{-1} \widetilde{P}^{j_1}  .
\end{equation}
Now that we have constructed such a $\widetilde{\psi}$, it remains to enforce that
\begin{align}
\widetilde{\psi} \notin \text{ker} \Bigg( CV\Lambda_{j_2} \Bigg) .
\end{align}
Thus, a sufficient condition so that either $\widetilde{\psi} \notin \text{ker}(CV\Lambda_{j_1}) \cap \text{ker}(CV\Lambda_{j_2})$ is that
\begin{align}
& \Bigg| \Bigg| C V \text{column} \Bigg( \frac{ - [(Q^{j_1})^{-1} \widetilde{P}^{j_1}]_1 }{ 1 - e^{\lambda_1 \Delta t - 2 \pi i {j_1} / L }}, \hdots, \frac{ - [(Q^{j_1})^{-1} \widetilde{P}^{j_1}]_m }{ 1 - e^{\lambda_m \Delta t - 2 \pi i {j_1} / L }}, \frac{ 1}{ 1 - e^{\lambda_{m+1} \Delta t - 2 \pi i {j_1} / L }},\hdots, \frac{ 1 }{ 1 - e^{\lambda_n \Delta t - 2 \pi i {j_1} / L }} \Bigg) 
\\
&  - C V \text{column} \Bigg( \frac{ - [(Q^{j_1})^{-1} \widetilde{P}^{j_1}]_1 }{ 1 - e^{\lambda_1 \Delta t - 2 \pi i {j_2} / L }}, \hdots, \frac{ - [(Q^{j_1})^{-1} \widetilde{P}^{j_1}]_m }{ 1 - e^{\lambda_m \Delta t - 2 \pi i {j_2} / L }}, \frac{ 1 }{ 1 - e^{\lambda_{m+1} \Delta t - 2 \pi i {j_2} / L }},\hdots, \frac{ 1 }{ 1 - e^{\lambda_n \Delta t - 2 \pi i {j_2} / L }} \Bigg)  \Bigg| \Bigg|_2^2 
\\
& = \Bigg| \Bigg| C V \text{column} \Bigg( \frac{ - [(Q^{j_1})^{-1} \widetilde{P}^{j_1}]_1 }{ 1 - e^{\lambda_1 \Delta t - 2 \pi i {j_2} / L }}, \hdots, \frac{ - [(Q^{j_1})^{-1} \widetilde{P}^{j_1}]_m }{ 1 - e^{\lambda_m \Delta t - 2 \pi i {j_2} / L }}, \frac{ 1 }{ 1 - e^{\lambda_{m+1} \Delta t - 2 \pi i {j_2} / L }},\hdots, \frac{ 1 }{ 1 - e^{\lambda_n \Delta t - 2 \pi i {j_2} / L }} \Bigg)  \Bigg| \Bigg|_2^2  > 0 .
\end{align}
The first term evaluates to just zero by our construction. We have used the norm 
\begin{equation}
|| z ||_2  = \sqrt{ \sum_i ( \text{Re}(z_i)^2 + \text{Im}(z_i)^2 ) }
\end{equation}
when $z$ is complex. We will denote
\begin{equation}
\varphi_{j,j_2} = \text{column} \Bigg( \frac{ - [(Q^{j})^{-1} \widetilde{P}^{j}]_1 }{ 1 - e^{\lambda_1 \Delta t - 2 \pi i {j_2} / L }}, \hdots, \frac{ - [(Q^{j})^{-1} \widetilde{P}^{j}]_m }{ 1 - e^{\lambda_m \Delta t - 2 \pi i {j_2} / L }}, \frac{ 1 }{ 1 - e^{\lambda_{m+1} \Delta t - 2 \pi i {j_2} / L }},\hdots, \frac{ 1 }{ 1 - e^{\lambda_n \Delta t - 2 \pi i {j_2} / L }} \Bigg)  .
\end{equation}
Thus our loss function is 
\begin{equation}
\label{eqn:fourier_kernel_loss}
\mathcal{L} = \sum_{j_1} \sum_{j_2 \neq j_1} \text{relu} \Bigg( \text{positive constant} - \Bigg| \Bigg| CV \varphi_{j_1,j_1} -  CV \varphi_{j_1,j_2 } \Bigg| \Bigg|_2^2 \Bigg) .
\end{equation}
We can formulate $Q,P,\widetilde{Q}$ by formulating
\begin{equation}
CV \Lambda_j 
\end{equation}
and extracting the corresponding elements. Hence, we have shown there exists a $\psi$ such that
\begin{equation}
CV \Lambda_{j_1} V^{-1} (I - e^{A L \Delta t  } )  \psi = 0, CV \Lambda_{j_2} V^{-1} (I - e^{A L \Delta t  } )  \psi \neq 0 , 
\end{equation}
and so their rowspace is not the same. 

\vspace{2mm}

\vspace{2mm}

\textit{Remark.} The proof does not automatically follow by taking 
\begin{equation}
\text{det}( \mathcal{F}[Ce^{A \Delta t}B] ) > 0
\end{equation}
without the condition. Consider the counterexample $C=(I_m \ 0), A=I_n, B=(I_m \ 0)^T$. Thus, we have $\mathcal{F}[Ce^{At}B] = \mathcal{F}[e(\Delta t)I]$ . But we have
\begin{align}
[ \mathcal{F}[e(\Delta t)I] ]_j \propto I .
\end{align}
Clearly, $\text{det}([ \mathcal{F}[e(\Delta t)I] ]_j ) > 0$ for all $j$ but the system is clearly not observable, i.e. since $A = I$, no columns of $C$ shift, so the observability matrix is not full column rank.

\vspace{2mm}

Also, we illustrate our method is not contradicted by this counterexample. Observe 
\begin{align}
\varphi_j & = \Bigg( \frac{-[(Q^j)^{-1} \widetilde{P}^j]_1 }{1 - e^{\lambda_1 \Delta t - 2 \pi i j / L}}, \frac{[(Q^j)^{-1} \widetilde{P}^j]_2}{1 - e^{\lambda_2 \Delta t - 2 \pi i j / L}},\frac{1}{1 - e^{\Delta t - 2 \pi i j / L}} \Bigg)  = \Bigg( 1, 1 , \frac{1}{1 - e^{\Delta t - 2 \pi i j / L}} \Bigg) .
\end{align}
Observe we have 
\begin{equation}
CV \varphi_j = (1,1)^T
\end{equation}
which is constant, which also implies
\begin{equation}
\mathcal{L} > 0 ,
\end{equation}
and we do not get a contradiction with our theorem.

\vspace{2mm}

\textbf{Lemma 5.} Let $C \in \mathbb{R}^{m \times n}$ be wide and full row rank, $A \in \mathbb{R}^{n \times n}$ be square and positive semi-definite, and $Q \in \mathbb{R}^{n \times m}$ be tall and full column rank. Let $n \geq m > 1$. Let $C$ satisfy the permutation-invariant property as in Theorem 1. If $CAQ$ is full column rank for all possible column space choices of $C$, i.e.
\begin{equation}
\inf_{\sigma} \text{row} \Big( \text{concat}_i [C]_{\sigma_i(C)} A Q \Big) = m ,
\end{equation}
where $[C]_j$ is a column of $C$ and $\sigma$ is a permutation, then $A$ is necessarily full column rank.

\vspace{2mm}

\textit{Proof.} The overarching concept of this proof is, by the full row rank condition on $C$, we can permute the linearly independent columns of $C$ sufficiently well, which eventually forces a sufficient rank condition when all matrices are considered together.

\vspace{2mm}

Assume for contradiction that $A$ is not full column rank. Thus, since it is square, it cannot be full row rank either. This means there exists a $\psi$ such that $\psi^T A = 0$. Moreover, we also have $\psi^T AQ = 0$. We will show this can sometimes be true but never always be true for all considerations of $C$. In particular, we will show, by permuting $C$, the condition that $CAQ$ as full column rank is violated by permuting $C$ in such a way that a vector $\widetilde{\psi}$ such that $\widetilde{\psi}^T A Q =0$ is seen by the rowspace permutations. Thus, $\widetilde{\psi}^T AQ = 0$ is not possible, which means that $A$ cannot be not full rank either. It suffices to show
\begin{equation}
\text{dim} \Bigg( \text{row} \Bigg( [C_{\Sigma_0}=C] \oplus C_{\Sigma_1} \oplus \hdots \oplus C_{\Sigma_p} \Bigg) \Bigg) = n ,
\end{equation}
where $V \oplus W:= \{v+w: w \in V, w \in W\}$, and we have also denoted
\begin{align}
C_{\Sigma_p} & \in \{ C \ \text{such that all columns are permuted for some permutation} \ \Sigma_p \} 
\\
& \in \{ \text{concat}_i^p [C]_{\sigma_i(C)}, \Sigma_p = \{ \sigma_1(C), \hdots, \sigma_p(C) \} \ \text{is some permutation} \} .
\end{align}
In our proof,
Now, we will use notation $\Sigma^p$ to denote the power set of $\{[\Sigma_0=\text{id}],\Sigma_1,\hdots,\Sigma_p\}$, i.e. all possible combinations of size at least two, including those smaller in set size. First, we point out the fallacy
\begin{align}
& \text{dim} \Bigg( \text{row} \Bigg( [C_{\Sigma_0}=C] \oplus C_{\Sigma_1} \oplus \hdots \oplus C_{\Sigma_p} \Bigg) \Bigg) 
\\
& \neq  \sum_{i=1}^{p}  \text{dim} ( \text{row} ( C_{\Sigma_i} )) - \sum_{i, \Sigma = \{ \Sigma_{1_0},\hdots,\Sigma_{i_i}\} \in \Sigma^p} (-1)^{i+1} \text{dim} ( \text{row} ( \bigcap_{\Sigma_j \in \Sigma^p}  C_{\Sigma_j} )) .
\end{align}
The belief that this equality holds is a common misconception in mathematics. Instead, we will attempt to argue
\begin{equation}
\exists \ i_1,\hdots,i_d \ \text{such that} \ \text{dim} \Bigg( \text{row} \Bigg(C_{\Sigma_{i_1}} \oplus C_{\Sigma_{i_2}} \oplus \hdots \oplus C_{\Sigma_{i_d}} \Bigg) \Bigg)  = n ,
\end{equation}
but with a more written argument over using a closed formula. Trivially, since $C$ is full row rank, $\text{row}(C) = m$, but $m \leq n$ and more typically $m < n$ in a machine learning context. 

\vspace{2mm}

First, we note $m>1$ is important, otherwise of $m=1$ we can have degeneracy, i.e. all entries are identical, and our argument fails. Now, there must exist a column $[C]_j$ that is linearly dependent of the other columns. Since we are permuting columns, and since each $C_i$ is full row rank, there exists a permutation that places a column, linearly independent from at least one other column, into $[C]_{\sigma_j(C)} \rightarrow [C]_j$. Thus, there exists a new row that is replaced that spans a new dimension in the rowspace except under the condition the equation constructed is permutation-invariant, i.e. $c \sum x_i = 0$ for all constants $c$, i.e. $\sum x_i = 0$, where $x$ is a vector applied to $C_{\Sigma}$. Recall this is the exact condition as in Theorem 1. For example, the matrix
\begin{equation}
\Gamma = \begin{pmatrix} 1 & -1 & 0 & 0 \\ 0 & 1 & -1 & 0 \\ 0 & 0 & 1 & -1
\end{pmatrix}
\end{equation}
does not satisfy this permutation-invariant property since $\sum x_i = 0$ is true with respect to all rows. Aside from this scenario, this process is repeatable such that
\begin{equation}
\text{row}\Bigg( C_{\Sigma_i} \Bigg) \neq \text{row}\Bigg( C_{\Sigma_j} \Bigg) 
\end{equation}
for some permutations $\Sigma_i, \Sigma_j$, and by a pigeonhole-type argument, we eventually construct enough new matrices with row length $n$ where the collection of rows spans $\mathbb{R}^n$ in the sense that
\begin{equation}
\text{dim} \Bigg( \text{span} \Bigg\{ \psi_j \in \mathbb{R}^n : \psi_j \in \text{row} \Big( C_{\Sigma_i} \Big), i \ \text{such that} \ \Sigma_i \ \text{is a suitable permutation} \Bigg\} \Bigg) = n    
\end{equation}
and we are done. In particular, since $CAQ$ is full column rank under all necessary permutations of $C$, we never permute $C$ in such a way that it observes a vector in the perpendicular nullspace of $AQ$. This would mean $v^T A Q = 0$ for some row $v$ that is seen by the permutations of $C$, and this completes the proof.

$ \square $

\subsection{Observability with a more applicable Fourier transform}
\label{app:thm4}

\textbf{Theorem 4.} Consider the Fast Fourier transform of the matrix $\overline{K}$. Suppose $C \in \mathbb{R}^{m \times n}$ is full row rank, and $A \in \mathbb{R}^{n \times n}$ full column rank. Suppose $A = V \Lambda V^{-1}$ is diagonalizable with all distinct eigenvalues. Suppose it is true that 
\begin{align}
\text{ker} \Bigg( CV \Psi^{j_1}  V^{-1} \Bigg) \not\subseteq  \text{ker} \Bigg( CV \Psi^{j_2}  V^{-1} \Bigg)
\end{align}
for all $j_1, j_2$, where $\Psi$ is full column rank and is defined as 
\begin{align}
\Psi^j & =  \text{diag} \Bigg(  (1-\frac{e^{-2 \pi i \frac{j}{L} }}{1 - \frac{\Delta \lambda_1}{2}} + \frac{\Delta}{2} e^{-2 \pi i \frac{j}{L}} (\frac{1 + \frac{\Delta \lambda_1}{2}}{1 - \frac{\Delta \lambda_1}{2}}))^{-1} [ 1 - (\frac{1 + \frac{\Delta \lambda_1}{2}}{1 - \frac{\Delta \lambda_1}{2}})^L ],
\\
& \ \ \ \ \ \ \ \ \ \ \ \ \ \ \ \ \ \ \ \hdots ,  (1-\frac{e^{-2 \pi i \frac{j}{L} }}{1 - \frac{\Delta \lambda_n}{2}} + \frac{\Delta}{2} e^{-2 \pi i \frac{j}{L}} (\frac{1 + \frac{\Delta \lambda_n}{2}}{1 - \frac{\Delta \lambda_n}{2}}))^{-1} [ 1 - (\frac{1 + \frac{\Delta \lambda_n}{2}}{1 - \frac{\Delta \lambda_n}{2}})^L ] \Bigg) .
\end{align}
Then the system is observable with high probability.

\vspace{2mm}

\textit{Remark.} Our loss function to enforce the distinct eigenvalues and kernel condition is
\begin{align}
\mathcal{L} = & \ \sum_{j} \text{relu} \Bigg( \text{positive constant} - \min_{k_1,k_2} \Big| \Psi_{k_1 k_1}^{j} - \Psi_{k_2 k_2}^{j} \Big| \Bigg) 
\\
& + \sum_{j_1} \sum_{j_2 \neq j_1} \sum_k \text{relu} \Bigg( \text{positive constant} - \Bigg| \Psi_{kk}^{j_1} - \Psi_{kk}^{j_2} \Bigg|^2 \Bigg) 
\\
& +  \text{relu} \Bigg(  \Bigg| \frac{1}{n} \sum_k e^{i \theta_k(\Lambda) } \Bigg| - \text{positive constant}    \Bigg) .
\end{align}
If this is satisfied, then with high probability the system is observable.

\vspace{2mm}

\textit{Corollary.} Again, we can relax the condition $A$ is full column rank if we enforce
\begin{align}
\text{det} \Bigg( \mathcal{F} \Big[ \overline{K} \Big]_j \Bigg) > 0
\end{align}
for all $j$, but this is computationally harder.

\vspace{2mm}

\textit{Proof.} The key of this proof is in the fact that the observability matrix requires taking powers of $A$, but so does the Fourier transform. By ensuring sufficient row rank diversity across $j$, we have the system is guaranteed observable by this connection. We will use the fact that the Fourier transforms relies on the roots of unity as one of our key ingredients, which introduces nonlinearity and nonuniformity into a diagonalization.

\vspace{2mm}

To begin, we will show the Fourier transform can be written in diagonalized form. This will simplify our computation greatly. Starting with the Fourier transform, and following suit of \cite{gu2022efficientlymodelinglongsequences} and using the geometric series formula for matrices, we notice
\begin{align}
& \mathcal{F} \Big[ \overline{K}\Big]_j = \sum_{k=0}^{L-1} \overline{K}_k \exp \{ -2 \pi i \frac{jk}{L} \}
\\
& = \sum_{k=0}^{L-1} \overline{C} \overline{A}^k \overline{B} \exp \{ - 2\pi i \frac{jk}{L} \}
\\
& = C \Bigg( \sum_{k=0}^{L-1} \overline{A}^k \exp \{ - 2 \pi i \frac{jk}{L} \} \Bigg) \overline{B}
\\
& = C( I - \overline{A} \exp \{ -2 \pi i \frac{j}{L} \} )^{-1} (I - \overline{A}^{L} \exp \{ -2 \pi i j \} ) \overline{B} 
\\
& = C \Big( I - ( I - \frac{\Delta}{2} \cdot A)^{-1} ( I + \frac{\Delta}{2} \cdot A ) \exp \{ -2 \pi i \frac{j}{L} \} \Big)^{-1} 
\\
& \ \ \ \ \ \ \ \ \times \Bigg(I - \Bigg[ \prod_{k=1}^L ( I - \frac{\Delta}{2} \cdot A)^{-1} ( I + \frac{\Delta}{2} \cdot A )  \Bigg] \exp \{ -2 \pi i j \} \Bigg) ( I - \frac{\Delta}{2} \cdot A)^{-1} \Delta B  
\\
& = C \Big( I - [ V ( I - \frac{\Delta}{2} \cdot \Lambda) V^{-1} ]^{-1} ( I + \frac{\Delta}{2} \cdot A ) \exp \{ -2 \pi i \frac{j}{L} \} \Big)^{-1} 
\\
& \ \ \ \ \ \ \ \ \times \Bigg(I - \Bigg[ \prod_{k=1}^L [ V (I - \frac{\Delta}{2} \cdot \Lambda ) V^{-1}]^{-1} ( I + \frac{\Delta}{2} \cdot A )  \Bigg] \exp \{ -2 \pi i j \} \Bigg) ( I - \frac{\Delta}{2} \cdot A)^{-1} \Delta B  
\\
& = C \Big( I -  V \text{diag}(\frac{e^{-2 \pi i \frac{j}{L} }}{1 - \frac{\Delta \lambda_1}{2}},\hdots,\frac{e^{-2 \pi i \frac{j}{L} }}{1 - \frac{\Delta \lambda_n}{2}}) V^{-1}  ( I + \frac{\Delta}{2} \cdot A ) \Big)^{-1} 
\\
& \ \ \ \ \ \ \ \ \times \Bigg(I - \Bigg[ \prod_{k=1}^L V \text{diag}(\frac{e^{ -2 \pi i \frac{j}{L} }}{1 - \frac{\Delta \lambda_1}{2}},\hdots,\frac{e^{ -2 \pi i \frac{j}{L} }}{1 - \frac{\Delta \lambda_n}{2}}) V^{-1} (I + \frac{\Delta}{2} \cdot A) \Bigg] \Bigg) ( I - \frac{\Delta}{2} \cdot A)^{-1} \Delta B  .
\end{align}
Now, we make some notes. We will denote $(I + \frac{\Delta}{2} \cdot A) =: \Phi$, and 
\begin{equation}
Q := V \text{diag}(\frac{1}{1 - \frac{\Delta \lambda_1}{2}},\hdots,\frac{1}{1 - \frac{\Delta \lambda_n}{2}}) V^{-1} .   
\end{equation}
We make some other notes. First, notice
\begin{equation}
\label{eqn:FKj}
\mathcal{F} [ \overline{K} ]_j = C (I - Q \Phi e^{ -2 \pi i j / L })^{-1} ( I - (Q \Phi)^Le^{ -2 \pi i j } ) \overline{B} .
\end{equation}
Now, observe that
\begin{align}
& I - Q  \Phi e^{ -2 \pi i j / L }  \\
& = I - e^{ -2 \pi i j / L } V \text{diag}(\frac{1}{1 - \frac{\Delta \lambda_1}{2}},\hdots,\frac{1}{1 - \frac{\Delta \lambda_n}{2}}) V^{-1} (V \text{diag}(1 + \frac{\Delta \lambda_1}{2},\hdots,1 + \frac{\Delta \lambda_n}{2}) V^{-1} )
\\
&  = V\Bigg( \text{diag}(1 - e^{-2 \pi i \frac{j}{L}} ( \frac{1 + \frac{\Delta \lambda_1}{2}}{1 - \frac{\Delta \lambda_1}{2}} ) ,\hdots,1- e^{-2 \pi i \frac{j}{L}} ( \frac{1 + \frac{\Delta \lambda_n}{2}}{1 - \frac{\Delta \lambda_n}{2}} ) )\Bigg) V^{-1} .
\end{align}
Absorb the exponential into $Q^j$ for simplicity of notation. Thus,
\begin{align}
& (I - Q^j \Phi)^{-1} 
\\
& =  V\Bigg( \text{diag}\Big( (1-  e^{-2 \pi i \frac{j}{L}} (\frac{1 + \frac{\Delta \lambda_1}{2}}{1 - \frac{\Delta \lambda_1}{2}}))^{-1},\hdots,(1- e^{-2 \pi i \frac{j}{L}} (\frac{1 + \frac{\Delta \lambda_n}{2}}{1 - \frac{\Delta \lambda_n}{2}}))^{-1} \Big)\Bigg) V^{-1} .
\end{align}
Similarly, returning to \ref{eqn:FKj}, and using $e^{-2 \pi i j} = 1$, 
\begin{align}
& I - (Q \Phi)^L e^{ -2 \pi i j }
\\
& = I - \Bigg[ \prod_{i=1}^L V \text{diag}(\frac{1}{1 - \frac{\Delta \lambda_1}{2}},\hdots,\frac{1}{1 - \frac{\Delta \lambda_n}{2}}) V^{-1} (I + \frac{\Delta}{2} \cdot A) \Bigg]  
\\
& = I  -  \Bigg[ \prod_{i=1}^L V \text{diag}(\frac{1}{1 - \frac{\Delta \lambda_1}{2}},\hdots,\frac{1}{1 - \frac{\Delta \lambda_n}{2}}) \text{diag}(1 + \frac{\Delta \lambda_1}{2} ,\hdots,1 + \frac{\Delta \lambda_n}{2}) V^{-1} \Bigg] 
\\
& = I  - \Bigg[ \prod_{i=1}^L V \text{diag}(\frac{1 + \frac{\Delta \lambda_1}{2}}{1 - \frac{\Delta \lambda_1}{2}},\hdots,\frac{1 + \frac{\Delta \lambda_n}{2}}{1 - \frac{\Delta \lambda_n}{2}})  V^{-1}  \Bigg] 
\\
& = I  -  V \text{diag}\Big( (\frac{1 + \frac{\Delta \lambda_1}{2}}{1 - \frac{\Delta \lambda_1}{2}})^L,\hdots,(\frac{1 + \frac{\Delta \lambda_n}{2}}{1 - \frac{\Delta \lambda_n}{2}})^L \Big)  V^{-1} 
\\
& = V \text{diag}\Big( 1 - (\frac{1 + \frac{\Delta \lambda_1}{2}}{1 - \frac{\Delta \lambda_1}{2}})^L,\hdots,1-(\frac{1 + \frac{\Delta \lambda_n}{2}}{1 - \frac{\Delta \lambda_n}{2}})^L \Big)  V^{-1}
\end{align}
We mostly compute \ref{eqn:FKj} and conclude
\begin{align}
& (I - Q \Phi)^{-1} (I - (Q \Phi)^L) 
\\
& = V \text{diag} \Bigg(  \Big(1- e^{-2 \pi i \frac{j}{L}} (\frac{1 + \frac{\Delta \lambda_1}{2}}{1 - \frac{\Delta \lambda_1}{2}}) \Big)^{-1} \Bigg[ 1 - \Big(\frac{1 + \frac{\Delta \lambda_1}{2}}{1 - \frac{\Delta \lambda_1}{2}} \Big)^L \Bigg],
\\
& \ \ \ \ \ \ \ \ \ \ \ \ \ \ \ \ \ \ \ \hdots ,  \Big(1- e^{-2 \pi i \frac{j}{L}} (\frac{1 + \frac{\Delta \lambda_n}{2}}{1 - \frac{\Delta \lambda_n}{2}})\Big)^{-1} \Bigg[ 1 - \Big(\frac{1 + \frac{\Delta \lambda_n}{2}}{1 - \frac{\Delta \lambda_n}{2}} \Big)^L \Bigg] \Bigg) V^{-1} .
\end{align}
We have currently shown
\begin{align}
& \mathcal{F}[ \overline{K}]_j = C \Bigg( \sum_{k=0}^{L-1} \overline{A}^k \exp \{ - 2 \pi i \frac{jk}{L} \} \Bigg) \overline{B} 
\\
& = C V \Psi^j V^{-1} \overline{B} .
\end{align}
Here, we have defined (which is \ref{eqn:Psi_j})
\begin{align}
\label{eqn:psi_fourier}
\Psi^j & :=  \text{diag} \Bigg(  \Big(1- e^{-2 \pi i \frac{j}{L}} (\frac{1 + \frac{\Delta \lambda_1}{2}}{1 - \frac{\Delta \lambda_1}{2}}) \Big)^{-1} \Bigg[ 1 - \Big(\frac{1 + \frac{\Delta \lambda_1}{2}}{1 - \frac{\Delta \lambda_1}{2}} \Big)^L \Bigg], \hdots ,  \Big(1- e^{-2 \pi i \frac{j}{L}} (\frac{1 + \frac{\Delta \lambda_n}{2}}{1 - \frac{\Delta \lambda_n}{2}})\Big)^{-1} \Bigg[ 1 - \Big(\frac{1 + \frac{\Delta \lambda_n}{2}}{1 - \frac{\Delta \lambda_n}{2}} \Big)^L \Bigg] \Bigg) .
\end{align}
Our argument now lies in a permutation-type argument. Note that it is of interest to ensure
\begin{equation}
\text{ker}\Bigg( CV \Psi^{j_1} \Bigg) \not\subseteq \text{ker}\Bigg( CV \Psi^{j_2} \Bigg) 
\end{equation}
for all $j_1 \neq j_2$. By a pigeonhole-type argument. However, this argument is not airtight because the permuations are not guaranteed to fill out $\mathbb{R}^n$. The argument has greater satisfaction if $m$ is close to $n$ (i.e., if $m=n-1$, only one permutation is necessary to satisfy the rowspace condition), thus probability plays into this argument as well.

\vspace{2mm}

Instead, our argument is built upon $\Psi^j$ changes the rowspace in conjunction to the use of right multiplication of $V^{-1}$ on the right. In fact, it is only necessary to enforce 
\begin{equation}
\label{eqn:kernel_with_Vinv}
\text{ker}\Bigg( CV \Psi^{j_1} V^{-1} \Bigg) \not\subseteq \text{ker}\Bigg( CV \Psi^{j_2} V^{-1} \Bigg) ,
\end{equation}
and so $V^{-1}$ is now included. In particular, our argument lies in the fact that $\Psi^j$ sufficiently alters the nonzero columns of $CV$, and the rowspace union condition is satisfied with the way the other matrices interact with $V^{-1}$. As we will see, this will enforce observability.

\vspace{2mm}

\textbf{Claim.} Let $\Psi^j$ be the diagonal matrix defined in \ref{eqn:psi_fourier}. Let $C \in \mathbb{R}^{m \times n}$ be full row rank and wide, and let $V \in 
\mathbb{R}^{n \times n}$ be surjective. Then equation \ref{eqn:kernel_with_Vinv} is satisfied with high probability.

\vspace{2mm}

\textit{Sketch of proof.} We refer to arguments of section \ref{sec:fourier_nondegeneracy}, which we summarize again here more succinctly. The argument here is that $j$ is coupled nonlinearly with the eigenvalues of $\lambda^k$. This nonlinearity is more rich than simply exponentiating the eigenvalues of the form $\Lambda^k$, where $k$ is an exponent, and has greater probability to shift the eigenspace across $j$ under the condition that this nonlinearity is enforced with a loss. More specifically, eigenvalue exponentiation causes uniform scaling in the complex plane, since $(r e^{i \theta})^k = r^ke^{i k \theta}$ simply scales the radius and multiplies the angle the same across $\theta$. It is the same radial increment per power. On the other hand, our matrix $\Psi^j$ has intricate scaling, and the eigenvalues shift and disperse among the complex plane in a non-uniform manner. This helps contribute towards rowspace intersections of dimension less than $m$. It is with greater probability that $\Psi^j$ shifts the rowspace rather than exponentiating $\Lambda^k$. Moreover, we specifically constrain the eigenvalues to enforce this property. We have found the losses of
\begin{itemize}
\item (Sufficiently up to error tolerances) (D)istinct eigenvalues
\item Sufficient distinctness across $\Psi^j$ with respect to $j$
\item Sufficient angular diversity across the eigenvalues
\end{itemize}
work well. Intuitively, the first two are not hard to understand. As a remark, distinct eigenvalues do not imply distinctiveness across $\Psi^j$, as well as the constant can be different. The third one we discovered this through use of experimentation (see Figure \ref{fig:kernel_with_Psi_loss}). We demonstrate the phenomenon that $\Psi^j$ promotes kernel diversity empirically.

\vspace{2mm}

This motivates us to develop our loss function. By enforcing diversity of $\Psi^j$ across $j$, which correspond to the eigenvalues of $V \Psi^j V^{-1}$, we have a more mathematically rich basis to enforce the permutations. For example, if $\Psi^j$ is constant with respect to $j$, this causes no permutations at all, and observability is lost. This leads us to develop the loss term
\begin{align}
\mathcal{L} = \sum_{j_1} \sum_{j_2 \neq j_1} \sum_k \text{relu} \Bigg( \text{positive constant} - \Bigg| \Psi_{kk}^{j_1} - \Psi_{kk}^{j_2} \Bigg|^2 \Bigg) .
\end{align}
This will in term promote diverseness and help facilitate the permutation-like qualities of $\Psi^j$. Moreover, to ensure the dispersion of eigenvalues across the complex plane, we can also penalize angular differences in the complex exponents of $re^{i \theta}$. We also develop the loss
\begin{align}
\mathcal{L} = \text{relu} \Bigg(  \Bigg| \frac{1}{n} \sum_k e^{i \theta_k(\Lambda) } \Bigg| - \text{positive constant}    \Bigg)  .
\end{align}
Here, we denote $\theta(\cdot)$ the angular part of the complex number as an exponential. This loss function works by omitting the radial aspect of the eigenvalues, thus by only including the angles with a sense of normalization, angular diversity should lead to a balancing among the vectors. We desire the absolute summation close to zero. We have found empirical success in choosing angular diversity among the eigenvalues rather than across $j$ with $\Psi^j$.

\begin{figure}[!htbp]
  \vspace{0mm}
  \centering
  \includegraphics[scale=0.88]{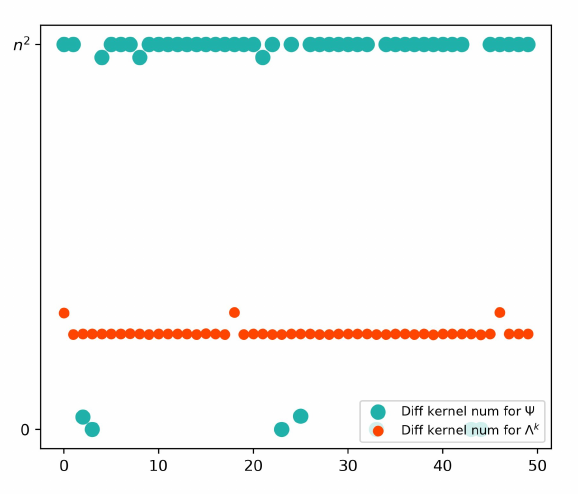}
  \caption{We illustrate the phenomenon that $\Psi^j$ is more likely to change the rowspace than simply taking powers of $\Lambda$. We randomly generate 50 matrices $C,V$ and compare the rowspace across all $n^2$ possibilities (since there are $n^2$ rows in the observability matrix) when multiplied together using both $\Psi^j$ and $\Lambda$ taken to a power. We have chosen $n=30, m =20$. The plotted point is the number of matrix combinations in which the kernels are not the same. Clearly, this plot illustrates what we need, as $\Psi^j$ rarely causes kernel intersection, while $\Lambda^k$ always does. In particular, the eigenvalues diminish in value across exponentiation, and cause an effect of triviality. This does not happen with our method. We remark just because taking powers of the eigenvalues causes numerical triviality, this does not imply a lack of observability, just that the test fails. Instead, we emphasize more closely that our method is observable rather than taking $\Lambda^k$ implies a lack thereof.}
  \label{fig:kernel_permutations}
\end{figure}

\begin{figure}[!htbp]
  \vspace{0mm}
  \centering
  \includegraphics[scale=0.88]{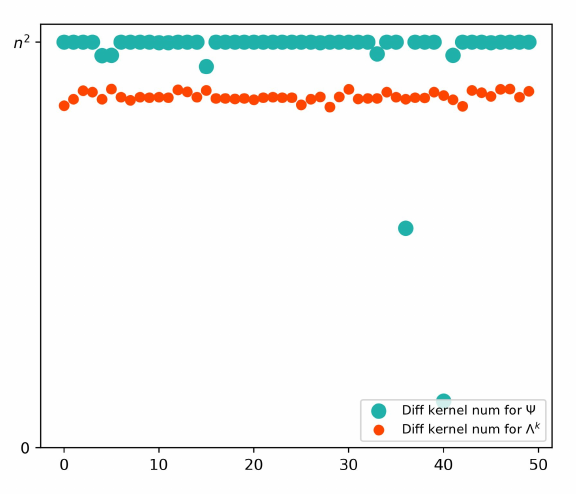}
  \caption{We illustrate the same phenomenon as Figure \ref{fig:kernel_permutations}, except we specifically choose eigenvalues here that are not diminished in value, and exhibit greater randomness. What we propose outperforms.}
  \label{fig:kernel_permutations_2}
\end{figure}

\begin{figure}[!htbp]
  \vspace{0mm}
  \centering
  \includegraphics[scale=0.88]{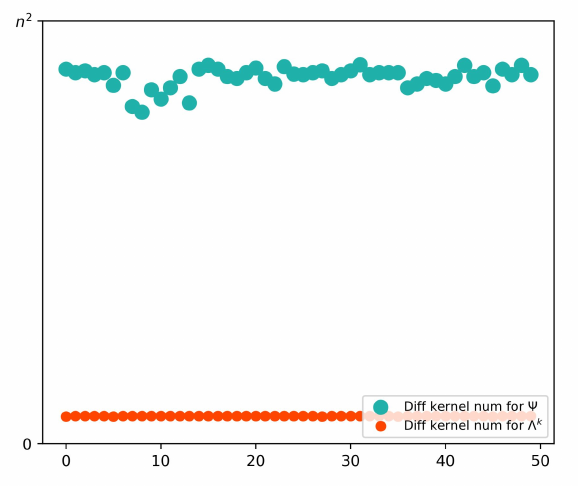}
  \caption{This figure goes hand-in-hand with Figure \ref{fig:eigenvalue_mamba}, as here we take eigenvalues of order $1\text{e}{-4}$. The reason our method outperforms so well here is because we can scale the eigenvalues however we wish in the definition of $\Psi^j$ using $\Delta$, which is a theoretical constant that can be changed in the learning task without drastically altering other facets of the learning. Here, $\Lambda$ is taken to the $k$-th power as before. In this figure, we have chosen $\Delta = 1$. Note that, in our proof purposes, $\Delta$ is irrelevant to that used in the state-space model.}
  \label{fig:kernel_permutations_4}
\end{figure}

\begin{figure}[!htbp]
  \vspace{0mm}
  \centering
  \includegraphics[scale=0.68]{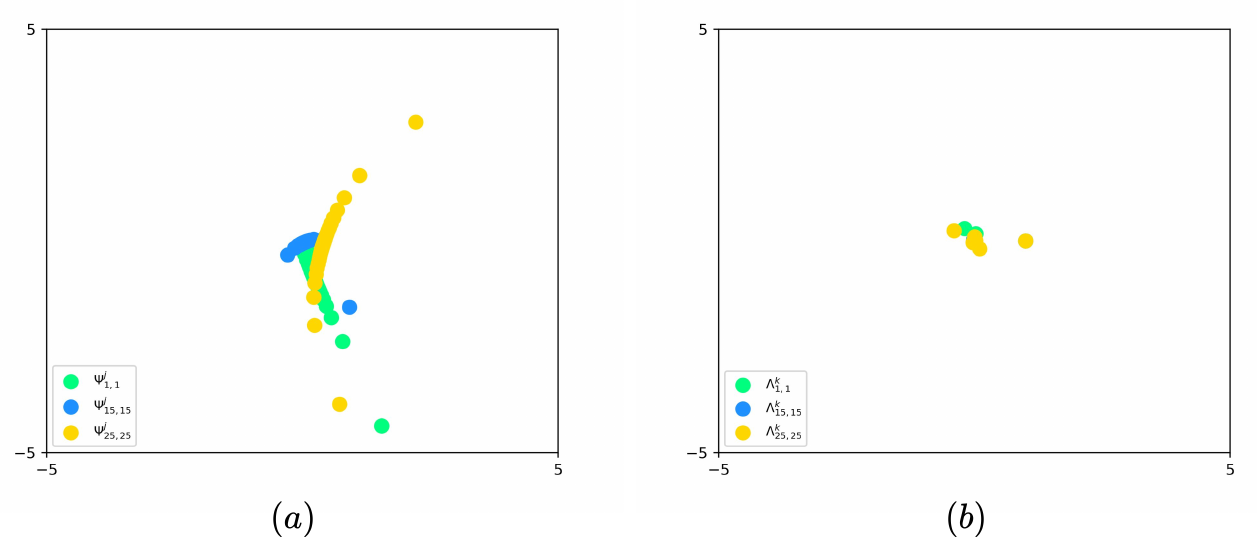}
  \caption{We compare the eigenvalues in a prototypical example of our method (a) versus standard diagonalization of (b). We plot three sequences of elements across $j$ of $\Psi^j$ in (a) and three sequences of elements across powers $k$ in $\Lambda^k$ in (b).}
  \label{fig:eigenvalue_comparison}
\end{figure}

\begin{figure}[!htbp]
  \vspace{0mm}
  \centering
  \includegraphics[scale=0.78]{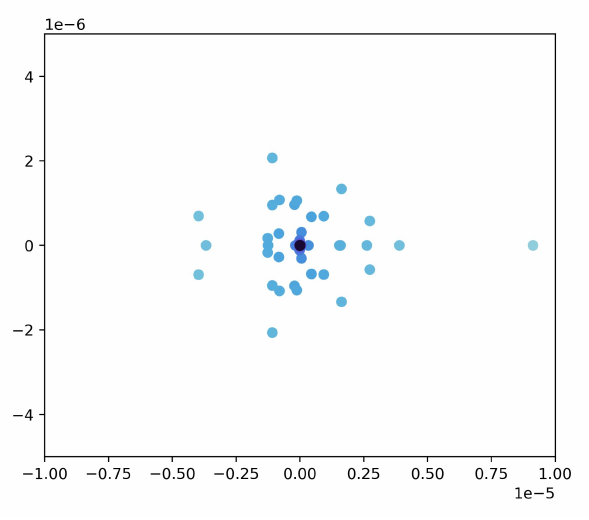}
  \caption{We train a Mamba state-space model on simple data of mappings between sinusoidal discretizations and examine the eigenvalues of $A$ in the complex plane. As we can see, we get very small values. Also, as we saw in the previous figures, what we propose outperforms with smaller eigenvalues in satisfying the kernel conditions.}
  \label{fig:eigenvalue_mamba}
\end{figure}

\begin{figure}[!htbp]
  \vspace{0mm}
  \centering
  \includegraphics[scale=0.78]{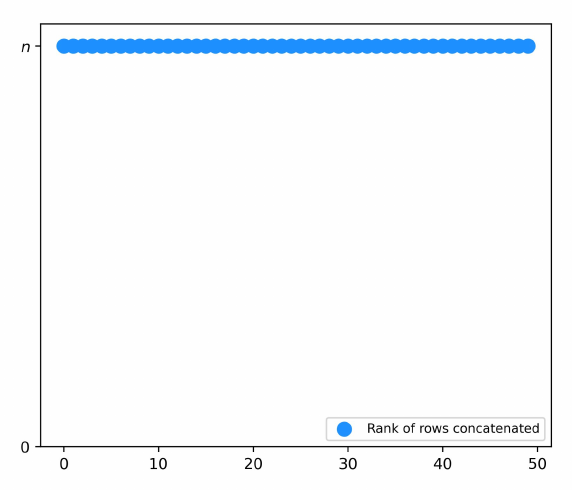}
  \caption{We illustrate the rank of the rows concatenated, or the union of their space, with our method, over 50 samples of $C,V$ and randomly sampled, fixed $\Lambda$. As we can see, all concatenations are of rank $n$, which is precisely what we desire. We taken eigenvalues of the order $1\text{e}{-1}$ in norm. We choose $m=50,n=100$. We remark that we even use a high tolerance of $1\text{e}{-5}$ (as opposed to say $1\text{e}{-10}$, which is significantly lower but still not unreasonable) to check the full rank condition (by checking the rank of the SVD).}
  \label{fig:rank_fig}
\end{figure}

\vspace{2mm}

We make a few crucial arguments. Note that the equation
\begin{equation}
C V \Psi^{j_1} V^{-1}  = Q C V \Psi^{j_2} V^{-1}  
\end{equation}
for some $Q$ if the kernels are the same. Such a $Q$ does indeed exist if the diagonal matrix $\Psi^{j_1}, \Psi^{j_2}$ are constant along the diagonal. This is because we can take $Q = \text{constant} \times I$ for that diagonal entry as the constant. Thus, it is necessary to also enforce all eigenvalues of $A$ are distinct. This is simple and and easy to enforce with machine learning by taking the loss
\begin{align}
\mathcal{L} = \text{relu} \Bigg( \text{positive constant} - \min_{j_1,j_2} \Big| \Lambda_{j_1,j_1} - \Lambda_{j_2,j_2} \Big| \Bigg) .
\end{align}

\vspace{2mm}

We make a final remark on the argument of the claim. It is possible that the new permutations do not observe enough new columns that add to the rowspace; however, this is unlikely considering we take a very large sample of permutations. There are many possible permutations that add at least one rowspace dimension, i.e. a column is swapped that adds rowspace. Moreover, permutations can add more than one column that adds rowspace at a time. It is with sufficiently high probability we make this argument. This argument holds more sufficiently well depending on the scale of $n$, since this governs the number of permutations, and how close $m$ is to $n$.

\vspace{2mm}

We show this gives the result. Take $\widetilde{\psi}$ to be any vector. For at least one $j$,  $CV\Psi^j V^{-1} \widetilde{\psi}  = C ( \sum_k \overline{A}^k \exp \{ -2 \pi i \frac{jk}{L} \})  \widetilde{\psi} \neq 0$ (we have used the fact that $V$ is surjective here). We must have $C \overline{A}^k b \neq 0$ for at least one select $k$. Thus, the pair $(C,\overline{A})$ is observable.

\vspace{2mm}

Since we defined $\overline{A} = ( I - \frac{\Delta}{2} \cdot A)^{-1} ( I + \frac{\Delta}{2} \cdot A )$, which is a  Cayley transform. Now, take an eigenvector of $A$, $v$. The Hautus lemma tells us we must have $Cv \neq 0$. Now, also recall if $v$ if an eigenvector of $Q$, then it is also an eigenvector of $Q^{-1}$ with an eigenvalue of the reciprocal. Thus, we have
\begin{align}
\label{eqn:cayley}
\overline{A}v = ( I - \frac{\Delta}{2} \cdot A)^{-1} ( I + \frac{\Delta}{2} \cdot A ) v = \frac{1 + \frac{\Delta}{2} \lambda }{ 1 - \frac{\Delta}{2} \lambda } v.
\end{align}
Thus $v$ is also an eigenvector of $\overline{A}$, so by the Hautus lemma we are done.

$ \square$

\vspace{2mm}

\textit{Remark.} Just as in Theorem 3, we can systematically construct $\psi$, but this comes at greater computational expense. We will simplify this proof by leaving our construction in terms of $\Psi^j$ and $V$. Construct $\widetilde{\psi}$ such that
\begin{align}
\widetilde{\psi} = \Bigg( \widetilde{\psi}_1,\hdots,\widetilde{\psi}_m, \underbrace{ 1, 1, \hdots, 1}_{n-m \ \text{times}} \Bigg) .
\end{align}
Letting $l$ be any row, we must have
\begin{align}
\sum_{k=1}^m \sum_{p=1}^n C_{lp} V_{pk} \Psi_{kk}^j \widetilde{\psi}_k + \sum_{k=m}^{n} \sum_{p=1}^n C_{lp} V_{pk} \Psi_{kk}^j  = 0.
\end{align} 
Equivalently, we define matrices with elements
\begin{align}
Q_{lk}^j =  \sum_{p=1}^n C_{lp} V_{pk} \Psi_{kk}^j \widetilde{\psi}_k , \ \ \ \ \ P_{lk}^j = \sum_{p=1}^n C_{lp} V_{pk} \Psi_{kk}^j, \ \ \ \ \ P_l^j = \sum_k \sum_p C_{lp} V_{pk} \Psi_{kk}^j . 
\end{align}
Thus, we solve
\begin{align}
Q^{j} ( \widetilde{\psi}_{1}, \widetilde{\psi}_{2}, \hdots, \widetilde{\psi}_{m} )^T + \widetilde{P}^{j}  = 0
\end{align}
Hence, we can solve the remaining $\widetilde{\psi}$ using
\begin{equation}
(\widetilde{\psi}_{1}, \widetilde{\psi}_{2  }, \hdots, \widetilde{\psi}_m )^T = - (Q^{j})^{-1} \widetilde{P}^{j}  .
\end{equation}
Hence, the sufficient condition so that $\widetilde{\psi} \in \text{ker}(CV \Psi^{j_1})$ or $\widetilde{\psi} \in \text{ker}(CV\Psi^{j_2})$ but not both is
\begin{align}
& \Bigg| \Bigg|   CV \text{column} \Bigg( \Psi_{11}^{j_2} [(Q^{j_2})^{-1} \widetilde{P}^{j_2}]_1, \hdots,  \Psi_{mm}^{j_2} [(Q^{j_2})^{-1} \widetilde{P}^{j_2}]_m,  \Psi_{(m+1)(m+1)}^{j_2}, \hdots,  \Psi_{nn}^{j_2} \Bigg) \Bigg| \Bigg| > 0 . 
\end{align}

\vspace{2mm}

\textit{Remark.} The proof does not automatically follow by taking $ C ( \sum_k \overline{A}^k \exp \{ -2 \pi i \frac{jk}{L} \}) b \neq 0$ for all columns of $b$ of $\overline{B}$. This is because $B$ is not surjective. Please see the counterexample in Theorem 3 for further exemplification.

\subsection{Observability for Mamba: an adapation of the Hautus eigenvector condition}
\label{app:thm5}

\textbf{Theorem 5 (modification of the Hautus-Lemma eigenvector test without testing eigenvectors).} Suppose $A \in \mathbb{R}^{n \times n}$ takes the form
\begin{equation}
A = V \text{diag} \Big( \lambda_1, \lambda_2, \hdots, \lambda_n \Big) V^{-1}.
\end{equation}
Here $A$ is diagonalized, $V$ is surjective. Suppose that $C \in \mathbb{R}^{m \times n}$. Suppose the loss
\begin{align}
\mathcal{L} = & \ \text{relu} \Bigg( \text{positive constant}  - \min_{k_1,k_2} \Bigg| \lambda_{k_1}  - \lambda_{k_2} \Bigg| \Bigg)  + \sum_j \text{relu} \Bigg( \text{positive constant}  - | \widetilde{C}_{1j} | \Bigg) 
\end{align}
is exactly satisfied, where $\widetilde{C} = CV$. Then the pair $(C,\overline{A})$ is observable. Moreover, the pair $(C,A)$ is observable.

\vspace{2mm}

\textit{Proof.} Recall $A$ is of the form
\begin{equation}
A = V \text{diag} \Big( \lambda_1, \lambda_2, \hdots, \lambda_n \Big) V^{-1} .
\end{equation}
Now, using the definition of $\overline{A}$, 
\begin{align}
\overline{A} & = \exp\{ \Delta \cdot A\} = \exp \{ \Delta V \text{diag} ( \lambda_1, \lambda_2, \hdots, \lambda_n )  V^{-1} \}  = \sum_{k=0}^{\infty} \frac{(V \Delta \text{diag} ( \lambda_1, \lambda_2, \hdots, \lambda_n ) V^{-1})^k }{k!}
\\
& = V  \Bigg( \sum_{k=0}^{\infty} \frac{( \Delta \text{diag} ( \lambda_1, \lambda_2, \hdots, \lambda_n ) )^k }{k!} \Bigg) V^{-1} = V \Bigg(   \text{diag}( \sum_{k=0}^{\infty} \frac{\Delta^k \lambda_1^k}{k!}, \sum_{k=0}^{\infty} \frac{\Delta^k \lambda_2^k}{k!},\hdots,\sum_{k=0}^{\infty} \frac{\Delta^k\lambda_n^k }{k!}) \Bigg) V^{-1}
\\
& = V \text{diag}(e^{\Delta \lambda_1}, e^{\Delta \lambda_2}, \hdots, e^{\Delta \lambda_n}) V^{-1} .
\end{align}

Now, consider
\begin{align}
\overline{O} &  = \begin{pmatrix}
C 
\\
C V \text{diag}(e^{\Delta \lambda_1}, e^{\Delta \lambda_2}, \hdots, e^{\Delta \lambda_n})
\\
C V \text{diag}(e^{2\Delta \lambda_1}, e^{2\Delta \lambda_2}, \hdots, e^{2\Delta \lambda_n})
\\
\vdots
\\
C V \text{diag}(e^{(n-1) \Delta \lambda_1}, e^{(n-1)\Delta \lambda_2}, \hdots, e^{(n-1)\Delta \lambda_n})
\end{pmatrix} 
V^{-1}  .
\end{align}
We will denote $\widetilde{C} = CV$. Since $V^{-1}$ is surjective,
\begin{align}
& \text{dim} \Bigg( \text{col} \Bigg( \begin{pmatrix}
C
\\
\widetilde{C} \text{diag}(e^{\Delta \lambda_1}, e^{\Delta \lambda_2}, \hdots, e^{\Delta \lambda_n})
\\
\widetilde{C} \text{diag}(e^{2\Delta \lambda_1}, e^{2\Delta \lambda_2}, \hdots, e^{2\Delta \lambda_n})
\\
\vdots
\\
\widetilde{C} \text{diag}(e^{(n-1) \Delta \lambda_1}, e^{(n-1)\Delta \lambda_2}, \hdots, e^{(n-1)\Delta \lambda_n})
\end{pmatrix} 
V^{-1} \Bigg) \Bigg)
\\
& = \text{dim} \Bigg( \text{col} \Bigg( \begin{pmatrix}
C
\\
\widetilde{C} \text{diag}(e^{\Delta \lambda_1}, e^{\Delta \lambda_2}, \hdots, e^{\Delta \lambda_n})
\\
\widetilde{C} \text{diag}(e^{2\Delta \lambda_1}, e^{2\Delta \lambda_2}, \hdots, e^{2\Delta \lambda_n})
\\
\vdots
\\
\widetilde{C} \text{diag}(e^{(n-1) \Delta \lambda_1}, e^{(n-1)\Delta \lambda_2}, \hdots, e^{(n-1)\Delta \lambda_n})
\end{pmatrix}  \Bigg) \Bigg) .
\end{align}
Now, observe
\begin{align}
\overline{O} = \begin{pmatrix}
\Phi_1 \otimes [\widetilde{C}]_1  
& 
\Phi_2 \otimes [\widetilde{C}]_2 
&
\hdots
&
\Phi_n \otimes [\widetilde{C}]_{n} 
\end{pmatrix} ,
\end{align}
for $i$-th column $[\widetilde{C}]_i$, and
\begin{align}
\label{eqn:vandermonde}
\Phi = 
\begin{pmatrix}
\Phi_1 & \Phi_2 & \hdots \Phi_n 
\end{pmatrix}
= 
\begin{pmatrix}
1 & 1 & \hdots & 1
\\
e^{\Delta \lambda_1} & e^{\Delta \lambda_2 } & \hdots & e^{\Delta \lambda_n}
\\
e^{2\Delta \lambda_1} & e^{2\Delta \lambda_2 } & \hdots & e^{2\Delta \lambda_n}
\\
\vdots & \vdots & \ddots & \vdots \\
e^{(n-1)\Delta \lambda_1} & e^{(n-1)\Delta \lambda_2 } & \hdots & e^{(n-1)\Delta \lambda_n}
\end{pmatrix}
.
\end{align}
Now, note that $\{ \Phi_j \}_j$ are all linearly independent, but clearly $\{ [\widetilde{C}]_j \}_j$ cannot be since $C$ is wide. Now, it is suffcient to enforce the condition
\begin{equation}
\text{dim} \Bigg( \text{col} \Bigg( ( \Phi_1 \otimes [\widetilde{C}]_1  ) \oplus ( \Phi_2 \otimes [\widetilde{C}]_2  ) \oplus \hdots \oplus ( \Phi_n \otimes [\widetilde{C}]_n ) \Bigg) \Bigg) = n
\end{equation}
to attain observability. 

\vspace{2mm}

We will enforce observability given the Kronecker product structure with the Vandermonde matrix. We will introduce a codependence on $C$ and $V$, in which our method improves efficiency by large scale. We will extract every $n$-th elements and concatenate this, and enforce that newly created matrix to be full column rank. In particular, we desire using \ref{eqn:vandermonde}
\begin{equation}
\Gamma = 
\begin{pmatrix}
\widetilde{C}_{11} & \widetilde{C}_{12} & \hdots & \widetilde{C}_{1n}
\\
e^{\Delta \lambda_1} \widetilde{C}_{11} & e^{\Delta \lambda_2 } \widetilde{C}_{12} & \hdots & e^{\Delta \lambda_n} \widetilde{C}_{1n}
\\
e^{2\Delta \lambda_1} \widetilde{C}_{11} & e^{2\Delta \lambda_2 } \widetilde{C}_{12} & \hdots & e^{2\Delta \lambda_n} \widetilde{C}_{1n}
\\
\vdots & \vdots & \ddots & \vdots \\
e^{(n-1)\Delta \lambda_1} \widetilde{C}_{11} & e^{(n-1)\Delta \lambda_2 } \widetilde{C}_{12} & \hdots & e^{(n-1)\Delta \lambda_n} \widetilde{C}_{1n}
\end{pmatrix}
\end{equation}
to be full column rank. This is very simple to enforce as full column rank. We only need the eigenvalues are distinct and all $\widetilde{C}_{1j}$ are nonzero.
Thus our loss function is 
\begin{align}
\mathcal{L} = & \ \text{relu} \Bigg( \text{positive constant}  - \min_{k_1,k_2} \Bigg| \lambda_{k_1}  - \lambda_{k_2} \Bigg| \Bigg)  + \sum_j \text{relu} \Bigg( \text{positive constant}  - | \widetilde{C}_{1j} | \Bigg)  ,
\end{align}
and we have the result. In particular, by enforcing this block as full column rank, the entire matrix is full column rank. Alternatively, we can relax this condition and instead consider the condition running over multiple sets of rows by taking
\begin{align}
\mathcal{L} = & \ \text{relu} \Bigg( \text{positive constant}  - \min_{k_1,k_2} \Bigg| \lambda_{k_1}  - \lambda_{k_2} \Bigg| \Bigg)  + \sum_i \sum_j \text{relu} \Bigg( \text{positive constant}  - | \widetilde{C}_{ij} | \Bigg)  .
\end{align}

\vspace{2mm}

Note that the columns of $\widetilde{C}$ can be expressed as a Kronecker product of a Vandermonde matrix. Note that we provide Lemma 6. The columns in our Vandermonde matrix are linearly independent by the distinctiveness of eigenvalues. By this lemma, we have the exact same result as the Hautus lemma.

\vspace{2mm}

Now, we are done in the $(C,\overline{A})$ case. The proof in the $(C,A)$ case follows immediately by using the $Ce^{A}h(0)=0 \implies h(0) = 0$ definition of observability. In particular, we have $Ce^{k \Delta A} h(0) = 0 \implies h(0) = 0$ for all $k$, which implies $Ce^{t A} h(0) \implies h(0) = 0$. Taking $t = k\Delta$, we have $(C,A)$ is observable.

$ \square $

\vspace{2mm}

\textbf{Lemma 6.} Let $\{v_i\}_{i=1}^n, v_i \in \mathbb{R}^n$ be linearly independent column vectors, and let $\{c_i\}_{i=1}^m, c_i \in \mathbb{R}^m$ be such that $c_i \neq 0$ for all $i$. Then the collection $\{ v_i \otimes c_i\}_i$ is linearly independent.
\vspace{2mm}

\textit{Proof.} Suppose for the sake of contradiction there exist $\{\alpha_i\}_i$ with at least two $\alpha_i \neq 0$ such that
\begin{equation}
\label{eqn:alphavc}
\sum_i \alpha_i ( v_i \otimes c_i ) = 0 .
\end{equation}
Let $v_{ji}$ denote the $j$-th element of $v_i$. Thus, it must also be true that from \ref{eqn:alphavc}
\begin{equation}
\sum_i \alpha_i \times v_{ji} \times  c_i  = 0 .
\end{equation}
In particular, $c_i \in \mathbb{R}^m$ is a column vector here, while $\alpha_i, v_{ji} \in \mathbb{R}$ are nonzero real numbers. Now, take a vector of nonzero constants, call it $\phi \in \mathbb{R}^m$, and take
\begin{equation}
\phi^T ( \sum_i \alpha_i \times v_{ji} \times  c_i ) = \sum_i \alpha_i v_{ji} (\phi^T c_i) = \sum_i \alpha_i \times v_{ji} \times  \phi_i = 0 ,
\end{equation}
and $\phi_i \in \mathbb{R}$, meaning we have reduced the vector $c_i$ to a nonzero real number. We have deliberately constructed $\phi$ such that $\phi^T c_i \neq 0$ for all $i$. Since the above holds for all $j$, we can regroup the coefficients as $\beta_i = \alpha_i \phi_i \in \mathbb{R}$, and we get
\begin{equation}
\sum_i \beta_i v_i = 0 .
\end{equation}
Thus, we have the contradiction of linear independence on $v$ and we are done.

$ \square $

\vspace{2mm}

\textit{Remark.} Alternatively, we can do this proof without exponentiation, and it follows similarly. The matrix in question to be full column rank in this case is
\begin{equation}
\Gamma' = 
\begin{pmatrix}
\widetilde{C}_{11} & \widetilde{C}_{12} & \hdots & \widetilde{C}_{1n}
\\
\lambda_1 \widetilde{C}_{11} & \lambda_2 \widetilde{C}_{12} & \hdots & \lambda_n \widetilde{C}_{1n}
\\
\lambda_1^2 \widetilde{C}_{11} & \lambda_2^2 \widetilde{C}_{12} & \hdots & \lambda_n^2 \widetilde{C}_{1n}
\\
\vdots & \vdots & \ddots & \vdots \\
\lambda_1^{n-1} \widetilde{C}_{11} & \lambda_2^{n-1} \widetilde{C}_{12} & \hdots &  \lambda_n^{n-1} \widetilde{C}_{1n}
\end{pmatrix} .
\end{equation}
This strategy does not change the loss function.

\subsection{Codependence: parameter reduction with efficient exponentiation for observability}
\label{app:codependence}

\begin{figure}[!h]
  \vspace{0mm}
  \centering
  \includegraphics[scale=0.98]{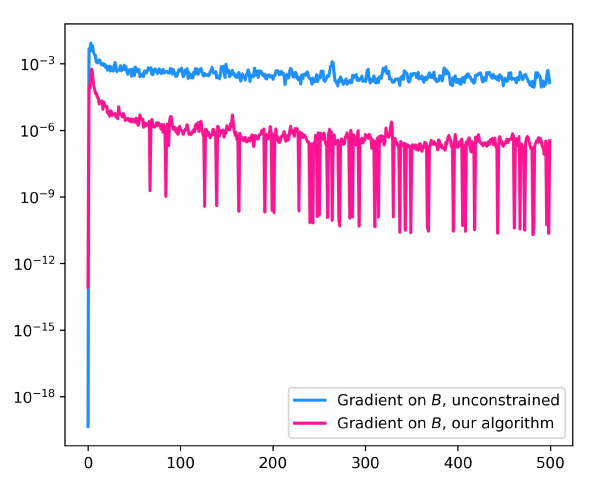}
  \caption{We illustrate results upon our proposed algorithm versus a more traditional algorithm. In this experiment, we use a Mamba architecture on basic sinusoidal sequences, and replace parameters with those we create. In particular, our algorithm yields more stable (stable in the sense of smaller in value, not in the sense of stochasticity) gradients with respect to $B, \mathcal{B}$, whereas a vanilla algorithm yields a gradient that is much higher. We mention training loss across both methods is relatively consistent, thus we remark the diminished gradient of our method is both stable and does not interfere with the training process. Our method is the gradient $\alpha \nabla_{\mathcal{B}} \mathcal{L}$ at a new step where $\alpha = || \mathcal{B}_k - \mathcal{B}_{k-1}||_F^{1/q}$ is the appropriate scaling for choice of $k$, and gradients at older steps are frozen. Gradients for the classical algorithm are the per step gradient $\nabla_{\mathcal{B}} \mathcal{L}$. In particular, our method slowly smooths into the correct parameter, thus is predictable and stable, whereas the gradients maintain large value in a more baseline method.}
  \label{fig:lipschitz_alg}
\end{figure}

\begin{figure}[!h]
  \vspace{0mm}
  \centering
  \includegraphics[scale=0.98]{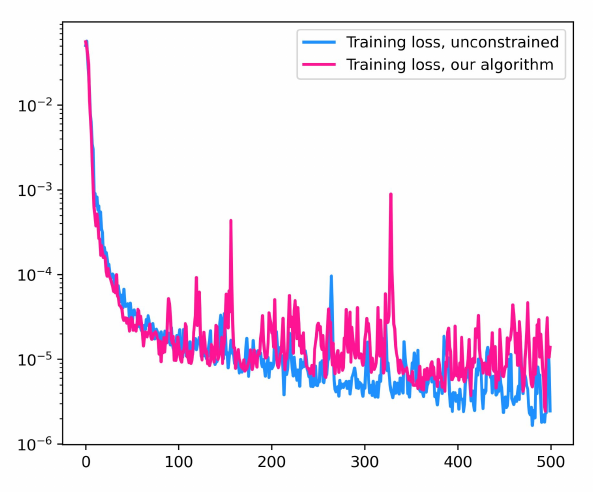}
  \caption{We illustrate training loss with our proposed algorithm versus a classical objective.}
  \label{fig:training_loss_coupled_eigendecomp}
\end{figure}

\textbf{Theorem 6.} Suppose loss $f$ decays at a rate
\begin{equation}
\EX \Big[ || (U_{k+1},Q_{k+1},S_{k+1}) - (U_{k},Q_{k},S_{k}) || \Big] = \EX \Big[ || \nabla f(U_k,Q_k,S_k) || \Big] \sim \Theta ( \frac{1}{k^q} ) ,
\end{equation} 
where $\frac{1}{2} \leq q \leq 1$. We remark this assumption is reasonable in certain nonconvex settings \cite{ward2021adagradstepsizessharpconvergence} \cite{arjevani2022lowerboundsnonconvexstochastic} \cite{carmon2017lowerboundsfindingstationary} \cite{carmon2019lowerboundsfindingstationary} \cite{convergence_rate_nonconvex} \cite{bubeck2015convexoptimizationalgorithmscomplexity} \cite{ghadimi2013stochasticfirstzerothordermethods} \cite{lei2019stochasticgradientdescentnonconvex}. Define
\begin{align}
& \mathcal{A}(Q,U,\Lambda) = (Q \otimes I_p)^T ( U \otimes I_p)^T \text{diag}(\lambda_1,\hdots,\lambda_n) (U \otimes I_p) (Q \otimes I_p)   
\\
& \mathcal{B}(\Lambda,U,S) = f(\text{diag}(\lambda_1,\hdots,\lambda_n))^{1/2}(U \otimes I) S .
\end{align}
Here, $p$ is some integer. Consider the training algorithm
\begin{align}
& A_{k+1} \leftarrow A_k + || \mathcal{A}(Q_k,U_k) - \mathcal{A}(Q_{k-1},U_{k-1}) ||_F^{1/2q} \cdot \mathcal{A}(Q_k,U_k,\Lambda_k)
\\
& B_{k+1} \leftarrow B_k + || \mathcal{B}(U_k,S_k) - \mathcal{B}(U_{k-1},S_{k-1}) ||_F^{1/q}  \cdot \mathcal{B}(\Lambda_k,U_k,S_k)
\\
& \text{Update} \ Q_{k+1},U_{k+1},S_{k+1}  \ \text{with a gradient descent-type optimizer} \ \mathcal{G} .
\end{align}
Suppose $Q,U$ are orthogonal and take the diagonalized form $Q = V_Q D_Q V_Q^T, U_i = V_U D_U V_U$ (we remark the entries can be complex-valued) where $D_Q, D_U$ are diagonal and learnable. Suppose that orthogonal $V$ are either fixed or updated only sometimes and finitely often. Then the algorithm satisfies a Robbins-Monro condition.

\vspace{2mm}

\textit{Remark.} The idea behind this algorithm is that gradient descent on $Q,U,S$ forces the updates in these parameters to diminish in differences over time, forcing the differences in $\mathcal{A}, \mathcal{B}$ to diminish as well. Note that our definition of $\mathcal{A}$ is quadratic in $Q, U$, so we will use this fact as well.

\vspace{2mm}

\textit{Proof.}  We begin with the case of $A$. It suffices to analyze $k \geq K$. For sake of argument, we will ignore the parameter updates on $V$ since they occur only finitely often and not every iteration, and it overall does not change the proof.  We will show
\begin{align}
& \sum_{k=K}^{\infty} \Bigg| \Bigg|    \mathcal{A}(Q_k,U_k) - \mathcal{A}(Q_{k-1},U_{k-1}) \Bigg| \Bigg|_F^{1/2q} = \infty 
\\
& \sum_{k=K}^{\infty} \Bigg| \Bigg|    \mathcal{A}(Q_k,U_k) - \mathcal{A}(Q_{k-1},U_{k-1}) \Bigg| \Bigg|_F^{1/q} < \infty  .
\end{align}
First, notice, by Lemma 7,
\begin{align}
& \sum_{k=1}^{\infty} \Bigg| \Bigg|    \mathcal{A}(Q_k,U_k) - \mathcal{A}(Q_{k-1},U_{k-1}) \Bigg| \Bigg|_F^{1/2q}
\\
& = \sum_k \Bigg| \Bigg| (Q_k \otimes I_p)^T ( U_k \otimes I_p)^T \text{diag}(\lambda_1,\hdots,\lambda_n) (U_k \otimes I_p) (Q_k \otimes I_p) 
\\
& \ \ \ \ \ \ \ \ \ \ \ \ \ \ \ \ \ \ \ \  - (Q_{k-1} \otimes I_p)^T ( U_{k-1} \otimes I_p)^T \text{diag}(\lambda_1,\hdots,\lambda_n) (U_{k-1} \otimes I_p) (Q_{k-1} \otimes I_p)  \Bigg| \Bigg|_F^{1/2q}
\\
& \geq \sum_k \Bigg| \Omega(\frac{1}{k^{2q}}) ( \widetilde{E} )  - | \text{other terms} | \Bigg|^{1/2q}
\\
& \geq \sum_k \frac{\Xi}{k} = \infty ,
\end{align}
where $\Xi$ is some constant and $\widetilde{E}$ is some matrix which is $\Theta(1)$ in norm. This finishes this case since the harmonic series diverges. Observe the other terms are irrelevant as long as they do not cancel out: if it is of order $1/k^{2q}$, we can regroup the terms up a constant; if it is order $1/k^{x}, x<2q$, then it still diverges; if it is of order $1/k^x, x>2q$, the totality of the terms still diverge by the first term dependent on $\Omega(1/k^{2q})$. Now, observe
\begin{align}
& \sum_{k=K}^{\infty} \Bigg| \Bigg|    \mathcal{A}(Q_k,U_k,\Lambda_k) - \mathcal{A}(Q_{k-1},U_{k-1},\Lambda_{k-1}) \Bigg| \Bigg|_F^{1/q}
\\
& = \sum_{k = K}^{\infty} \Bigg| \Bigg| (Q_k \otimes I_p)^T ( U_k \otimes I_p)^T \text{diag}(\lambda_1,\hdots,\lambda_n) (U_k \otimes I_p) (Q_k \otimes I_p) 
\\
& \ \ \ \ \ \ \ \ \ \ \ \ \ \ \ \ \ \ \ \  - (Q_{k-1} \otimes I_p)^T ( U_{k-1} \otimes I_p)^T \text{diag}(\lambda_1,\hdots,\lambda_n) (U_{k-1} \otimes I_p) (Q_{k-1} \otimes I_p)  \Bigg| \Bigg|_F^{1/q} = (1).
\end{align}
By Lemma 8 and its corollary, and since $X \otimes I$ is orthogonal if $X$ is, we see
\begin{align}
(1) & \leq \sum_{ k = K}^{\infty} \Bigg( \xi_1 \Big| \Big| ( Q_k \otimes I_p) - (Q_{k-1} \otimes I_p) \Big| \Big|_F^2 + \xi_2 \Big| \Big| (Q_k \otimes I_p) - (Q_{k-1}\otimes I_p) \Big| \Big|_F^4 
\\
& \ \ \ \ + \xi_3 \Big| \Big| (U_k \otimes I_p) - (U_{k-1} \otimes I_p) \Big| \Big|_F^2 + \xi_4 \Big| \Big| (U_k \otimes I_p) - (U_{k-1} \otimes I_p) \Big| \Big|_F^4 \Bigg)^{1/q}
\\
& = \sum_{k = K}^{\infty} \Bigg( \xi_1 p ||Q_k - Q_{k-1} ||_F^2 - \xi_2 p ||Q_k - Q_{k-1} ||_F^4 + \xi_3 ||U_k - U_{k-1} ||_F^2 + \xi_4 ||U_k - U_{k-1} ||_F^4 \Bigg)^{1/q}
\\
& \leq \sum_{k = K}^{\infty} \Bigg( \frac{ C_1\xi_1 p }{k^{2q}} + \frac{ C_2 \xi_2 p }{k^{4q}} + \frac{ C_3\xi_3 p }{k^{2q}} + \frac{ C_4\xi_4 p }{k^{2q} } \Bigg)^{1/q} < \infty ,
\end{align}
since the leading order term is $O(\frac{1}{k^2})$.

\vspace{2mm}

Next, we show
\begin{align}
& \sum_{k=K}^{\infty} \Bigg| \Bigg|    \mathcal{B}(U_k, S_k) - \mathcal{B}(U_{k-1}, S_{k-1}) \Bigg| \Bigg|_F^{1/q} = \infty 
\\
& \sum_{k=K}^{\infty} \Bigg| \Bigg|    \mathcal{B}(U_k,S_k) - \mathcal{B}(U_{k-1}, S_k) \Bigg| \Bigg|_F^{2/q} < \infty  .
\end{align}
This case is simpler. Notice that both $U,S \sim \Theta(\frac{1}{k^p})$
\begin{align}
& \sum_{k=K}^{\infty} \Bigg| \Bigg|    \mathcal{B}(U_k, S_k) - \mathcal{B}(U_{k-1}, S_{k-1}) \Bigg| \Bigg|_F^{1/q}
\\
& = \sum_{k = K}^{\infty} \Bigg| \Bigg| f(\text{diag}(\lambda_1,\hdots,\lambda_n))^{1/2} ( U_k \otimes I) S_k - f(\text{diag}(\lambda_1,\hdots,\lambda_n))^{1/2} ( U_{k-1} \otimes I) S_{k-1} \Bigg| \Bigg|_F^{1/q}
\\
& = \sum_{k = K}^{\infty} \xi_1 \Bigg| \Bigg|  ( U_k \otimes I) S_k -  ( U_{k-1} \otimes I) S_{k-1} \Bigg| \Bigg|_F^{1/q}
\\
& = \sum_{k = K}^{\infty} \xi_1 \Bigg| \Bigg|  ( (U_k -U_{k-1}) \otimes I) S_k +  ( U_{k-1} \otimes I) (S_k - S_{k-1}) \Bigg| \Bigg|_F^{1/q}
\\
& \geq \sum_{k = K}^{\infty} \Bigg| \xi_2 \Big| \Big|  U_k -U_{k-1} \Big| \Big|_F -  \xi_3 \Big| \Big| S_k - S_{k-1} \Big| \Big|_F \Bigg|^{1/q} 
\\
& \geq \sum_{k = K}^{\infty} \frac{\Xi}{k} = \infty .
\end{align}
Again, we use both the lower and upper bounds from $\Theta(1/k^q)$. Note that again the constants do not cancel. Following a similar argument but with the inequality the other way,
\begin{align}
& \sum_{k=K}^{\infty} \Bigg| \Bigg|    \mathcal{B}(U_k,S_k) - \mathcal{B}(U_{k-1}, S_k) \Bigg| \Bigg|_F^{2/q}
\\
& = \sum_{k = K}^{\infty} \xi_1 \Bigg| \Bigg|  ( (U_k -U_{k-1}) \otimes I) S_k +  ( U_{k-1} \otimes I) (S_k - S_{k-1}) \Bigg| \Bigg|_F^{2/q}
\\
& \leq \sum_{k = K}^{\infty} \Bigg( \xi_1 \Bigg| \Bigg| U_k - U_{k-1} \Bigg| \Bigg|_F + \xi_2 \Bigg| \Bigg| S_k - S_{k-1} \Bigg| \Bigg|_F \Bigg)^{2/q}
\\
& \leq \sum_{k = K}^{\infty} \Bigg( \frac{\xi_3}{k^q} + \frac{\xi_4}{k^q} \Bigg)^{2/q} < \infty.
\end{align}
This is quadratic in the denominator in its highest order, thus we have the result and the proof is completed.

$ \square $

\vspace{2mm}

\textbf{Lemma 7.} Let $B_i$ be a square matrix, and let $A_2^T A_1$ be symmetric. Suppose $||A_1 - A_2|| \sim \Theta(1/k^q)$. Then
\begin{align}
& ||(A_1 \otimes I) B_1 (A_1 \otimes I)^T - (A_2 \otimes I) B_2 (A_2 \otimes I)^T || \\
& \geq  \Bigg|  \Omega(\frac{1}{k^{2q}}) || ( E \otimes I) B_2 + B_2 ( E \otimes I) ||  -  || B_1 - B_2 + \Omega(\frac{1}{k^{2q}}) (E \otimes I) B_2 (E \otimes I)^T || \Bigg|  .
\end{align}

\vspace{2mm}

\textit{Proof.} Define $D = A_2^T A_1$. Since $D$ is symmetric by orthogonality of $A_1$, observe
\begin{align}
& ||(A_1 \otimes I) B_1 (A_1 \otimes I)^T - (A_2 \otimes I) B_2 (A_2 \otimes I)^T ||
\\
& =  ||(A_1 \otimes I) B_1 (A_1 \otimes I)^T - (A_1 \otimes I) (D \otimes I) B_2 (D \otimes I)^T (A_1 \otimes I)^T ||
\\
& = ||(A_1 \otimes I) ( B_1  -  (D \otimes I) B_2 (D \otimes I)^T  ) (A_1 \otimes I)^T || 
\\
\label{eqn:lemma_7_lastline}
& =  || B_1  -  (D \otimes I) B_2 (D \otimes I)^T  || .
\end{align}

\vspace{2mm}

\textbf{Claim.} We have
\begin{equation}
\xi || A_1 - A_2 ||^2  \leq || A_2^T A_1  - I || ,
\end{equation}
for some constant $\xi$.

\vspace{2mm}

\textit{Proof of claim.} First, we remark this proof does not need symmetry, only the upper bound does (see Lemma 8). Denote $X = A_1 - A_2$. Observe $A_1^T A_1 = (X + A_2)^T (X + A_2) = I$. After expanding, we see $X^T A_2 + A_2^T X = - X^T X$. Also note $A_2^T A_1 - I = A_2^T X$. Next, denote $Y = A_2^T X$. Notice that $Y^T + Y = -X^T X$, and so $||Y^T + Y|| = ||X||^2$. Thus, $2||Y|| \geq ||X||^2$. Substituting in what we have, we see $\frac{1}{2} ||A_1 - A_2||^2 \leq ||A_2^T X|| = ||A_2^T A_1 - I ||$ and we have the result.

\vspace{2mm}

Returning to the lemma, we can denote by the claim $D = I +\Omega(\frac{1}{k^{2q}})E $ for some matrix $E$. Hence, revisiting \ref{eqn:lemma_7_lastline}
\begin{align}
& \xi_1 || B_1  -  (D \otimes I) B_2 (D \otimes I)^T  ||
\\
& \xi_1 || B_1 - [(I + \Omega(\frac{1}{k^{2q}}) E) \otimes I] B_2 [(I + \Omega(\frac{1}{k^{2q}}) E) \otimes I]^T ||
\\
& =  || B_1 - B_2 + ( \Omega(\frac{1}{k^{2q}})  E \otimes I) B_2 ( \Omega(\frac{1}{k^{2q}})  E \otimes I)^T 
\\
& \ \ \ \ + (\Omega(\frac{1}{k^{2q}}) E \otimes I) B_2 + B_2 (\Omega(\frac{1}{k^{2q}}) E \otimes I)^T ||
\\
& \geq   \Bigg|  \Omega(\frac{1}{k^{2q}}) || ( E \otimes I) B_2 + B_2 (E \otimes I)||  -  || B_1 - B_2 + \Omega(\frac{1}{k^{2q}}) (E \otimes I) B_2 (E \otimes I)^T || \Bigg|  
\end{align}
by the reverse triangle inequality and the uniform estimates.

\vspace{2mm} 

\textbf{Lemma 8.} Let $A_i,B_i$ be square, symmetric matrices, and let $A_i^T A_j$ be symmetric. We have
\begin{equation}
|| A_1 B_1 A_1^T - A_2 B_2 A_2^T || \leq  \xi_1 ||A_1 - A_2||^2 + \xi_2 ||A_1 - A_2||^4 + 2 || B_1 - B_2 ||.
\end{equation}

\vspace{2mm}

\textit{Proof.} Define $D = A_2^T A_1$. Again using orthogonality, notice
\begin{align}
& || A_1 B_1 A_1^T - A_2 B_2 A_2^T ||
\\
& = || A_1 B_1 A_1^T - A_2 B_1 A_2^T + A_2 B_1 A_2^T - A_2 B_2 A_2^T ||
\\
& \leq || A_1 B_1 A_1^T - A_2 B_1 A_2^T || +  || A_2 B_1 A_2^T - A_2 B_2 A_2^T ||
\\
& = || A_2 D B_1 D A_2 - A_2 B_1 A_2^T || + || A_2 ( B_1 - B_2 ) A_2^T ||
\\
& \leq || A_2 (D B_1 D - B_2 ) A_2^T|| + ||A_2||^2 || B_1 - B_2 ||
\\
& = \xi_1 || D B_1 D - B_2  || + \xi_2 || B_1 - B_2 || .
\end{align}

\vspace{2mm}

\textbf{Claim.} We have
\begin{equation}
|| A_2^T A_1  - I || = || A_1 - A_2 ||^2  .
\end{equation}

\vspace{2mm}

\textit{Proof of claim.} We can first note
\begin{align}||A_2^T A_1 - I||_F^2 &= \text{Tr}( (A_2^T A_1 - I)^T (A_2^T A_1 - I) ) = \text{Tr}( (A_1^T A_2 - I)(A_2^T A_1 - I) ) \\
&= \text{Tr}( A_1^T A_2 A_2^T A_1 - A_1^T A_2 - A_2^T A_1 + I ) = \text{Tr}( 2I - A_1^T A_2 - A_2^T A_1 ) .
\end{align}
Now, we evaluate the distance between $A_1$ and $A_2$,
\begin{align}||A_1 - A_2||_F^2 &= \text{Tr}( (A_1 - A_2)^T (A_1 - A_2) ) \\
&= \text{Tr}( A_1^T A_1 - A_1^T A_2 - A_2^T A_1 + A_2^T A_2 ) \\
&= \text{Tr}( 2I - A_1^T A_2 - A_2^T A_1 ) .
\end{align}
Therefore, $||A_2^T A_1 - I||_F = ||A_1 - A_2||_F$ holds unconditionally.

\vspace{2mm}

Returning to the proof of the Lemma, Now, we denote $D = I + O(||A_1 - A_2||^2) E$ for some matrix $E$. We have
\begin{align}
|| D B_1 D - B_2  || & = ||  ( I + O(||A_1 - A_2||^2) E ) B_1 ( I + O(||A_1 - A_2||^2) E ) - B_2 ||
\\
&  \leq || B_1 - B_2 || + || O(||A_1 - A_2||^2)E  B_1|| 
\\
&  \ \ \ \ + || B_1 O(||A_1 - A_2||^2)E  ||  
\\
& \ \ \ \ + || O(||A_1 - A_2||^2) E B_1 O(||A_1 - A_2||^2)E  ||
\\
& \leq \xi_1 ||A_1 - A_2||^2 + \xi_2 ||A_1 - A_2||^4 +  || B_1 - B_2 ||
\end{align}
Here, $\xi_1,\xi_2 < \infty$ are constants, which are known and finite by the compactness of the domain under gradient descent. This completes the lemma.

$ \square $

\vspace{2mm}

\textit{Corollary.} We can iterate this inequality to get
\begin{align}
& || A_1 C_1 B C_1^T A_1^T - A_2 C_2 B C_2^T A_2^T ||
\\
& \leq \xi_1 ||A_1 - A_2 ||^2 + \xi_2||A_1 - A_2||^4 +  ||C_1 B_1 C_1^T - C_2 B_2 C_2^T ||
\\
& \leq \xi_1 ||A_1 - A_2 ||^2 + \xi_2||A_1 - A_2||^4 + \xi_3 ||C_1 - C_2 ||^2 + \xi_4 ||C_1 - C_2||^4 + \xi_5 ||B_1 - B_2 || .
\end{align}

\vspace{2mm}

\textbf{Theorem 7.} Suppose the hypotheses of Lemma 9 are satisfied. Define
\begin{align}
& \mathcal{A}(Q,U,\Lambda) = (Q \otimes I_p)^T ( U \otimes I_p)^T \text{diag}(\lambda_1,\hdots,\lambda_n) (U \otimes I_p) (Q \otimes I_p)   
\\
& \mathcal{B}(\Lambda,U,S) = f(\text{diag}(\lambda_1,\hdots,\lambda_n))^{1/2}(U \otimes I) S .
\end{align}
Here, $Q,U \in \mathbb{R}^{m \times m}$ are any matrices. Consider the training algorithm
\begin{align}
& A \leftarrow \mathcal{A}(Q,U)
\\
& B \leftarrow \mathcal{B}(U,S)
\\
& \text{Update} \ Q,U,S  \ \text{with a gradient descent-type optimizer} \ \mathcal{G} .
\end{align}
Then the algorithm does not satisfy a contraction mapping with Lipschitz constant $L \gg 1$.

\vspace{2mm}

\textit{Proof.} Consider any inequality on Euclidean metrics
\begin{align}
\Bigg| \Bigg| \mathcal{A}_1 - \mathcal{A}_2 \Bigg| \Bigg|_F^2 + \Bigg| \Bigg| \mathcal{B}_1 - \mathcal{B}_2 \Bigg| \Bigg|_F^2 \leq (L_A^2 + L_B^2) \Bigg| \Bigg| (Q_1,U_1,S_1) - (Q_2,U_2,S_2) \Bigg| \Bigg|_F^2 .
\end{align}
Here, $L_1,L_2$ are the appropriate Lipschitz constants with respect to $\mathcal{A},\mathcal{B}$. we require $L := L_A^2 + L_B^2 < 1$. The smallest $L$ possible is based on the supremum of the derivative, and it must be true that
\begin{align}
L  & \geq  \sup_{(B)}  \Big| \Big| \frac{\partial}{\partial B}( \text{Eigendecomposition}((U \otimes I)^T f( \text{diag}(\lambda_1,\hdots,\lambda_n)) ( U \otimes I))) \Big| \Big|_2^2  
\end{align}
By Lemma 9, and since we assumed the hypotheses are satisfied, we have this is greater than $1$, and we have the result.

\vspace{2mm}

\vspace{2mm}

\textbf{Lemma 9.} Let $Q,\lambda_j$ be the associated quantities with the eigendecomosition of a $\mathbb{R}^{n \times n}$ matrix.
The Lipschitz constant of the Eigendecomposition on such a $\mathbb{R}^{n \times n}$ matrix is greater than $1$ if 
\begin{equation}
\sqrt{   \frac{(n-1)}{ \lambda_{\text{max, gap}}^2} \sum_j  \lambda_j  }  -  \sqrt{  \sum_{j}  \frac{ 1 }{4 \lambda_j} 
 } > 1 .
\end{equation}
We have denoted $\lambda_{\text{max, gap}}$ the maximum difference in eigenvalues. Typically, this condition is satisfied, especially if $n$ is large.

\vspace{2mm}

\textit{Proof.} Observe
\begin{align}
& ||  \text{concat}_{ijkl} \frac{ \partial A{ij}}{\partial B_{kl}} ||{F}^2
\\ & = \sum{ijkl} \left| \sqrt{\lambda_j} \sum_{p \neq i} \frac{ Q_{pk} Q_{jl} }{\lambda_i - \lambda_p } Q_{ip} + \frac{ Q_{ij}Q_{kj}Q_{lj}}{2 \sqrt{\lambda_j}} \right|^2
\\
& \geq \Bigg( \sqrt{ \sum_{ijkl} \left| \sqrt{\lambda_j} \sum_{p \neq i} \frac{ Q_{pk} Q_{jl} }{\lambda_i - \lambda_p } Q_{ip} \right|^2 } - \sqrt{ \sum_{ijkl} \left| \frac{ Q_{ij}Q_{kj}Q_{lj}}{2 \sqrt{\lambda_j}} \right|^2 } \Bigg)^2
\\
& = \Bigg( \sqrt{ \sum_{ijkl} \lambda_j \sum_{p \neq i} \sum_{q \neq i} \frac{ Q_{ip} Q_{iq}  Q_{pk} Q_{jl} Q_{qk} Q_{jl}}{(\lambda_i - \lambda_p)(\lambda_i - \lambda_q)} } - \sqrt{ \sum_{ijkl} \frac{ Q_{ij}^2 Q_{kj}^2 Q_{lj}^2}{4 \lambda_j} } \Bigg)^2
\\
& = \Bigg( \sqrt{ \sum_{i,j} \lambda_j \sum_{p \neq i} \sum_{q \neq i} \frac{ Q_{ip} Q_{iq} }{(\lambda_i - \lambda_p)(\lambda_i - \lambda_q)} \Big(\sum_k Q_{pk} Q_{qk}\Big) \Big(\sum_l Q_{jl}^2\Big) } \\
& \ \ \ \ \ \ \ \ \ \ \ \ \ \ \ \ \ \ \ \ \ \ \ \ \ \ \ \ \ \ \ \ \ \ \ \ \ \ \ \ \ \ - \sqrt{ \sum_{j} \frac{ (\sum_i Q_{ij}^2 )(\sum_k Q_{kj}^2 )(\sum_l Q_{lj}^2)}{4 \lambda_j} } \Bigg)^2
\\
& = \Bigg( \sqrt{ \sum_{i,j} \lambda_j \sum_{p \neq i} \sum_{q \neq i} \frac{ Q_{ip} Q_{iq} \delta_{pq} (1) }{(\lambda_i - \lambda_p)(\lambda_i - \lambda_q)} } - \sqrt{ \sum_{j} \frac{ 1 }{4 \lambda_j} } \Bigg)^2
\\
& = \Bigg( \sqrt{ \sum_{i,j} \lambda_j \sum_{p \neq i} \frac{ Q_{ip}^2 }{(\lambda_i - \lambda_p)^2} } - \sqrt{ \sum_{j} \frac{ 1 }{4 \lambda_j} } \Bigg)^2
\\
& \geq \Bigg( \sqrt{ \sum_{i,j} \lambda_j \sum_{p \neq i} \frac{ Q_{ip}^2 }{\lambda_{\text{max, gap}}^2} } - \sqrt{ \sum_{j} \frac{ 1 }{4 \lambda_j} } \Bigg)^2
\\
& = \Bigg( \sqrt{ \frac{1}{\lambda_{\text{max, gap}}^2} \sum_j \lambda_j \sum_i (1 - Q_{ii}^2) } - \sqrt{ \sum_{j} \frac{ 1 }{4 \lambda_j} } \Bigg)^2
\\
& = \Bigg( \sqrt{ \frac{(n - \sum_i Q_{ii}^2)}{\lambda_{\text{max, gap}}^2} \sum_j \lambda_j } - \sqrt{ \sum_{j} \frac{ 1 }{4 \lambda_j} } \Bigg)^2 .
\end{align}
The first term is typically quite large, and the second term is typically quite small, and we have the result. We also remark, in this proof, we have assumed the first term is larger than the second, but the assumption implies this.

$ \square $

\vspace{2mm}

\textbf{Lemma 10.} Denote $A = \text{Eigendecomposition}(B) = \text{Eigen}(B)$. We have
\begin{equation}
\frac{ \partial A_{ij}}{\partial B_{kl}} = \sqrt{\lambda_j} \sum_{p \neq i} \frac{ Q_{pk} Q_{jl} }{\lambda_i - \lambda_p } Q_{ip}   +  \frac{ Q_{ij}Q_{kj}Q_{lj}}{2 \sqrt{\lambda_j}} .
\end{equation}

\vspace{2mm}

\textit{Proof.} Note that
\begin{equation} 
A_{ij} = Q_{ij} \sqrt{\lambda_j} .
\end{equation}
Also observe that this formula holds when $A$ is not square as long as $A = Q\text{diag}(\sqrt{\lambda_1},\hdots,\sqrt{\lambda_n})$, where $Q$ is a collection of eigenvectors, but not necessarily all of them. The proof proceeds as the chain rule. Observe
\begin{align}
\frac{ \partial A_{ij} }{\partial B_{kl}} & = \frac{\partial}{\partial B_{kl}} (  Q_{ij} \sqrt{\lambda_j} )
\\
& = \sqrt{\lambda_j} \frac{ \partial Q_{ij}}{\partial B_{kl}} + Q_{ij} \frac{1}{2\sqrt{\lambda_j}} \frac{\partial \lambda_j}{\partial B_{kl}} .
\end{align}
Observe the formula for $\partial Q_{ij} / \partial B_{kl}$ is given by \cite{eigenvector_deriv}
\begin{equation}
\frac{\partial Q_{ij}}{\partial B_{kl}} = \sum_{p \neq i} \frac{ Q_{pk} Q_{jl} }{\lambda_i - \lambda_p } Q_{ip}
\end{equation}
Now, the derivative of the eigenvalue is
\begin{equation}
\frac{ \partial \lambda_j}{\partial B_{kl} } = \langle [Q]_j, e_k \rangle \langle [Q]_j, e_l \rangle = Q_{kj} Q_{lj} .
\end{equation}
Putting everything together, we have the result.

$\square$

\subsection{Eigenvalues distinct does not imply $\Psi$ distinct}

\textbf{Lemma 11.} For sequence length $L \geq 2$, the mapping from the eigenvalues $\lambda_k$ to the diagonal entries $\Psi_{kk}^j$ is not injective. That is, distinct eigenvalues $\lambda_1 \neq \lambda_2$ do not guarantee distinct diagonal entries $\Psi_{11}^j \neq \Psi_{22}^j$.

\vspace{2mm}

\textit{Proof.} Let the mapping from the eigenvalue $\lambda_k$ to the $k$-th diagonal entry of $\Psi^j$ be denoted by $f(\lambda_k) = \Psi_{kk}^j$. We define the Cayley transformation $z: \mathbb{C} \setminus \{\frac{2}{\Delta}\} \to \mathbb{C} \setminus \{-1\}$ as
\begin{align}
\label{eqn:z}
z_k = \frac{1 + \frac{\Delta \lambda_k}{2}}{1 - \frac{\Delta \lambda_k}{2}} .
\end{align}
Because this is a Möbius transformation, it is bijective. Therefore, proving that $f$ is non-injective with respect to the eigenvalues $\lambda_k$ is equivalent to proving it is non-injective with respect to the transformed $z_k$. Recall the definition of $\Psi^j$ in \ref{eqn:Psi_j}, where the $k$-th diagonal entry is
\begin{align}
\Psi_{kk}^j = \left(1 - e^{-2 \pi i \frac{j}{L}} \left(\frac{1 + \frac{\Delta \lambda_k}{2}}{1 - \frac{\Delta \lambda_k}{2}}\right)\right)^{-1} \left[ 1 - \left(\frac{1 + \frac{\Delta \lambda_k}{2}}{1 - \frac{\Delta \lambda_k}{2}}\right)^L \right] .
\end{align}
Let $c_j = e^{-2 \pi i \frac{j}{L}}$ be a root of unity. By substituting $z_k$ and $c_j$ into the expression for $\Psi_{kk}^j$, the map $z_k$ to $\Psi_{kk}^j$ simplifies to
\begin{align}
\label{eqn:Psi_zk}
\Psi_{kk}^j = \frac{1 - z_k^L}{1 - c_j z_k} .
\end{align}
To show this mapping is not one-to-one for $L \ge 2$, we fix a complex output $y \in \mathbb{C}$ and set $\Psi_{kk}^j = y$, so \ref{eqn:Psi_zk} becomes
\begin{align}
y = \frac{1 - z_k^L}{1 - c_j z_k} .
\end{align}
Rearranging this yields
\begin{align}z_k^L - (c_j y) z_k + (y - 1) = 0 .
\end{align}
For any fixed $y$, this is a polynomial in $z_k$ of degree $L$. By the Fundamental Theorem of Algebra, this polynomial has exactly $L$ roots in the complex plane, counting multiplicity. Because the set of critical values (where roots are degenerate) for any polynomial is finite, there exist infinitely many choices of $y$ that yield distinct roots. Therefore, for $L \ge 2$, there exist distinct $z_1 \neq z_2$ such that $f(z_1) = f(z_2) = y$. Because the Cayley transform is bijective, these correspond to distinct eigenvalues $\lambda_1 \neq \lambda_2$ that yield identical diagonal entries $\Psi_{11}^j = \Psi_{22}^j$.

$ \square $

\end{document}